\documentclass{article}

\PassOptionsToPackage{numbers,sort&compress}{natbib}

\usepackage[main, final]{neurips_2026}

\usepackage[utf8]{inputenc}
\usepackage[T1]{fontenc}
\usepackage{hyperref}
\usepackage{url}
\usepackage{booktabs}
\usepackage{multirow}
\usepackage{amsfonts}
\usepackage{nicefrac}
\usepackage{microtype}
\usepackage[dvipsnames]{xcolor}
\usepackage{graphicx}
\usepackage{subcaption}
\usepackage{enumitem}
\usepackage{placeins}
\usepackage{caption}
\usepackage{float}
\usepackage{wrapfig}
\usepackage{amsmath}
\usepackage{amssymb}
\usepackage{mathtools}
\usepackage{amsthm}
\usepackage{algorithm}
\usepackage{algpseudocode}
\usepackage[capitalize,noabbrev]{cleveref}
\usepackage{thm-restate}
\usepackage[textsize=tiny]{todonotes}
\usepackage{mdframed}
\usepackage[most]{tcolorbox}

\hypersetup{
    colorlinks,
    linkcolor={red!50!black},
    citecolor={blue!50!black},
    urlcolor={blue!80!black}
}

\usepackage{minitoc}

\doparttoc                   %
\mtcsetdepth{parttoc}{3}

\theoremstyle{plain}
\newtheorem{theorem}{Theorem}[section]
\newtheorem{proposition}[theorem]{Proposition}
\newtheorem{lemma}[theorem]{Lemma}
\newtheorem{corollary}[theorem]{Corollary}
\theoremstyle{definition}

\theoremstyle{remark}
\newtheorem{remark}[theorem]{Remark}

\newmdenv[
  linecolor=pink!60!purple,
  backgroundcolor=pink!30,
  linewidth=0pt,
  roundcorner=2pt,
  skipabove=3pt,
  skipbelow=1pt,
  innerleftmargin=3pt,
  innerrightmargin=3pt,
  innertopmargin=3pt,
  innerbottommargin=3pt
]{pinkbox}

\newmdenv[
  linecolor=blue!60!cyan,
  backgroundcolor=blue!10,
  linewidth=0pt,
  roundcorner=2pt,
  skipabove=3pt,
  skipbelow=1pt,
  innerleftmargin=3pt,
  innerrightmargin=3pt,
  innertopmargin=3pt,
  innerbottommargin=3pt
]{bluebox}

\tcbset{colback=blue!5!white, colframe=blue!40!black, boxrule=0.5pt, arc=3pt, left=6pt, right=6pt, top=3pt, bottom=3pt}
\newtcolorbox{propbox}{colback=blue!5!white, colframe=blue!40!black, boxrule=0.5pt, arc=3pt, left=3pt, right=3pt, top=2pt, bottom=2pt}

\title{Test-time Scaling of Diffusions with Flow Maps}

\author{%
  Amirmojtaba Sabour$^{*,1,2,3}$ 
  \And
  Michael S. Albergo$^{*,4,5,6}$ 
  \And
  Carles Domingo-Enrich$^{*,7}$ 
  \AND
  Nicholas M. Boffi$^{8}$ 
  \And
  Sanja Fidler$^{1,2,3}$ 
  \And
  Karsten Kreis$^{\dagger, 1}$ 
  \And
  Eric Vanden-Eijnden$^{\dagger, 9, 10}$
}

\begin{document}

\maketitle
\def\thefootnote{$*$}\footnotetext{Equal contribution}
\def\thefootnote{$\dagger$}\footnotetext{Equal supervision}
\def\thefootnote{\arabic{footnote}}

\begin{center}
    \vspace{-8mm}
    \footnotesize
    $^1$ NVIDIA \hspace{1mm}
    $^2$ University of Toronto \hspace{1mm}
    $^3$ Vector Institute \hspace{1mm}
    $^4$ Harvard University \hspace{1mm}
    $^5$ Kempner Institute \hspace{1mm} 
    $^6$ IAIFI \hspace{1mm}
    $^7$ Microsoft Research \hspace{1mm}
    $^8$ Carnegie Mellon University \hspace{1mm}
    $^9$ Courant Institute, New York University \hspace{1mm} 
    $^{10}$ Capital Fund Management (CFM) 
\end{center}

\begin{center}   
    \textit{Project Page: }
    \href{https://flow-map-trajectory-tilting.github.io/}{
    \textcolor{RubineRed}{https://flow-map-trajectory-tilting.github.io/}}
\end{center}

\begin{figure}[H]
  \centering
  \includegraphics[width=\linewidth]{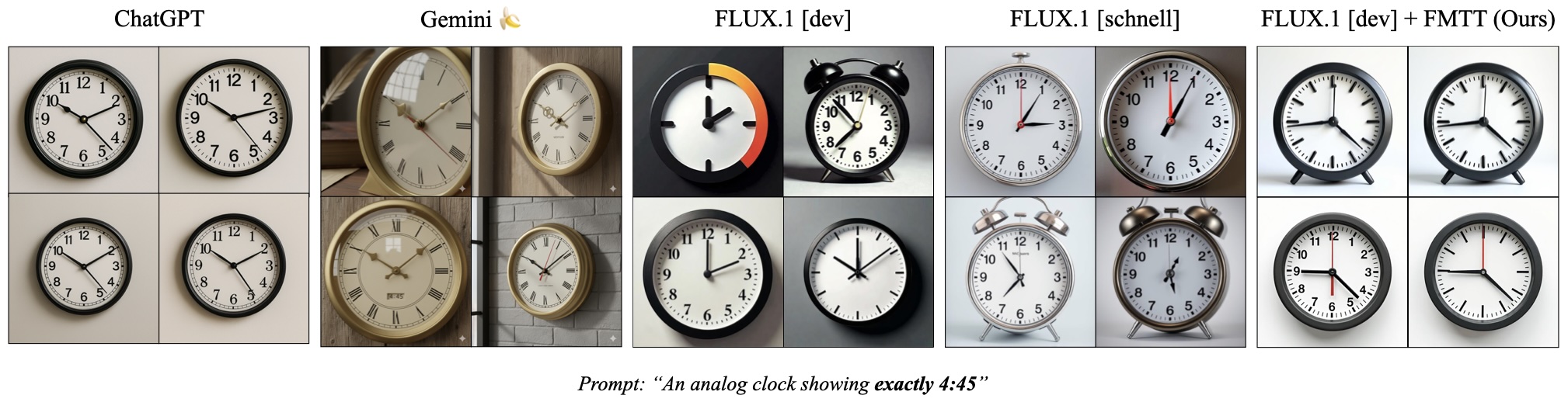}
  \caption{Flow Map Trajectory Tilting (FMTT) implements efficient and accurate test-time search with flow map look-ahead that can overcome model biases and reliably sample from regions of the distribution (e.g., precise clock times) that baselines fail to capture.\looseness=-1}
  \label{fig:teaser}
\end{figure}

\doparttoc %
\faketableofcontents %

\begin{abstract}
A common recipe to improve diffusion models at test-time so that samples score highly against a user-specified reward is to introduce the gradient of the reward into the dynamics of the diffusion itself. This procedure is often ill posed, as user-specified rewards are usually only well defined on the data distribution at the end of generation. While common workarounds to this problem are to use an approximate denoiser to estimate what a sample would have been at the end of generation, we propose a simple solution to this problem by working directly with a flow map. By exploiting a relationship between the flow map and velocity field governing the instantaneous transport, we construct an algorithm, Flow Map Trajectory Tilting (FMTT), which leverages a fast yet precise flow map look-ahead and provably performs better ascent on the reward than standard test-time methods involving the gradient of the reward. The approach can be used to either perform exact sampling via importance weighting or principled search that identifies local maximizers of the reward-tilted distribution. We demonstrate the efficacy of our approach against other look-ahead techniques, and show how the flow map enables engagement with complicated reward functions that make possible new forms of image editing, e.g. by interfacing with vision language models. \looseness=-1
\end{abstract}

\section{Introduction}

Large-scale foundation models built from diffusions \citep{ho2020denoising, song2020score} or flow-based transport \citep{lipman2022flow, albergo2022building, albergo2023stochastic, liu2022flow} have become highly successful tools across computer vision and scientific domains.
In this paradigm, generation amounts to numerically solving an ordinary or stochastic differential equation (ODE/SDE) whose coefficients are learned neural networks.
An active area of research is how to best \textit{adapt} these dynamics at inference time to produce samples that align with a user-specified reward. 
For example, as shown in Figure \ref{fig:teaser}, a user may want an image of a clock displaying a precise time, which is often generated inaccurately without suitable adaptation.
These approaches, often referred to as \textit{guidance}, require no additional re-training and are therefore orthogonal to fine-tuning methods, which modify the model itself through an additional learning stage to modify the quality of generated samples.\looseness=-1

\begin{wrapfigure}[17]{r}{0.45\textwidth}
    \vspace{-5mm}
    \centering
    \includegraphics[width=\linewidth]{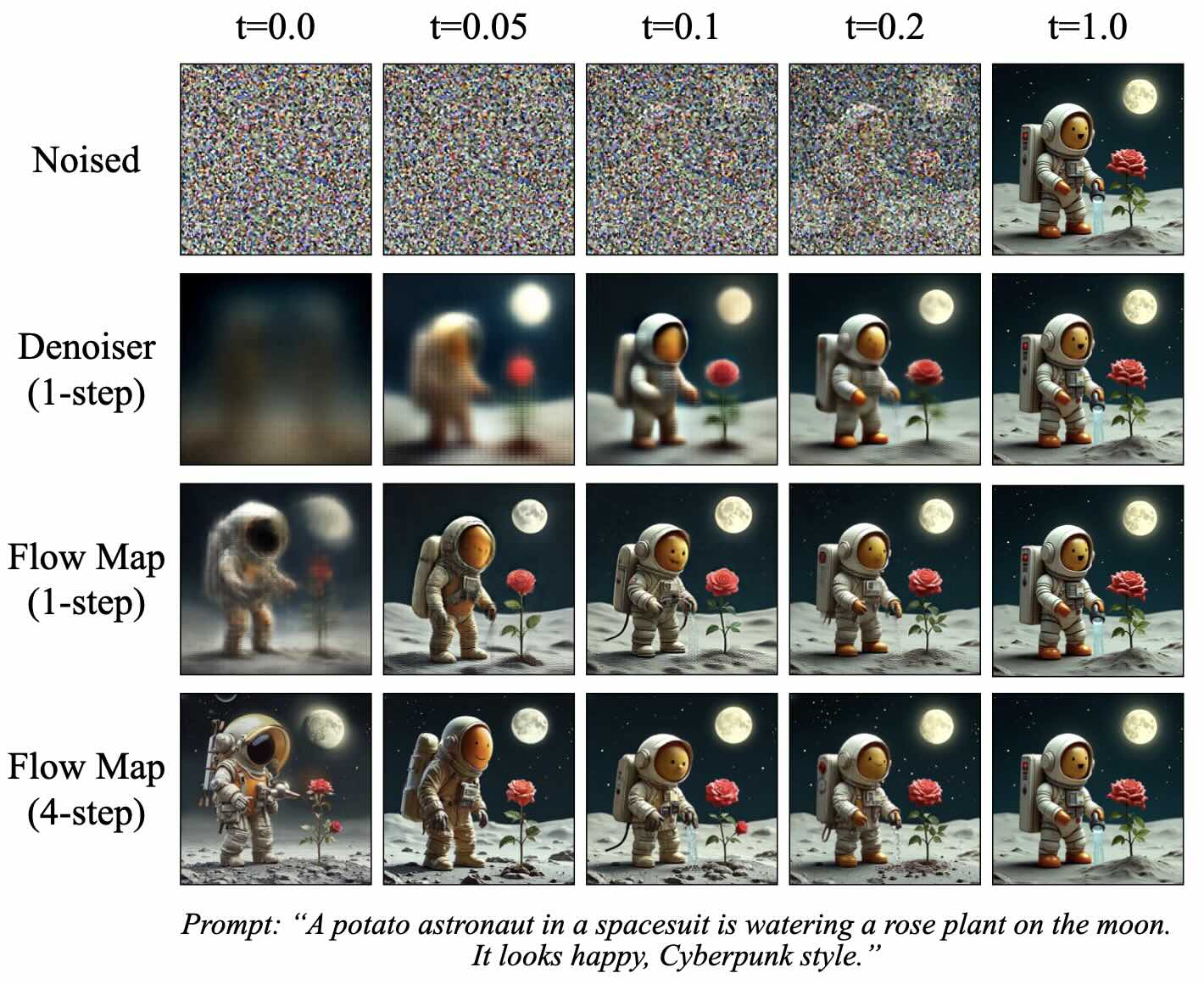}
    \vspace{-5mm}
    \caption{Comparing look-aheads across noise levels $t$: corrupted inputs, 1-step denoiser, 1-step flow map, and 4-step flow map.\looseness=-1}
    \label{fig:viz_lookaheads}
\end{wrapfigure}

While guidance-based approaches can often be made to work well in practice, most methods are somewhat \textit{ad-hoc}, and proceed by postulating a term that may drive the generative equations towards the desired goal.
To this end, a common approach is to incorporate the gradient of the reward model, which imposes a gradient ascent-like structure on the reward throughout generation.
Despite the intuitive appeal of this approach, typical rewards are defined only at the terminal point of generation -- i.e., over a clean image -- rather than over the entire generative process.
This creates a need to ``predict'' where the current trajectory will land, in principle necessitating an expensive additional differential equation solve per step of generation.
To avoid the associated computational expense of this nested solve, common practice is to employ a heuristic approximation of the terminal point, such as leveraging an approximate, often blurry, one-step denoiser estimate that can be derived from a learned score- or flow-based model. See \Cref{fig:viz_lookaheads} for visualizations.\looseness=-1

In this paper, we revisit reward guidance through flow maps, a recent approach to flow-based generative modeling that directly learns the solution operator of a probability flow ODE rather than its drift~\citep{boffi_flow_2024,boffi2025build,Sabour2025AlignYF, geng2025mean}.
By leveraging a simple identity of the flow map, we show that an implicit flow can be used to define a reward-guided generative process as in the case of standard flow-based models.
With access to the flow map in addition to the implicit flow, we can accurately predict the terminal point of a trajectory using a few function evaluations, vastly improving the approximate prediction relative to denoiser-based techniques -- see again \Cref{fig:viz_lookaheads} -- and leading to significantly improved optimization of the reward. 
In addition, we highlight how to incorporate time-dependent weights throughout the generative process to account for the gradient ascent's failure to equilibrate on the timescale of generation, leading to several new and effective ways to sample high-reward outputs.\looseness=-1

\textbf{Contributions.} 
(\textit{i}) We introduce Flow Map Trajectory Tilting (FMTT), a principled inference-time adaptation procedure for flow maps that uses their efficient yet accurate look-ahead capabilities to incorporate complex rewards into diffusion guidance and search algorithms. 
(\textit{ii}) Using properties of the flow map, we derive a simple expression for the importance weights in our Jarzynski/SMC scheme, and prove that this scheme is optimal in a simplified Gaussian setting with linear rewards.
(\textit{iii}) We show that FMTT has better test-time scaling characteristics than standard ways of embedding rewards into diffusion sampling, and can effectively leverage natural-language rewards specified via pretrained VLMs. On these complex rewards, existing test-time search baselines often offer little to no improvement over naive best-of-$N$, while flow maps make search reliably effective, significantly improve performance, and solve cases for which prior methods fail.\looseness=-1

\section{Related Work}

\begin{figure}[t]
  \centering
  \includegraphics[width=\linewidth]{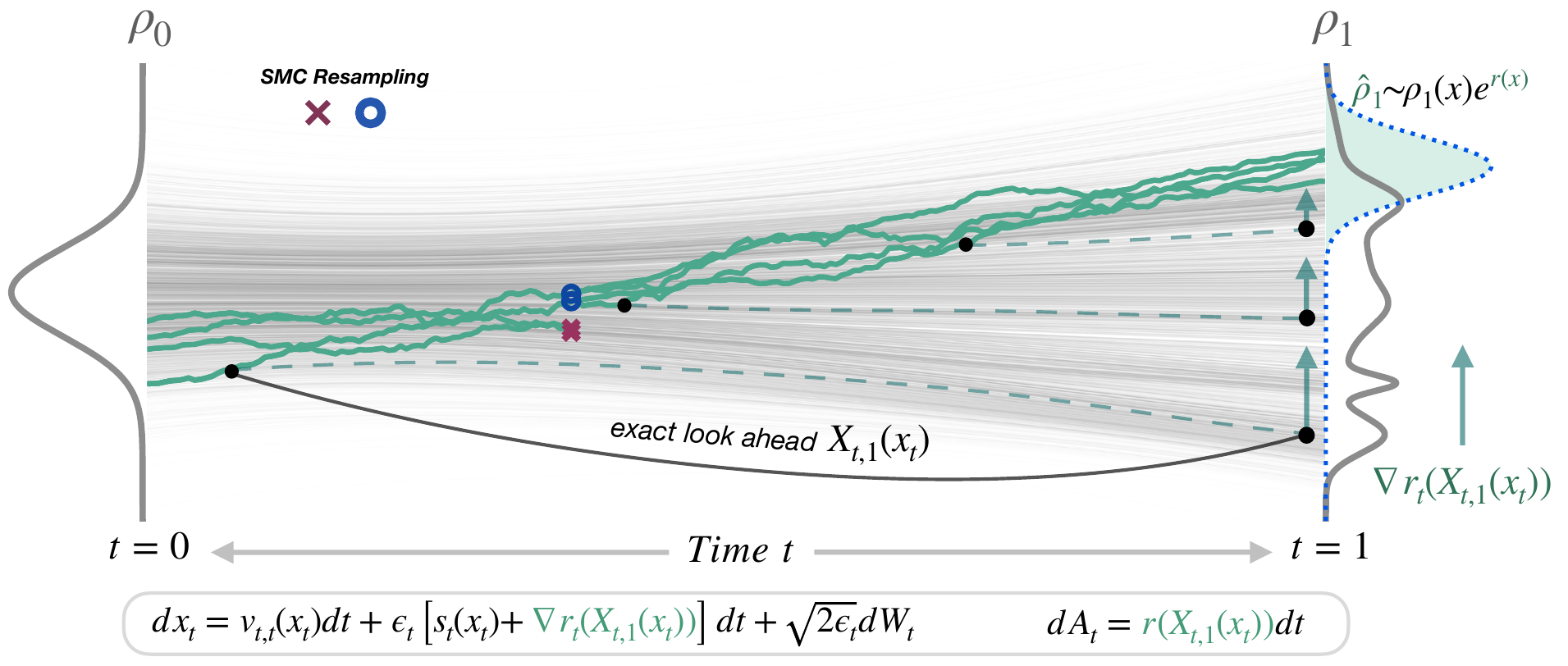}
  \caption{\textbf{Overview of Flow Map Trajectory Tilting}. Using the look-ahead map $X_{t,1}(x_t)$ in the diffusion inside the reward, reward information can be principly used through the tilted trajectories (green lines). This allows us to perform better ascent on the reward, and the importance weights $A_t$ take on a remarkably simple form that can be used for both exactly sampling $\hat \rho_t$ and search for maximizers of $\hat \rho_t$.\looseness=-1}
  \label{fig:overview}
  \vspace{-4mm}
\end{figure}

\paragraph{Flows and diffusions.}
Diffusion \citep{song2020score, ho2020denoising} and flow models \citep{lipman2022flow, albergo2022building, liu2022flow} are the backbone of efficient, state-of-the-art generative model for continuous data. They are learned by regressing the coefficients that appear in ordinary or stochastic differential equations that fulfill the transport of samples from one distribution to samples from another. Dual to the instantaneous picture of transport is the flow map \citep{song_consistency_2023, kim_consistency_2024, boffi_flow_2024, geng2025mean, Sabour2025AlignYF,boffi2025build}, in which we learn not the coefficients in a differential equation that needs to be integrated, but the arbitrary integrator itself. This enables few-step sampling. Our approach in this paper is to combine these perspectives to modify diffusions using the flow map.\looseness=-1

\vspace{-4mm}
\paragraph{Test-time scaling for diffusions.}
Test-time scaling in diffusions refers to the line of work that trades off compute at inference time to improve the performance of a model or align the generation with a user-specified reward~\citep{ma2025inference}. 
Certain works use the approximate denoiser associated with the score model to perform look-ahead on the dynamics \citep{wu2023practical, singhal2025general,zhang2025vfscale,das2025} or do not use any look-ahead at all \citep{mousavi2025flowsemidiscrete,skreta2025feynman}. However, as discussed later, there is little signal from the denoiser at early times in the generative trajectory. Other works rely on Monte Carlo search algorithms~\citep{Lee2025AdaptiveIS,ramesh2025test}, which monotonically increase the reward but reduce sample diversity. 
As we will see, many of these approaches are compatible with the flow map approach presented here.\looseness=-1

\vspace{-2mm}
\paragraph{Posterior sampling.}
Reward alignment can also be viewed through the lens of posterior sampling. Introducing a binary optimality variable $\mathcal O$ with $p(\mathcal O{=}1 \mid x_1) \propto e^{r(x_1)}$, the tilted density $\hat \rho_1$ is precisely the posterior $p(x_1 \mid \mathcal O{=}1)$, which is the perspective taken by DAS~\citep{das2025}. This makes test-time scaling closely related to solving inverse problems with diffusions~\citep{chung2023diffusion,song2023pseudoinverse,zhang2025improving,ye2026cldps,basu2026contrastive}, where the likelihood is similarly only defined on clean samples and must be approximated along the generative trajectory. DPS~\citep{chung2023diffusion}, for example, evaluates the likelihood at the one-step denoiser estimate of the clean sample, which for a reward-based likelihood recovers the denoiser look-ahead discussed in \Cref{sec:methodology} and hence faces the same issue of a weak gradient signal at high noise values. Such methods that evaluate can instead use the flow map to make that prediction; DAPS~\citep{zhang2025improving}, which obtains its endpoint from a short diffusion rollout, is a natural example. Other approaches rely on structure specific to inverse problems, such as a known measurement operator, a pseudoinverse~\citep{song2023pseudoinverse}, or a learned likelihood surrogate~\citep{ye2026cldps,basu2026contrastive}, and do not directly extend to arbitrary rewards.\looseness=-1

\vspace{-2mm}
\section{Methodology} \label{sec:methodology}
\vspace{-2mm}
We consider the task of generative modeling via continuous-time flow maps, wherein samples $x_0\in \mathbb R^d$ from a base distribution with probability density function (PDF) $\rho_0$  are mapped via a diffeomorphism to samples $x_1$ from the target PDF $\rho_1$ known through empirical data. From there, we will detail how the instantaneous dynamics of this map can be directly adapted (without retraining) to sample a \textbf{tilted distribution} favoring a reward, i.e. to sample $\hat \rho_1(x) = \rho_1 e^{r(x) + \hat F}$ where $r(x)$ is a user specified reward function  and $\hat F = - \ln \int_{\mathbb R^d} \rho_1(x) e^{r(x)} dx$ is a normalization factor.

\vspace{-2mm}
\subsection{Background on Dynamical Generative Modeling}
\vspace{-2mm}
An effective means of instantiating the transport from the base PDF $\rho_0$ to the target PDF $\rho_1$ relies on formulating it as the solution to an ordinary differential equation (ODE) \refstepcounter{equation}\label{eq:ode}\(\dot x_t = b_t(x_t), \ x_{t=0} \sim \rho_{t=0}\)~(\theequation),
where $b_t: [0,1] \times \mathbb R^d \rightarrow \mathbb R^d$ is a velocity field that governs the transport, adjusted so that the solutions to the ODE~\eqref {eq:ode} satisfy $x_{t=1}{\sim}\rho_1$. As the time-dependent PDF $\rho_t(x)$ of the solutions to \eqref{eq:ode} at time $t$ satisfies the  continuity equation \refstepcounter{equation}\label{eq:ce}\(\partial_t \rho_t = -\nabla \cdot (b_t \rho_t), \ \rho_{t=0} = \rho_0\)~(\theequation),\looseness=-1
this requirement on $b_t$ implies that the solution to~\eqref{eq:ce} is such that $\rho_{t=1}=\rho_1$.

Associated with these dynamics is the two-time \textbf{flow map} $X_{s,t}: [0,1]^2 \times \mathbb R^d \rightarrow \mathbb R^d$, which satisfies \refstepcounter{equation}\label{eq:map}\(X_{s,t}(x_s) = x_t, \ \forall s,t \in [0,1]\)~(\theequation).
That is, the map jumps along solutions of \eqref{eq:ode} from time $s$ to time $t$. Notably, if $s=0$ and $t=1$, we could produce a sample under $\rho_1$ in a single step, though we have the freedom to use more if we so choose. This property, and the relation between the flow map and $b_t$ will be exploited below to devise a principled adaptation procedure for $X_{s,t}$. Importantly, the  flow map satisfies the Eulerian equation \refstepcounter{equation}\label{eq:euler}\(\partial_s X_{s,t}(x) + b_s(x) \cdot \nabla X_{s,t}(x)=0\)~(\theequation),\looseness=-1
which will play a role in simplifying our analysis later.
Equation~\eqref{eq:euler} can be obtained by taking the total derivative of~\eqref{eq:map} with respect to $s$, using the ODE~\eqref{eq:ode}, and evaluating the result at $x_s=x$.\looseness=-1

\vspace{-2mm}
\paragraph{Stochastic interpolants.}
\vspace{-2mm}
One way to instantiate the generative models above is to construct a PDF $\rho_t$ that connects $\rho_0$ to $\rho_1$ and then learn the associated velocity field $b_t$ that gives rise to this evolution. A common strategy to construct such a path and regress $b_t$ is that of stochastic interpolants \citep{albergo2022building, albergo2023stochastic, lipman2022flow, liu2022flow}, in which $\rho_t$ is defined as the law of the stochastic process $I_t(x_0, x_1) = \alpha_t x_0 + \beta_t x_1$ with $(x_0, x_1) \sim \rho(x_0, x_1)$, where $\rho(x_0, x_1)$ is some coupling from which $x_0, x_1$ are drawn that marginalizes onto $\rho_0$, $\rho_1$,
and $\alpha_t, \beta_t$ are scalar coefficients that satisfy $\alpha_0 = \beta_1 = 1$ and $\alpha_1 = \beta_0 = 0$. 
A common choice is to use $\alpha_{t} = 1-t$ and $\beta_t = t$, which we will use throughout this work for simplicity. 
Importantly, using these coefficients, the law of this process satisfies \eqref{eq:ce} with velocity field \refstepcounter{equation}\label{eq:si:b}\(b_t(x) = \mathbb E[\dot I_t | I_t = x] = \mathbb E[x_1 | I_t] - \mathbb E[x_0 | I_t]\)~(\theequation), where we used $\dot I_t = x_1-x_0$ and $\mathbb E[\cdot | I_t=x]$ denotes expectation over $\rho(x_0, x_1)$ conditional on $I_t = x$. By Stein's identity, the score is given by  $s_t(x) = \nabla \log \rho_t(x) = -\tfrac{1}{1-t} \mathbb E[x_0 | I_t=x]$, and using  $x = \mathbb E[I_t | I_t=x]= (1-t) \mathbb E[x_0|I_t] + t \mathbb E[x_1 | I_t]$, it can be expressed in terms of $b_t$ as \refstepcounter{equation}\label{eq:score}\(s_t(x) = (t b_t(x) - x)(1-t)^{-1}\)~(\theequation).\looseness=-1

The velocity field $b_t(x)$ is also the minimizer of a simple quadratic objective \citep{lipman2022flow, albergo2022building}
which, once learned, can be translated into a function for the score via \eqref{eq:score}. 
Using the score, the deterministic ODE can be converted to a stochastic dynamics
\looseness=-1
\refstepcounter{equation}\label{eq:sde}\(dx_t = [b_t(x_t)+\epsilon_t s_t(x_t)]dt + \sqrt{2\epsilon_t}dW_t\)~(\theequation),
where $\epsilon_t\ge0$ is an arbitrarily tunable diffusion coefficient and $dW_t$ is an incremental Brownian motion~\citep{albergo2023stochastic}. 
The solutions to \eqref{eq:sde} sample the same PDF $\rho_t$ as \eqref{eq:ode}, as can be seen from the fact that the PDF of \eqref{eq:sde} satisfies the Fokker-Planck equation \refstepcounter{equation}\label{eq:fpe}\(\partial_t \rho_t = - \nabla \cdot (b_t \rho_t) + \epsilon_t\nabla \cdot [-s_t \rho_t + \nabla \rho_t]\)~(\theequation), which reduces to \eqref{eq:ce} since $s_t \rho_t = \nabla \rho_t$.  The velocity field given in \eqref{eq:si:b} is related to the two-time flow map $X_{s,t}$ via the tangent identity~\citep{kim_consistency_2024,boffi2025build} \refstepcounter{equation}\label{eq:tangent}\(\lim_{s\rightarrow t} \partial_t X_{s,t}(x) = b_t(x)\)~(\theequation),
which says that for infinitesimally small steps, the variation of the flow map output in time is characterized by the velocity. 
To automatically enforce the boundary condition~$X_{s,s}(x) = x$, we can express the flow map as \refstepcounter{equation}\label{eq:map:resid}\(X_{s,t}(x) = x + (t-s)v_{s,t}(x)\)~(\theequation),
where $v_{s,t}(x): [0,1]^2 \times \mathbb R^d \rightarrow \mathbb R^d$ is a function of $s$, $t$, and $x$ defined through this relation. Importantly, \textit{the velocity field is accessible directly from the flow map by using \eqref{eq:tangent} on \eqref{eq:map:resid}:}
\refstepcounter{equation}\label{eq:map:b}\(v_{t,t}(x) = b_t(x)\)~(\theequation).
As such, training a flow map model instead of just a velocity gives access to both the drifts in \eqref{eq:ode} and \eqref{eq:sde} as well as the any-step model, i.e. the ability to \textit{look ahead} our trajectory.\looseness=-1

\begin{figure}[t]
  \centering
  \vspace{-3mm}
  \includegraphics[width=\linewidth]{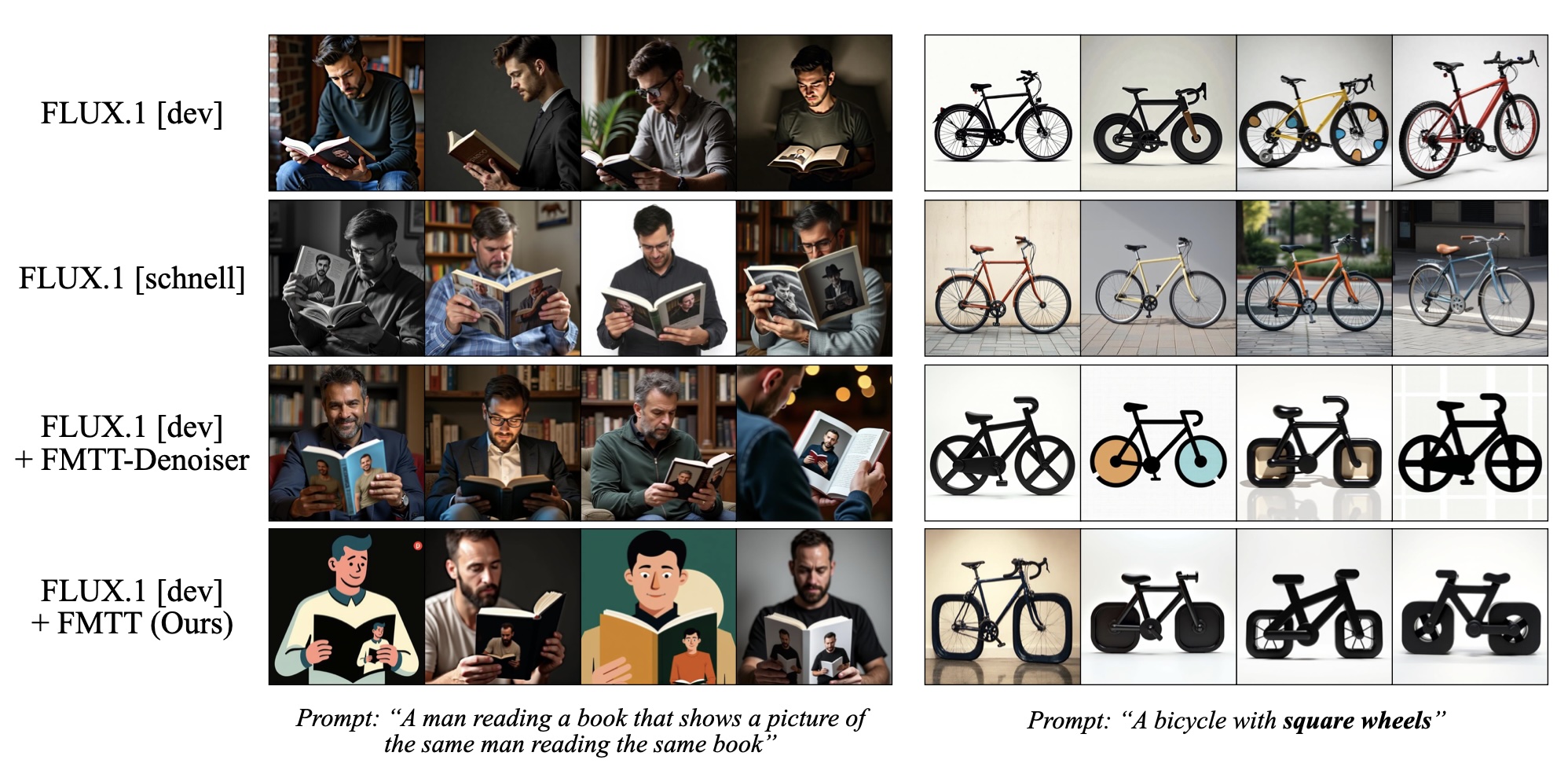}
  \caption{Qualitative results using VLM-based rewards. Prompts where the base model fails to generate aligned outputs are corrected by FMTT. Flow map look-ahead produces the most reliable improvements, whereas denoiser look-ahead often remains insufficient to achieve precise prompt following. For additional qualitative results, see \Cref{app:additional_vlm_examples}.\looseness=-1}
  \label{fig:vlm_single_image}
  \vspace{-5mm}
\end{figure}

\vspace{-1mm}
\subsection{Fixing Inference-time Adaptation of Diffusions with Flow Maps} \label{subsec:fixing_inference_time}
\vspace{-1mm}

A contemporary question is how best to adapt the SDE \eqref{eq:sde} to tilt toward samples that score highly against a time-dependent reward $r_t(x)$ satisfying $r_0{=}0$ and $r_{t=1}{=}r$ so that the time-dependent tilted PDF $\hat \rho_t{=}\rho_t e^{r_t(x) + \hat F_t}$ satisfies $\hat \rho_{t=0}{=}\rho_0$ and $\hat \rho_{t=1}{=}\hat \rho_1{=}\rho_1 e^{r+\hat F}$. One may want to \textit{sample exactly} under the tilted PDF $\hat \rho_t{=}\rho_t e^{r_t(x) + F_t}$ for scientific applications, or one may want to \textit{find local maximizers} of $\hat \rho_t$, such as in image generation to ensure user prompts align with image content. \looseness=-1

\vspace{-3mm}
\paragraph{Tilting the diffusion.}  
A natural way to modify the SDE~\eqref{eq:sde} is to add the gradient of the time-dependent reward $r_t$ to the score, i.e. use
\begin{equation}
\label{eq:sde:r}
\begin{aligned}
d\tilde x_t = [b_t(\tilde x_t) + \epsilon_t s_t(\tilde x_t) + \epsilon_t \nabla r_t(\tilde x_t) ] dt + \sqrt{2 \epsilon_t} dW_t, \quad \tilde x_0 \sim \rho_0,
\end{aligned}
\end{equation}
assuming $r_{0}(x) = 0$ to avoid changing the prior $\rho_0$.
However, in practice this faces an issue:\looseness=-1

\begin{pinkbox}
\textit{A meaningful~$r_t(x)$ is  not readily available, as user-specified rewards are usually learned only on the data-distribution, i.e. at time $t=1$.}
\end{pinkbox}
One could think of several solutions to this problem: 
\textbf{Na\"ive (no) look-ahead:} This amounts to using e.g. $r_t(x){=}t \cdot  r(x)$. Unfortunately, the gradient dynamics from $t \nabla r(x)$ provides no clear signal at small times when $\tilde x_t$ is still far from the region where the reward~$r(x)$ is meaningful. \textbf{Denoiser look-ahead:} A common workaround for the fact that the reward has no signal for most of the trajectory is to use the \textit{denoiser} $\mathsf D_t(x){=}\mathbb E[x_1 | I_t {=}x]$ to estimate where the sample would have gone. That is, instead of $r_t(x){=}t \cdot  r(x)$, we could instead use $r_t(x){=}t \cdot  r( \mathsf D _t(x))$. This strategy is tractable because the denoiser is readily available from the score. However, this still does not provide useful information early on in the dynamics, as the denoiser is only effective at producing samples close to the data distribution later in the evolution (\Cref{fig:viz_lookaheads}). \textbf{Flow map look-ahead:} 
Intuitively, the above dynamics are  better if one works instead with the flow map defined above. Because the flow map allows us to look ahead at any point on the trajectory, e.g. by taking $\tilde x_t$ at time $t$ and computing $X_{t,1}(\tilde x_t)$, and because the velocity field associated to the flow map is accessible via $\lim_{s\rightarrow t} \partial_t X_{s,t}(x) = v_{t,t}(x)$ (from which we can recover the score function $s_t(x)$), we can instead sample with the following SDE:\looseness=-1
\begin{propbox}
\begin{equation}
\label{eq:position_SDE}
\begin{aligned}
d \tilde x_t
&= v_{t,t}(\tilde x_t)\,dt
 + \epsilon_t \Big[s_t(\tilde x_t)
   + t \nabla_{\tilde x_t} r\!\big(X_{t,1}(\tilde x_t)\big)\Big]\,dt + \sqrt{2\epsilon_t}\, dW_t, \quad \tilde x_0 \sim \rho_0.
\end{aligned}
\end{equation}
\end{propbox}
where $r_t(x)=t \cdot r(X_{t,1}(x))$ makes use of the \textit{exact look-ahead} to properly evaluate the reward for any $t$. 
The above could be interpreted as a mixture of denoising and ascent on the reward.\looseness=-1

\subsection{Correcting the Dynamics for Unbiased Sampling} \label{sec:correcting}
\vspace{-2mm}

While using the flow map composed with the reward makes possible the precise use of the reward for all times in the diffusion trajectory, in applications where it is important to \textit{exactly} sample the tilted distribution $\hat \rho_t$, the dynamics in \eqref{eq:position_SDE} are not sufficient
for this.
This leads to the second issue, for any version of $r_t(x)$, with or without flow map:\looseness=-1
\begin{pinkbox}
\textit{The PDF associated with \eqref{eq:sde:r} is not the tilted density $\hat \rho_t$.}
\end{pinkbox}
To see why this is true, note that we can explicitly compare the PDF $\tilde \rho_t$ of $\tilde x_t$ to that of $\hat \rho_t$. Indeed, the PDF $\tilde \rho_t(x)$ of $\tilde x_t$ satisfies:
\begin{equation}
    \partial_t \tilde \rho_t + \nabla \cdot (b_t \tilde \rho_t) = \epsilon_t \nabla \cdot ([-s_t - \nabla r_t ]\tilde \rho_t + \nabla \tilde \rho_t),
\end{equation}
and we can show explicitly that $\hat \rho_t$ satisfies a different, imbalanced equation which can be obtained by expanding $\partial_t \hat \rho_t + \nabla \cdot (b_t \hat \rho_t) $:
\begin{equation}
\label{eq:stePrhor}
\begin{aligned}
\partial_t \hat \rho_t + \nabla \cdot (b_t \hat \rho_t) &= \partial_t ( e^{r_t + \hat F_t} \rho_t) + \nabla \cdot (b_t e^{r_t + \hat F_t} \rho_t) \\
&= (\partial_t r_t + \partial_t \hat F_t) \hat \rho_t  + b_t \cdot \nabla r_t \hat\rho_t, \\
\end{aligned} 
\end{equation}
where we used the FPE~\eqref{eq:fpe} with $\epsilon_t=0$ to get the last equality.  Since $\nabla \hat \rho_t = (s_t + \nabla r_t) \hat \rho_t$ we can add the diffusion term $\epsilon_t \nabla\cdot (-(s_t + \nabla r_t) \hat \rho_t +\nabla \hat \rho_t) = 0 $ to~\eqref{eq:stePrhor}  to arrive at
\begin{equation}
\label{eq:dtrho:emp}
\begin{aligned} 
   \partial_t \hat \rho_t +  \nabla \cdot (b_t\hat \rho_t) = &\epsilon_t \nabla \cdot((-s_t - \nabla r_t) \hat \rho_t + \nabla\hat \rho_t ) + (b_t \cdot \nabla r_t + \partial_t r_t + \partial_t \hat F_t) \hat \rho_t.
\end{aligned}
\end{equation}
As we see, the  extra term  $(b_t \cdot \nabla r_t + \partial_t r_t + \partial_t \hat F_t) \hat\rho_t$ on the RHS of this equation differentiates it from being the law of \eqref{eq:sde:r}. Nonetheless, we can account for this term with weights $A_t$ emerging as the solution of a different differential equation coming from an adaptation of the Jarzynski equality \citep{jarzynski1997, vaikuntanathan2008}:\looseness=-1

\begin{propbox}
\begin{restatable}[Jarzynski's estimator]{proposition}{jar}
\label{prop:jar}
    Assume that $r_0=0$ so that $\rho^r_0 = \rho_0$.
	Let $\tilde x_t$ solve the SDE~\eqref{eq:sde:r} with $\tilde x_0 \sim \rho^r_0$ and define
	\begin{alignat}{2}
  \label{eq:sde:A}
		A_t = \int_0^t \left( b_s(\tilde{x}_s) \cdot \nabla r_s(\tilde{x}_s)  + \partial_s r_s(\tilde{x}_s)\right) ds,
	\end{alignat}
	Then for all $t\in[0,1]$ and any test function $h:\mathbb R^d\to\mathbb R$, we have 
	\begin{equation}
 \label{eq:expect:map}
		\int_{\mathbb R^d} h(x) \hat \rho_t(x) dx = \frac{\mathbb E[ e^{A_t} h(\tilde x_t)]}{\mathbb E[e^{A_t}]},
	\end{equation}
	where the expectations at the right-hand side are taken over the law of $\tilde x= (\tilde x_t)_{t\in[0,T]}$.
\end{restatable}
\end{propbox}
The proof of this statement in \Cref{app:proof:generic} relies on manipulating the augmented FPE of the joint PDF $f_t(x,a)$ of $(\tilde x_t,A_t)$. This relation ensures that the lag associated to naively using the gradient of the reward in the diffusion can be compensated for by reweighting the trajectories, and, in addition, these weights can be used to perform resampling of the trajectories as is done in Sequential Monte Carlo (SMC) and birth/death processes, as is depicted in Figure \ref{fig:overview}. Here, as the trajectories walk out, the walkers can be \textit{resampled} using the importance weights, removing some and duplicating others.

\vspace{-3mm}
\paragraph{Simplicity of importance weights with the flow map.} Interestingly, the importance weights in \eqref{eq:sde:A} take on a remarkably simple form when using a flow map look-ahead, as stated in the following:\looseness=-1
\vspace{-0mm}
\begin{propbox}
\begin{proposition}[Unbiased Flow Map Trajectory Tilting]
\label{prop:map:tilt}
    Using the same notations as in Proposition~\ref{prop:jar}, if $r_t(x) = t \cdot  r(X_{t,1}(x))$, then the importance weights defined in \eqref{eq:sde:A} reduce to
    \begin{align} 
    \label{eq:position_weights}
        A_t &= \int_0^t r(X_{s,1}(\tilde x_s))ds.
    \end{align}
\end{proposition}
\end{propbox}
\vspace{-0mm}
This result is proven in \Cref{app:proof:map:tilt}, and relies on a simple modification of the proof of \Cref{prop:jar} and the Eulerian equation \eqref{eq:euler}. Thanks to the flow map, the complicated derivatives appearing in \eqref{eq:sde:A} reduce to simply compounding the reward over the look-ahead trajectory.

\vspace{-3mm}
\paragraph{Reward-modified drift.} A variant of \Cref{prop:jar}
makes use of not only augmenting the score with the gradient of the reward with the look-ahead, but also the drift itself. Because $v_{t,t}$ is the velocity field of a stochastic interpolant and is related to the score via \eqref{eq:score}, we can replace the SDE given in \eqref{eq:position_SDE} with
\begin{equation}
\label{eq:sde:mod}
\begin{aligned}
d\tilde x_t &= \left[ v_{t,t}(\tilde x_t) \! + \! \chi_t \nabla r_t(\tilde x_t)\right] dt + \epsilon_t \left[s_t(\tilde x_t) \! + \! \nabla r_t(\tilde x_t)\right] dt + \sqrt{2\epsilon_t} dW_t, \quad \tilde x_0 \sim \rho_0,
\end{aligned}
\end{equation}
where $(\chi_t)_{t \in [0,1]}$ is arbitrary, 
and the log-weight ODE with 
\begin{equation}
\label{eq:a_t:mod}
\begin{aligned}
\frac{d\tilde{A}_t}{dt} &= b_t(\tilde{x}_t) \cdot \nabla r_t(\tilde{x}_t) \! + \! \partial_t r_t(\tilde{x}_t) + \chi_t \big( \| \nabla r_t(\tilde{x}_t)\|^2 \! + \! \Delta r_t(\tilde{x}_t) \! + \! \langle \nabla r_t(\tilde{x}_t), s_t(\tilde{x}_t) \rangle \big),
\end{aligned}
\end{equation}
where 
$\tilde A_0 = 0$. It is proven in Appendix \ref{app:proof:map:tilt:reward} that these equations also provide an unbiased sampler of the tilted distribution by ensuring \eqref{eq:expect:map} holds for any choice of $(\chi_t)_{t \in [0,1]}$, and three choices besides the default $\chi_t \equiv 0$ are studied (\Cref{app:additional_MNIST_experiments}). This allows further augmenting the sampling process toward the tilt. We use $\chi_t = \lambda \cdot (1-t)/(t)$ in our experiments, where $\lambda$ is a hyperparameter controlling the strength of the additional tilt. We found $\lambda=0.1$ to work well in most cases.\looseness=-1

\vspace{-3mm}
\paragraph{From sampling to search: making the most of rewards in practice.} Notably, the importance weights defined in either \Cref{eq:position_weights,eq:a_t:mod} do not need to be used to perform exact sampling. They can also be used to perform various greedy search algorithms that \textbf{search} for samples with high reward. That is, we are free to use top-n sampling approaches in place of the resampling one would usually do in SMC. Unlike \citet{ma2025inference}, this use of top-n still makes use of the gradient of the reward along the trajectory, while also making use of the better signal thanks to the flow map. A full pseudocode can be seen in \Cref{alg1:inference_time_adaptation}.\looseness=-1

\subsection{Optimality of Flow Map Look-Ahead}
\label{sec:flowmap_optimality}
\vspace{-2mm}

\begin{figure}[t]
    \centering
    \vspace{-3mm}
    \includegraphics[width=\linewidth]{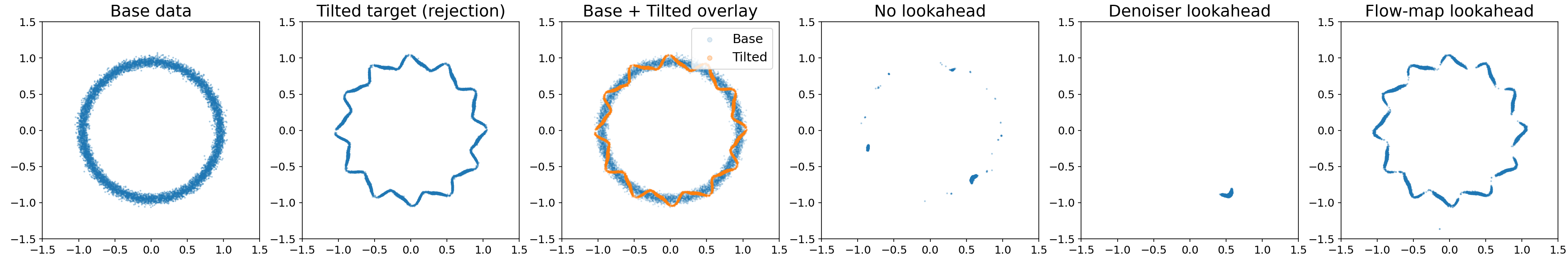}
    \caption{
    Tilting a 2D toy distribution. We train a flow model on the base ring and define a reward that tilts it toward a zigzag target. With 500-step SMC, no look-ahead and denoiser look-ahead fail to recover the target, while flow map look-ahead captures it effectively. See \Cref{app:toy_2d_experiments} for details.\looseness=-1}
    \label{fig:toy_2d_comparison}
    \vspace{-5mm}
\end{figure}

In \Cref{fig:toy_2d_comparison}, we visualize the SMC output using different look-aheads on a 2D toy dataset. We train a flow matching model on a ring-shaped distribution and define a reward that tilts this base distribution toward a zigzag-shaped target. Running the SMC algorithm in \Cref{eq:position_SDE,eq:sde:A} with 500 steps and 10000 particles, we find that both naive and denoiser look-aheads require over 150 resampling steps, which degenerates the particle system and produces collapsed blob-like outputs. In contrast, flow map look-ahead uses only 6 resampling steps and accurately captures the tilted target. This difference stems from the large variance of the SMC importance weights when using the unsuccessful look-aheads. For more details, please see \Cref{app:toy_2d_experiments}.\looseness=-1

To better understand this effect, we analyze a simple setting where the data distribution is an isotropic Gaussian with standard deviation $c$ and the reward is linear. The following theorem shows that the flow map is the optimal linear look-ahead that minimizes the variance of the incremental log-weights $A_t$.\looseness=-1
\vspace{-1mm}
\begin{propbox}
\begin{theorem}[Proof in \Cref{app:toy_theory_flowmap_optimal}]
\label{thm:flow-map-linear-gaussian-optimal}
Let $\rho_{data}(x) = \mathcal{N}(0, c^2 I)$ be the data distribution, and let $r(x) = u^\top x$ be a linear reward for some $u$. 
For the default SMC correction specified by \Cref{eq:position_SDE,eq:sde:A}, the flow map is the optimal linear look-ahead that minimizes the variance of the incremental log-weights $A_t$. 
\end{theorem}
\end{propbox}
\vspace{-1mm}

Although the theorem is proved in a simplified, analytically tractable setting, together with the 2D results it suggests that flow map look-ahead yields more stable SMC behavior than prior look-aheads. Directly measuring this stability is difficult: while SMC gives an unbiased estimate $\hat{Z}_{\mathrm{SMC}}$ of $\mathbb E[e^{A_t}]$, its variance can be exponentially large, hard to estimate, and sensitive to choices such as particle count, resampling, and schedule. To address this, in \Cref{app:analyzing} we relate $\mathrm{Var}[\hat{Z}_{\mathrm{SMC}}]$ to the \textit{thermodynamic length}, a simpler proxy computable from the $A_t$ updates. In \Cref{app:additional_MNIST_experiments}, class-conditional MNIST experiments show that flow map look-ahead again gives the most stable SMC behavior, achieving the lowest thermodynamic length. These theoretical analyses motivate our novel flow map look-ahead framework as a principled and broadly relevant technique applicable also in other diffusion model search and reward guidance methods~\citep{das2025,singhal2025general,skreta2025feynman}. This underscores the general impact of FMTT.

\vspace{-2mm}
\section{Experiments}
\vspace{-2mm}

Next, we evaluate our approach on text-to-image generation using a 4-step distilled flow map following Align Your Flow~\citep{Sabour2025AlignYF}, which is trained by distilling the open-source FLUX.1 [dev] model~\citep{flux2024}. The flow map is trained in roughly 4 hours on 8 NVIDIA A100 GPUs. For these experiments,  $\mathrm{Sampling} = \mathrm{False}, \mathrm{Searching} = \mathrm{True}$ is used in \Cref{alg1:inference_time_adaptation}. 
We consider three categories of reward functions:  
1) \textit{VLM-based rewards} defined through natural language queries, 
2) \textit{Geometric rewards} enforcing structural constraints such as symmetry or layout control,  
3) \textit{Human preference rewards} capturing visual quality and text alignment.
Note that for these experiments, unlike the experiments discussed in \Cref{sec:flowmap_optimality} which perform exact sampling, we use the search version of FMTT.\looseness=-1

We compare against several \textit{gradient-free} and \textit{gradient-based} baselines. Gradient-free approaches such as Best-of-$N$~\citep{chatterjee2018sample}, Multi-Best-of-$N$~\citep{Lee2025AdaptiveIS}, and beam search~\citep{fernandes2025latent} rely on sampling and selection, and remain confined to the base model’s distribution. Gradient-based methods use reward gradients, but differ in how they apply them: ReNO~\citep{eyring2024reno} performs gradient ascent in the initial noise latent space, keeping samples tied to the base distribution, whereas our FMTT algorithm (and its ablations) and DAS~\citep{das2025} use the gradient to modify the generative  process itself, enabling exploration beyond the model’s support and generation of out-of-distribution (OOD) samples. Notably, the gradient of the reward used in FMTT efficiently gives meaningful signal for the whole trajectory thanks to the flow map, enabling OOD sample generation for highly nuanced rewards.\looseness=-1

\begin{table*}[b]
\caption{Summary of key quantitative results on UniGenBench++. Entries show the percentage of images judged by the evaluation model to satisfy each criterion. We use 4 particles for the base model and flowmap, 256 particles for best-of-$N$ and FMTT, and 64 particles for DAS. See the extended \Cref{tab:unigenbench} for the complete results.}
\vspace{-2mm}
\centering
\label{tab:unigenbench_mini}
\resizebox{0.99\linewidth}{!}{
\begin{tabular}{lcccccccccccc}
\toprule
\textbf{Method} & \textbf{Mean} & \textbf{Style} & \textbf{World Knowledge} & \textbf{Attribute} & \textbf{Action} & \textbf{Relationship} & \textbf{Compound} & \textbf{Grammar} & \textbf{Logical Reasoning} & \textbf{Entity Layout} & \textbf{Text Generation} & \textbf{NFE}   \\
\midrule
FLUX.1 [dev] & 61.78 & 85.20 & 86.23 & 64.53 & 60.46 & 63.07 & 41.24  & 60.96  & 24.08  & 67.16  & 30.17  & 180 \\
Flow Map & 58.16 & 79.70 & 80.22 & 59.51 & 57.22 & 62.44 & 42.01  & 57.62  & 25.92  & 70.90  & 3.74  & 16 \\
\midrule
FLUX.1 [dev] + Best-of-$N$ & 73.75   & 91.10 & \textbf{92.88} & 76.39 & 71.58 & 75.51 & 61.86 & 69.52 & 45.18 & 80.97 & 44.83 & 11520 \\
\midrule
FMTT - 1-step denoiser look-ahead & 72.90 & 91.00 & 90.51 & 73.93 & 70.72 & 78.30 & 60.31 & 69.65 & 42.89 & 81.53 & 39.94 & 11520  \\
FMTT - 1-step flow map look-ahead & 74.86 & \textbf{91.90} & 92.72 & 77.78 & 71.86 & 77.41 & 62.50 & 72.59 & 46.33 & \textbf{81.72} & 46.55 & 3840  \\
FMTT - 2-step flow map look-ahead & \textbf{75.55} & \textbf{91.90} & \textbf{92.88} & \textbf{78.21} & \textbf{72.34} & \textbf{79.19} & \textbf{65.08} & \textbf{72.73} & \textbf{46.79} & 81.53 & \textbf{47.70} & 6720  \\
\midrule
DAS - 1-step denoiser look-ahead  & 66.42 & 86.90 & 87.66 & 68.38 & 64.35 & 69.29 & 52.32 & 63.90 & 33.72 & 72.01 & 32.76 & 2880  \\
DAS - 1-step flow map look-ahead & 68.14 & 88.50 & 89.72 & 70.09 & 66.25 & 70.05 & 54.16 & 63.77 & 33.26 & 73.51 & 34.77 & 1920  \\
\bottomrule
\end{tabular}
}
\vspace{-3mm}
\end{table*}

\begin{table*}[b]
\caption{Quantitative GenEval results. NFE is total number of function evaluations during generation.}
\centering
\label{tab:geneval}
\resizebox{0.99\linewidth}{!}{
\begin{tabular}{lcccccccc}
\toprule
\textbf{Method} & \textbf{Mean $\uparrow$} & \textbf{Single Obj. $\uparrow$} & \textbf{Two Obj. $\uparrow$} & \textbf{Counting $\uparrow$} & \textbf{Colors $\uparrow$} & \textbf{Position $\uparrow$} & \textbf{Attr. Binding $\uparrow$} & \textbf{NFE}  \\
\midrule
\multicolumn{9}{l}
{\textbf{Diffusions + Flow Maps}} \\
\midrule
FLUX.1 [dev] & 0.65 & 0.99 &	0.78 &	0.70 & 0.78 & 0.18 & 0.45 & 180 \\
Flow Map  & 0.62 & 0.99 & 0.72 & 0.63 & 0.80 & 0.19 & 0.39 & 16  \\
\midrule{\textbf{Gradient-Free Search}} \\
\midrule
FLUX.1 [dev] + Best-of-$N$ & 0.75 & 0.99 & 0.94 & 0.83 & 0.86 & 0.26 & 0.57 & 1440 \\
Flow Map + Best-of-$N$ & 0.73 & 1.00 & 0.88 & 0.82 & 0.85 & 0.25 & 0.59 & 128 \\
Flow Map + Multi-best-of-$N$ & 0.76 & 1.00 & 0.95 & 0.84 & 0.85 & 0.26 & 0.69 & 1280 \\
Flow Map + Beam Search  & 0.75 & 1.00 & 0.92 & 0.86 & 0.85 & 0.29 & 0.58 & 1200 \\
\midrule
\multicolumn{9}{l}{\textbf{Gradient-Based Search}} \\
\midrule
Flow Map + ReNO & 0.71 & 0.98 & 0.89 & 0.79 & 0.89 & 0.20 & 0.57 & 1280   \\
FMTT (Ours) & \textbf{0.79} & \textbf{1.0} & \textbf{0.97} & \textbf{0.90} & \textbf{0.91} & \textbf{0.30} & \textbf{0.64} & 1400 \\
FMTT - 1-step denoiser look-ahead & 0.75 & 0.99 & 0.90 & 0.87 & 0.87 & 0.26 & 0.59 & 350  \\
FMTT - 4-step diffusion look-ahead & 0.75 & 0.99 & 0.93 & 0.86 & 0.89 & 0.27 & 0.57 & 1400 \\
\bottomrule
\end{tabular}
}
\end{table*}

\vspace{-2mm}
\subsection{VLMs as a Judge}
\vspace{-2mm}
In this section, we explore using pretrained vision language models (VLMs) to judge our images. The setup is as follows: we provide the generated image along with a binary yes/no question, and define the reward to be $\mathrm{sigmoid}(\text{logits}[\text{Yes}] - \text{logits}[\text{No}])$. This formulation allows rewards to be expressed entirely in natural language. Additionally, since some VLMs accept multiple image inputs, we can also define rewards that depend on comparisons between the generated image and additional context images. In our experiments, we use Skywork-VL Reward~\citep{wang2025skyworkVLreward} for single-image rewards and Qwen2.5-VL-7B-Instruct~\citep{bai2025qwen25VL} for multi-image cases (see \Cref{app:additional_vlm_examples} for details and examples).\looseness=-1

\vspace{-1mm}
To quantitatively benchmark FMTT, we evaluate our method on the UniGenBench++~\citep{UniGenBench++} benchmark using the Skywork-VL reward model. This benchmark contains 600 short English prompts, each requiring 4 generated images. The resulting images are scored by the UniGenBench evaluation model, a fine-tuned variant of Qwen2.5-VL-72B~\citep{bai2025qwen25VL}, across multiple dimensions including entity layout, text rendering, world knowledge, and more. We ask the following question of the Skywork-VL reward VLM: \emph{`Is \{PROMPT\} a correct caption for the image? Please answer no if the image is not in high definition (i.e., clear, sharp, not pixelated, and not blurry).''} 

\begin{figure}[t]
    \centering
    \vspace{-3mm}
    \begin{minipage}[c]{0.54\textwidth}
        \centering
        \includegraphics[width=\linewidth]{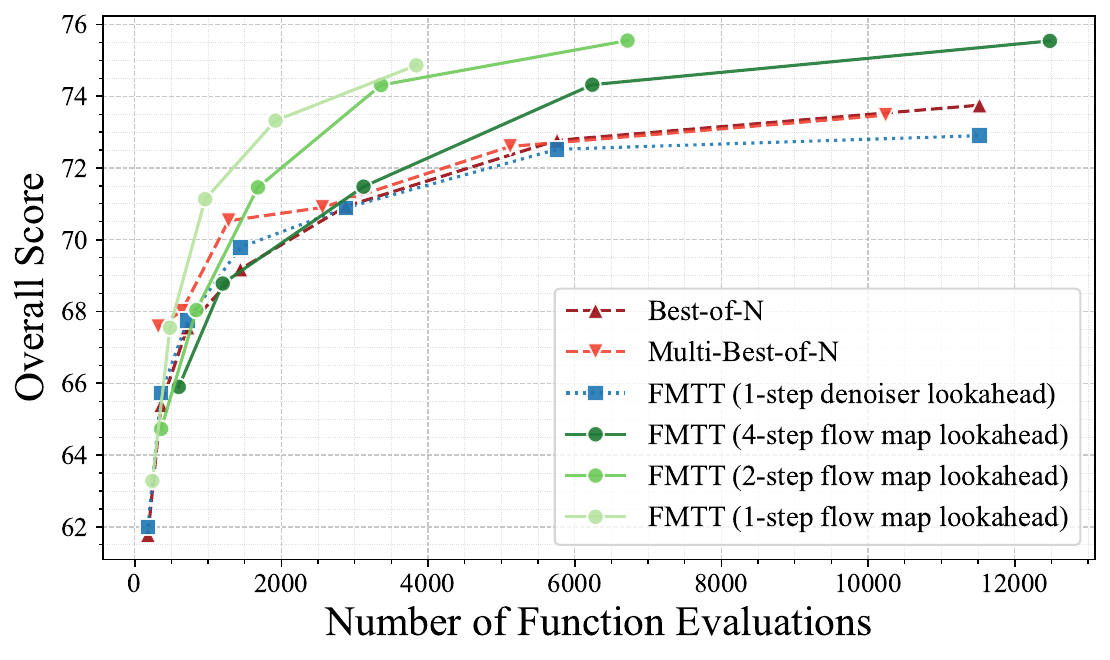}
    \end{minipage}
    \hfill
    \begin{minipage}[c]{0.43\textwidth}
        \caption{Scaling results on UniGenBench++. Overall score versus compute, measured in network evaluations. FMTT significantly outperforms both Best-of-$N$ and Multi-Best-of-$N$, achieving higher scores at comparable compute. FMTT with a 1-step denoiser look-ahead fails to beat Best-of-$N$, highlighting the ineffectiveness of reward gradients at blurry denoised states. See \Cref{app:unigenbench_details} for details.\looseness=-1}
        \label{fig:unigenbench_scaling}
    \end{minipage}
    \vspace{-6mm}
\end{figure}

\vspace{-1mm}
We summarize the results as a performance-vs-compute scaling plot in \Cref{fig:unigenbench_scaling}. We also provide a summary of all the metrics in \Cref{tab:unigenbench_mini} for the primary methods and the DAS baseline. The plot shows that, when comparing the flow map look-ahead, the 1-step denoiser look-ahead, and standard best-of-$N$ sampling, the denoiser look-ahead offers little to no improvement over the best-of-$N$ baseline. In contrast, FMTT with a flow map look-ahead consistently improves the performance at lower computational cost. We attribute the denoiser's weak performance to its heavily blurred predictions at early, high-noise timesteps, which fall far off the data manifold and result in unhelpful reward gradients. See \Cref{fig:viz_lookaheads} for side-by-side comparison of the different look-ahead strategies across different noise levels. Additionally, the strong performance obtained with a 1-step flow map look-ahead suggests that FMTT is not overly sensitive to the quality of the pretrained flow map, and is effective even with suboptimal or partially-trained flow maps that may be even cheaper to obtain. For a full performance breakdown including all metrics at all compute budgets, please see \Cref{tab:unigenbench}. Comparing the two DAS runs, we further find that flow map look-ahead improves prior test-time scaling methods as well.\looseness=-1

\begin{figure}[t]
  \centering
  \vspace{-3mm}
  \includegraphics[width=\linewidth]{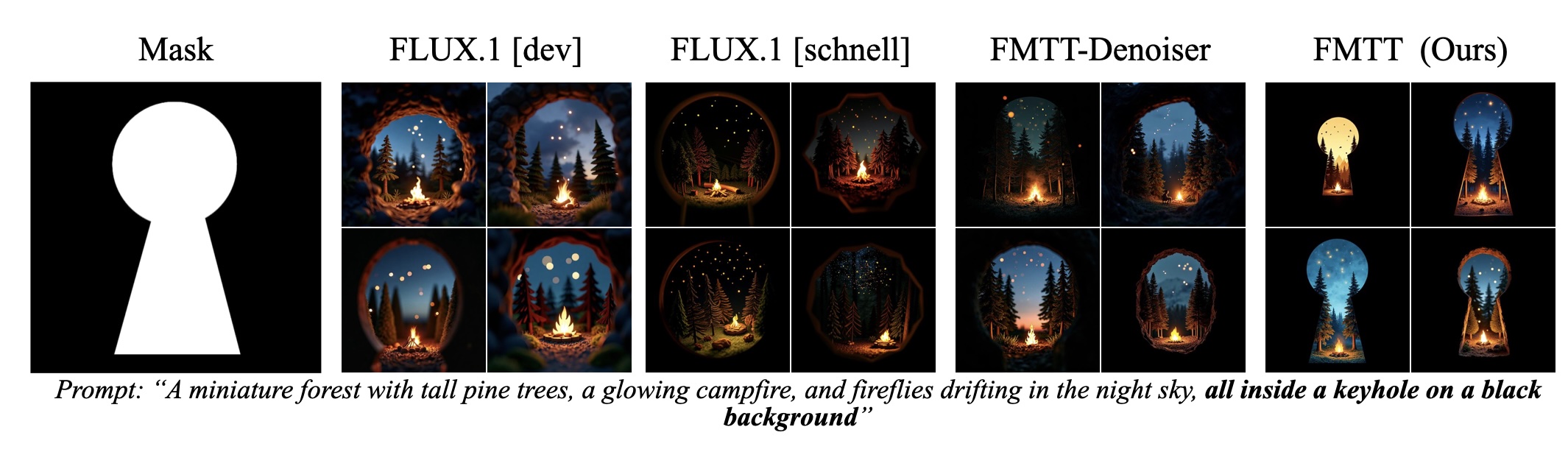}
  \vspace{-3mm}
  \caption{Qualitative comparison on masked rewards. Only our flow map-based FMTT reliably satisfies the constraints, only generating content in the unmasked regions. The base model (prompt only), and denoiser look-ahead fail to capture the constraint. For additional examples, see \Cref{app:additional_geometric_examples}.}
  \label{fig:masked_reward}
  \vspace{-6mm}
\end{figure}

\vspace{-1mm}
Finally, in \Cref{app:unigenbench_details}, we also evaluate the distilled FLUX.1 [Schnell] model, both with Best-of-$N$ and as an alternative look-ahead. Schnell + Best-of-$N$ achieves the best overall performance in this setting, likely because Schnell is distilled using latent adversarial distillation and trades off diversity for sample quality by concentrating on high-probability modes. However, as a gradient-free method, Best-of-$N$ remains limited to the model's original distribution and can fail on harder rewards that require generating samples outside these modes, such as \Cref{fig:teaser,fig:masked_reward}. Meanwhile, using Schnell as a look-ahead improves over denoiser look-ahead within FMTT, supporting our main claim that stronger look-aheads lead to more effective test-time trajectory tilting.\looseness=-1

\begin{figure}[t]
  \centering
  \vspace{-3mm}
  \includegraphics[width=\linewidth]{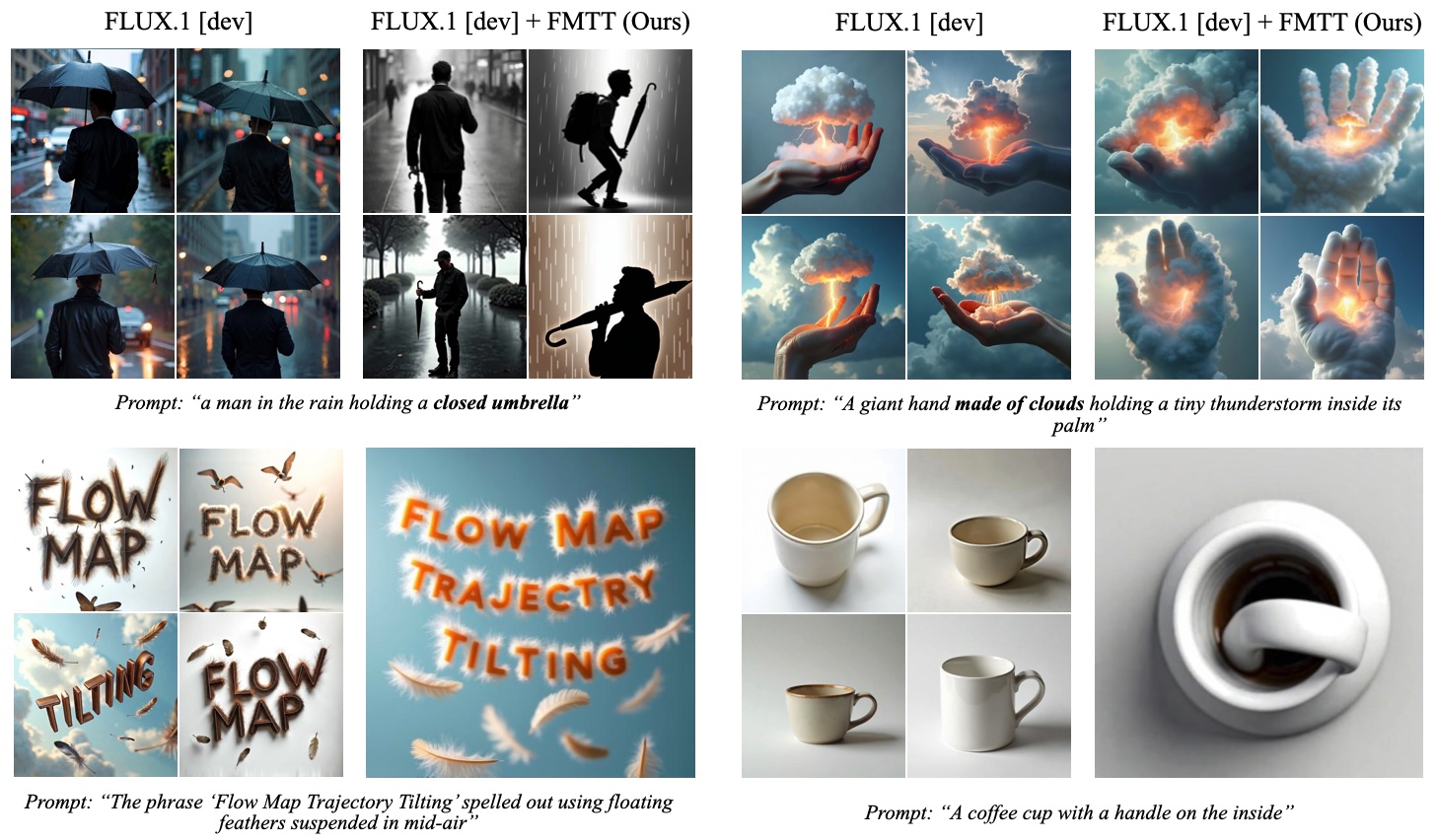}
  \caption{Additional qualitative results using VLM-based rewards on hard prompts. The base model defaults to the common interpretation of the prompt and ignores the unusual requirement highlighted in bold, whereas FMTT satisfies it while remaining realistic.\looseness=-1}
  \label{fig:vlm_single_image_extra}
  \vspace{-5mm}
\end{figure}

\vspace{-1mm}
Separately from the UniGenBench++ evaluation, we include additional qualitative examples in \Cref{fig:vlm_single_image}. The figure highlights two prompts where the base model fails to produce text-aligned outputs. In contrast, using the VLM-based reward, our FMTT algorithm generates outputs that match the input text much more accurately. \Cref{fig:vlm_single_image_extra} shows further examples on prompts that describe uncommon scenes, where the base model falls back to the more common interpretation of the prompt and FMTT recovers the intended one. For additional VLM-based experiments, including multi-image settings, please see \Cref{app:additional_vlm_examples}. 
One potential downside of FMTT is the possibility of reward hacking, as the search procedure explicitly maximizes the VLM reward which can contain unwanted artifacts. To mitigate this, we have observed that the yes/no questions must be written with sufficient detail to prevent the algorithm from exploiting loopholes~\citep{amodei2016concrete}. Further discussion and examples in  \Cref{app:vlm_reward_hacking}.\looseness=-1

\vspace{-2mm}
\subsection{Human Preference Rewards}
\vspace{-2mm}
Additionally, we quantitatively evaluate FMTT on  GenEval~\citep{Ghosh2023GenEvalAO}. This benchmark consists of ${\approx}$ 550 object-centric prompts and measures the quality of generated images using a pre-trained object detector. For the reward model, we follow prior work~\citep{eyring2024reno} and use a linear combination of PickScore~\citep{pickscore}, HPSv2~\citep{hpsv2}, ImageReward~\citep{xu2023imagereward}, and CLIPScore~\citep{clip}. The results are presented in \Cref{tab:geneval}.\looseness=-1  

The base model, FLUX.1 [dev], achieves strong scores due to its training on large amounts of object-centric data. Distillation into a 4-step flow map slightly reduces performance but significantly accelerates generation. A simple best-of-$N$ search on top of the flow map recovers this drop and surpasses the base diffusion model while remaining about 30\% faster. More advanced search methods, such as multi-best-of-$N$ or beam search, yield additional but modest gains. Using reward gradients with FMTT provides a further (small) improvement. Finally, we also ablate the use of the 4-step flow map look-ahead by comparing FMTT against variants using either a 1-step denoiser or a 4-step diffusion sampler. The flow map look-ahead consistently performs best, in line with our earlier findings.\looseness=-1  

We also note that FLUX.1 [dev] has already been post-trained with human preference data. This makes it harder for test-time optimization on similar rewards to substantially shift the output distribution further, which likely explains why the gains across the different methods are limited. Nevertheless, FMTT achieves the best overall performance, outperforming all other approaches by clear margin.\looseness=-1

\vspace{-2mm}
\subsection{Geometric Transformation Rewards}
\vspace{-2mm}
While human preference and VLM-based rewards are sufficient for many high-level objectives, they are often a poor fit when the goal requires fine, pixel-level control. A simple example is enforcing invariance under a geometric transform via $r(x) = -d(x, T(x))$, where $T(x):\mathbb{R}^d\!\to\!\mathbb{R}^d$ is a transformation and $d(\cdot,\cdot)$ is a distance metric. For example, if $T$ is a masking operator, maximizing this reward encourages the masked region to be black, which can be used to control where content is allowed to appear. Similar objectives can be defined for symmetry, anti-symmetry, rotation, and more. These rewards are highly specialized and are typically satisfied only in the far tails of the base model's output distribution, making them a challenging testbed for test-time scaling methods.

\Cref{fig:symm_rot_reward} shows that the base model roughly follows these objectives but does not fully satisfy them (e.g., small planets break symmetry, the cat's eyes are not anti-symmetric). Prior methods such as multi-best-of-$N$~\citep{Lee2025AdaptiveIS} and ReNO~\citep{eyring2024reno} also struggle, either violating the constraints or producing blurry images. 
In contrast, our gradient-based variants directly modify the dynamics, producing sharper outputs that reliably satisfy the constraints.\looseness=-1

For an even harder case, we use a masked-generation reward in \Cref{fig:masked_reward}. FMTT with a denoiser look-ahead increases the reward by darkening the masked region, but it does not consistently move all content into the unmasked area. Using a flow map look-ahead, however, successfully maximizes the reward and generates images that fully satisfy the constraint.\looseness=-1

\vspace{-2mm}
\section{Conclusions and Future Work} 
\vspace{-2mm}
We presented FMTT, a gradient-based test-time scaling framework for diffusion models that uses an efficient flow map for look-ahead prior to reward evaluation. 
FMTT consistently improves performance across benchmarks and enables sampling from long-tailed regions where the base model and prior test-time search methods often fail, thereby demonstrating state-of-the-art inference-time guidance capabilities. We also showed compatibility with diverse reward types, including human preference models, geometric transformation objectives, and VLM-based question-answering.
Promising directions for future work include integrating FMTT with online RL-based diffusion/flow map fine-tuning literature, applying these ideas to even more complex rewards, and scaling to domains such as image editing, video generation, world models, and scientific applications.\looseness=-1

\bibliographystyle{plainnat}
\bibliography{main}

\newpage
\appendix
\renewcommand\ptctitle{}
\addcontentsline{toc}{section}{Appendix} %
\part{Appendix} %
\vspace{6mm}
\parttoc %
\newpage

\section{Proofs}
\subsection{Proof of \Cref{prop:jar}} \label{app:proof:generic}

Consider the coupled SDE/ODE:
\begin{align}
\begin{split}
    d\tilde x_t &= \big( b_t(\tilde x_t) + \epsilon_t s_t(\tilde x_t) + \epsilon_t \nabla r_t(\tilde x_t) \big) dt + \sqrt{2 \epsilon_t} dW_t, \qquad \tilde x_0 \sim \rho_0, \\
    dA_t &= \big( b_t(\tilde{x}_t) \cdot \nabla r_t(\tilde{x}_t) + \partial_t r_t(\tilde{x}_t) \big) \, dt, \qquad A_0 = 0.
\end{split}
\end{align}
Let $f_t(x,a)$ be the joint probability density of $(\tilde x_t, A_t)$. Then, it satisfies the Fokker-Planck equation
\begin{align}
\begin{split}
    \partial_t f_t(x,a) &= - \nabla_x \cdot \big( \big( b_t(x) + \epsilon_t s_t(x) + \epsilon_t \nabla r_t(x) \big) f_t(x,a) \big) - \partial_a \big( \big( b_t(x) \cdot \nabla r_t(x) + \partial_t r_t(x) \big) f_t(x,a) \big) \\ &\quad + \epsilon_t \Delta_x f_t(x,a), \\
    f_0(x,a) &= \rho_0(x) \delta_0(a).
\end{split}
\end{align}
Observe that if we let $\bar{\rho}_t(x) = \int_{\mathbb{R}} f_t(x,a) e^{a} \, da$, $\bar{\rho}$ satisfies
\begin{align}
\begin{split} \label{eq:partial_t_hatrho}
    \partial_t \bar{\rho}_t(x) &= \int_{\mathbb{R}} \partial_t f_t(x,a) e^{a} \, da \\ &= \int_{\mathbb{R}} \big( - \nabla_x \cdot \big( \big( b_t(x) + \epsilon_t s_t(x) + \epsilon_t \nabla r_t(x) \big) f_t(x,a) \big) - \big( \big( b_t(x) \cdot \nabla r_t(x) + \partial_t r_t(x) \big) \partial_a f_t(x,a) \\ &\qquad\quad + \epsilon_t \Delta_x f_t(x,a) \big) e^{a} \, da \\ &= - \nabla_x \cdot \big( \big( b_t(x) + \epsilon_t s_t(x) + \epsilon_t \nabla r_t(x) \big) \int_{\mathbb{R}} f_t(x,a) e^{a} \, da \big) \\ &\qquad\quad - \big( b_t(x) \cdot \nabla r_t(x) + \partial_t r_t(x) \big) \int_{\mathbb{R}} \partial_a f_t(x,a) e^{a} \, da + \epsilon_t \Delta_x \bigg( \int_{\mathbb{R}} f_t(x,a) e^{a} \, da \bigg) \\ &= - \nabla_x \cdot \big( \big( b_t(x) + \epsilon_t s_t(x) + \epsilon_t \nabla r_t(x) \big) \bar{\rho}_t(x) \big) + \big( b_t(x) \cdot \nabla r_t(x) + \partial_t r_t(x) \big) \bar{\rho}_t(x) + \epsilon_t \Delta_x \bar{\rho}_t(x).
\end{split}    
\end{align}
where the fourth inequality holds through integration by parts: 
\begin{align}
\int_{\mathbb{R}} \partial_a f_t(x,a) e^{a} \, da = [f_t(x,a) e^{a}]_{-\infty}^{\infty} - \int_{\mathbb{R}} f_t(x,a) \partial_a e^{a} \, da = - \int_{\mathbb{R}} f_t(x,a) e^{a} \, da = - \bar{\rho}_t(x).
\end{align}
We can reinterpret equation \eqref{eq:partial_t_hatrho} as stating that for any test function $h$, 
\begin{align} \label{eq:barrho_int}
\int_{\mathbb{R}^d} h(x) \bar{\rho}_t(x) \, dx = \int_{\mathbb{R}^d} \int_{\mathbb{R}} f_t(x,a) e^{a} h(x) \, da \, dx = \mathbb{E}[e^{A_t} h(\tilde{x}_t)].
\end{align}
However, $\bar{\rho}_t$ is not a normalized density for $t \geq 0$:
if we integrate both sides of \eqref{eq:partial_t_hatrho} over $\mathbb{R}^d$, and we use the divergence theorem, we obtain that
\begin{align} \label{eq:partial_t_int}
    \partial_t \int_{\mathbb{R}^d} \bar{\rho}_t(x) \, dx = \int_{\mathbb{R}^d} \big( b_t(x) \cdot \nabla r_t(x) + \partial_t r_t(x) \big) \bar{\rho}_t(x) \, dx, \qquad \int_{\mathbb{R}^d} \bar{\rho}_0(x) \, dx = 1
\end{align}
If we define $F_t = \int_{\mathbb{R}^d} \bar{\rho}_t(x) \, dx$, and we define $\hat{\rho}_t(x) = \bar{\rho}_t(x) / F_t$, we obtain that
\begin{align}
\begin{split} \label{eq:partialhatrho}
    \partial_t \hat{\rho}_t(x) &= - \nabla_x \cdot \big( \big( b_t(x) + \epsilon_t s_t(x) + \epsilon_t \nabla r_t(x) \big) \frac{\bar{\rho}_t(x)}{F_t} \big) + \big( b_t(x) \cdot \nabla r_t(x) + \partial_t r_t(x) \big) \frac{\bar{\rho}_t(x)}{F_t} + \epsilon_t \Delta_x \frac{\bar{\rho}_t(x)}{F_t} \\ &\quad - \frac{\partial_t F_t}{F_t} \frac{\bar{\rho}_t(x)}{F_t},  
\end{split}
\end{align}
and by \eqref{eq:partial_t_int}, if we define $\hat{F}_t = \log F_t$, we have that
\begin{align}
    \partial_t \hat{F}_t = \frac{\partial_t F_t}{F_t} = \int_{\mathbb{R}^d} \big( b_t(x) \cdot \nabla r_t(x) + \partial_t r_t(x) \big) \frac{\bar{\rho}_t(x)}{F_t} \, dx = \int_{\mathbb{R}^d} \big( b_t(x) \cdot \nabla r_t(x) + \partial_t r_t(x) \big) \hat{\rho}_t(x) \, dx.
\end{align}
Plugging this into the right-hand side of \eqref{eq:partialhatrho} and substituting $\hat{\rho}_t(x) = \bar{\rho}_t(x) / F_t$ yields a PDE for $\hat{\rho}_t$ which matches \eqref{eq:dtrho:emp}:
\begin{align}
    \partial_t \hat{\rho}_t(x) &= - \nabla_x \cdot \big( \big( b_t(x) + \epsilon_t s_t(x) + \epsilon_t \nabla r_t(x) \big) \hat{\rho}_t(x) \big) + \big( b_t(x) \cdot \nabla r_t(x) + \partial_t r_t(x) - \partial_t \hat{F}_t \big) \hat{\rho}_t(x) + \epsilon_t \Delta_x \hat{\rho}_t(x).
\end{align}
To show that equation \eqref{eq:expect:map} holds, we rely on \eqref{eq:barrho_int} and the fact that $F_t = \int_{\mathbb{R}^d} \int_{\mathbb{R}} f_t(x,a) e^{a} \, da \, dx = \mathbb{E}[e^{A_t}]$:
\begin{align}
    \int_{\mathbb{R}^d} h(x) \hat{\rho}_t(x) \, dx = \frac{1}{F_t }\int_{\mathbb{R}^d} h(x) \bar{\rho}_t(x) \, dx = \frac{\mathbb{E}[e^{A_t} h(\tilde{x}_t)]}{\mathbb{E}[e^{A_t}]}.
\end{align}

\subsection{Proof of \Cref{prop:map:tilt}} \label{app:proof:map:tilt}

When $r_t(x) = t \cdot r(X_{t,1}(x))$, the log-weight $A_t$ defined in \eqref{eq:sde:A} satisfies the ODE
\begin{align}
\begin{split} \label{eq:dA_t_dt}
    \frac{\mathrm{d}A_t}{\mathrm{d}t} &= t \, b_t(\tilde{x}_t) \cdot \nabla_{\tilde{x}_t} r(X_{t,1}(\tilde{x}_t))  + \partial_t \big( t \, r(X_{t,1}(\tilde{x}_t)) \big) \\ &= t b_t(\tilde{x}_t) \cdot \nabla X_{t,1}(\tilde{x}_t)^{\top} \nabla r(X_{t,1}(\tilde{x}_t)) + r(X_{t,1}(\tilde{x}_t)) + t \, \partial_t X_{t,1}(\tilde{x}_t) \cdot \nabla r(X_{t,1}(\tilde{x}_t)) \\ &= 
    t \nabla r(X_{t,1}(\tilde{x}_t)) \cdot \big( \partial_t X_{t,1}(\tilde{x}_t) + \nabla X_{t,1}(\tilde{x}_t) b_t(\tilde{x}_t) \big) + r(X_{t,1}(\tilde{x}_t)),
\end{split}
\end{align}
The Eulerian identity states that
\begin{align} \label{eq:Eulerian}
    \partial_t X_{t,1}(x) + 
    \nabla X_{t,1}(x) b_t(x)
    = 0.
\end{align}
To prove \eqref{eq:Eulerian}, we write $0 = \partial_s (X_{s,1} \circ X_{t,s})(x) = \partial_s X_{s,1}(X_{t,s}(x)) + \nabla X_{s,1}(X_{t,s}(x)) \partial_s X_{t,s}(x)$, we use that $\partial_s X_{t,s}(x) = b(X_{t,s}(x))$, and we set $s = t$. Plugging \eqref{eq:Eulerian} into the right-hand side of \eqref{eq:dA_t_dt} yields
\begin{align}
    \frac{\mathrm{d}A_t}{\mathrm{d}t} = r(X_{t,1}(\tilde{x}_t)),
\end{align}
which concludes the proof.

\subsection{Modifying the Drift with the Flow Map Reward} \label{app:proof:map:tilt:reward}
While the position and log-weight dynamics in \Cref{prop:jar} and \Cref{prop:map:tilt} are remarkably simple to compute, they are not the only ones that can be used to sample from the tilted distribution. In particular, if we add an additional term $\chi_t \nabla r_t(\tilde{x}_t)$ to the position SDE, we can adjust the log-weights dynamics so that the coupled system still lets us compute expectations according to the tilted distribution $\hat{\rho}_t$ for all $t \in [0,1]$. Namely, the reward-modified position SDE reads:
\begin{align}
d\tilde x_t &= \left[ v_{t,t}(\tilde x_t) \! + \! 
    \chi_t \nabla r_t(\tilde x_t)
    \right] dt + \epsilon_t \left [s_t(\tilde x_t) \! + \! 
    \nabla r_t(\tilde x_t)
    \right] dt + \sqrt{2\epsilon_t} dW_t, \quad \tilde x_0 \sim \rho_0.
\end{align}
There are four choices of $\chi_t$ of particular interest:
\begin{enumerate}[label=(\roman*),left=2pt]
    \item \textbf{Default dynamics}: $\chi_t = 0$. This choice recovers the dynamics introduced in \Cref{sec:correcting}.
    \item \textbf{Tilted score dynamics}: $\chi_t = \eta_t :=\alpha_t (\frac{\dot{\beta}_t}{\beta_t} \alpha_t - \dot{\alpha}_t)$. When the dynamics \eqref{eq:ode} has been learned using stochastic interpolants, the score $s_t(x)$ and the vector field $b_t(x)$, or equivalently $v_{t,t}(x)$, are related to each other via the equation
    \begin{align} \label{eq:v_tt_score}
        b_t(x)
        = \alpha_t (\frac{\dot{\beta}_t}{\beta_t} \alpha_t- \dot{\alpha}_t) s_t(x) + \frac{\dot\beta_t}{\beta_t} x
    \end{align}
    which \eqref{eq:score} is a particular case of, when $\alpha_t = 1-t$, $\beta_t = t$. The score of the tilted distribution at time $t$ is $\hat{s}_t(x) = \nabla \log \hat{\rho}_t(x) = s_t(x) + \nabla r_t(x)$, and if we replace $s_t(x)$ by $\hat{s}_t(x)$ in \eqref{eq:v_tt_score}, we obtain
    \begin{align}
        b_t(x)
        = \alpha_t (\frac{\dot{\beta}_t}{\beta_t} \alpha_t- \dot{\alpha}_t) \hat{s}_t(x) + \frac{\dot\beta_t}{\beta_t} x = v_{t,t}(x) + \alpha_t (\frac{\dot{\beta}_t}{\beta_t} \alpha_t- \dot{\alpha}_t) \nabla r_t(x).
    \end{align}
    \item \textbf{Local tilt dynamics}: $\chi_t = \epsilon_t$. If we let $P(x_{(k+1)h}|x_{kh})$ be the transition kernel for the Euler-Maruyama discretization of the SDE 
    \begin{align}
        \mathrm{d}\tilde x_t = \big( v_{t,t}(\tilde x_t) + \epsilon_t s_t(\tilde x_t) 
        \big) \mathrm{d}t + \sqrt{2 \epsilon_t} \mathrm{d}W_t,
    \end{align} 
    the local tilt dynamics arises from tilting $P(x_{(k+1)h}|x_{kh})$ according to the reward $r_{(k+1)h}(x_{(k+1)h})$, which yields the transition kernel $P^{r}(x_{(k+1)h}|x_{kh}) \propto P(x_{(k+1)h}|x_{kh}) \exp(r_{(k+1)h}(x_{(k+1)h}))$. By \Cref{prop:cond_density_max_2}, when $h \to 0$, $P^{r}(x_{(k+1)h}|x_{kh})$ can be viewed as the transition kernel for the SDE
    \begin{align}
        d\tilde x_t &= \left[ v_{t,t}(\tilde x_t) \! + \! 
        \epsilon_t \nabla r_t(\tilde x_t)
        \right] dt + \epsilon_t \left [s_t(\tilde x_t) \! + \! 
        \nabla r_t(\tilde x_t)
        \right] dt + \sqrt{2\epsilon_t} dW_t,
    \end{align}
    which amounts to setting $\chi_t = \epsilon_t$ in \eqref{eq:sde:mod}.
    \item \textbf{Base dynamics}: $\chi_t = -\epsilon_t$. With this choice, the SDE \eqref{eq:sde:mod} becomes
    \begin{align}
        d\tilde x_t &= \left[ v_{t,t}(\tilde x_t) \! + \! 
        \epsilon_t s_t(\tilde x_t)
        \right] dt + \sqrt{2\epsilon_t} dW_t,
    \end{align}
    which does not involve the reward $r_t$.
\end{enumerate}

The following proposition, which generalizes \Cref{prop:jar}, shows the log-weight dynamics that we must use for generic reward multipliers $\chi_t$, written in three different ways. Observe that 
\begin{propbox}
\begin{proposition}[Unbiased Map Tilting with reward-modified vector field]
\label{prop:map:tilt:reward}
    Let 
    $(\chi_t)_{t \in [0,1]}$ be an arbitrary real valued function, and let $r_t$ be an arbitrary time-dependent reward.
    Let $(\tilde{x}_t,A_t)$ be a solution of 
    \begin{align} \label{eq:tilde_x_sde:reward}
        d\tilde x_t &= \big( 
        b_t(\tilde x_t) + \chi_t \nabla r_t(\tilde x_t) + \epsilon_t s_t(\tilde x_t) + \epsilon_t \nabla r_t(\tilde x_t) \big) dt + \sqrt{2 \epsilon_t} dW_t, \qquad \tilde x_0 \sim \rho_0, \\
        dA_t \! &= \! \big( b_t(\tilde{x}_t) \cdot \nabla r_t(\tilde{x}_t) \! + \! \partial_t r_t(\tilde{x}_t) \! + \! \chi_t \big( \| \nabla r_t(\tilde{x}_t)\|^2 \! + \! \Delta r_t(\tilde{x}_t) \! + \! \langle \nabla r_t(\tilde{x}_t), s_t(\tilde{x}_t) \rangle \big) \big) \, dt, \ A_0 = 0.
        \label{eq:a_ode:reward}
    \end{align}
    Then for all $t\in[0,1]$ and any test function $h:\mathbb R^d\to\mathbb R$, we have 
	\begin{equation}
 \label{eq:expect:map-mod}
		\int_{\mathbb R^d} h(x) \hat \rho_t(x) dx = \frac{\mathbb E[ e^{A_t} h( \tilde x_t)]}{\mathbb E[e^{A_t}]},
	\end{equation}
	where the expectations at the right-hand side are taken over the law of $(\tilde{x}, A)$. 
    The equation \eqref{eq:a_ode:reward} for $A_t$ can be rewritten as
    \begin{align}
        \begin{split}
        dA_t \! &= \! \big( b_t(\tilde{x}_t) \cdot \nabla r_t(\tilde{x}_t) \! + \! \partial_t r_t(\tilde{x}_t) \! + \! \chi_t \big( \| \nabla r_t(\tilde{x}_t)\|^2 
        \! + \! \langle \nabla r_t(\tilde{x}_t), s_t(\tilde{x}_t) \rangle \big) \big) \, dt \\ &\quad + \! \frac{\chi_t}{\sqrt{2 \epsilon_t}} 
        \nabla r_t (\tilde{x}_t)
        \cdot \mathrm{d}^{-}W_t 
        \! - \! \frac{\chi_t}{\sqrt{2 \epsilon_t}} \nabla r_t (\tilde{x}_t) \cdot \mathrm{d}W_t, \ A_0 = 0.
        \end{split}
        \label{eq:a_ode:reward2}
    \end{align}
    Equation \eqref{eq:a_ode:reward} can be rewritten in a third way as
    \begin{align}
        \begin{split}
        \mathrm{d}A_t \! &= \begin{cases} 
        \frac{\mathrm{d}}{\mathrm{d}s} \mathbb{E}[ e^{r_s(X^{\chi}_s)-r_t(\tilde{x}_t)} \,|\, X^{\chi}_t = \tilde{x}_t ] \big|_{s = t} \, \mathrm{d}t, &\text{if } \chi_t \geq 0, \\
        \big( 2 \frac{\mathrm{d}}{\mathrm{d}s} r_s(X^{0}_s) \big|_{\substack{X^{0}_t = \tilde{x}_t \\ s = t}}  - \frac{\mathrm{d}}{\mathrm{d}s} \mathbb{E}[ e^{r_s(X^{\chi}_s)-r_t(\tilde{x}_t)} \,|\, X^{\chi}_t = \tilde{x}_t ] \big|_{s = t} \big) \, \mathrm{d}t, &\text{if } \chi_t < 0,
        \end{cases}
        \\
         A_0 &= 0,
        \end{split}
        \label{eq:a_ode:reward3}
    \end{align}
    where $X^{\chi}$ is a solution of the SDE
    \begin{align} \label{eq:process_eta_t}
        \mathrm{d}X^{\chi}_t = \big( b_t(X^{\chi}_t) + |\chi_t| s_t(X^{\chi}_t) \big) \mathrm{d}t + \sqrt{2 |\chi_t|} \mathrm{d}W_t,
    \end{align}
    and $X^{0}$ is a solution of the ODE $\mathrm{d}X^{0}_t = b_t(X^{\chi}_t) \, \mathrm{d}t$.
\end{proposition}
\end{propbox}

\begin{proof}    
\textbf{(i) Proof of equation \eqref{eq:expect:map-mod}}. When we replace $b_t(x)$ by $\tilde{b}_t(x) = b_t(x) + \chi_t \nabla r_t(x)$, the analog of equation \eqref{eq:stePrhor} is:
\begin{equation}
\label{eq:stePrhor:reward}
\begin{aligned}
\partial_t \hat \rho_t + \nabla \cdot (\tilde{b}_t \hat \rho_t) &= \partial_t ( e^{r_t + \hat F_t} \rho_t) + \nabla \cdot (\tilde{b}_t e^{r_t + \hat F_t} \rho_t)\\
&= (\partial_t r_t + \partial_t \hat F_t) \hat \rho_t + e^{r_t + \hat F_t} \partial_t \rho_t
+ \big( b_t + \chi_t \nabla r_t \big) \cdot \nabla r_t \hat \rho_t + e^{r_t + \hat F_t} \nabla \cdot( (b_t + \chi_t \nabla r_t) \rho_t)\\
&= (\partial_t r_t + \partial_t \hat F_t) \hat \rho_t  + b_t \cdot \nabla r_t \hat\rho_t + \chi_t \| \nabla r_t\|^2 \hat \rho_t + 
\chi_t e^{r_t + \hat F_t} \nabla \cdot( \nabla r_t \rho_t)
\\
&= \big(b_t \cdot \nabla r_t + \partial_t r_t + \chi_t \big( \| \nabla r_t\|^2 + \Delta r_t + \langle \nabla r_t, s_t \rangle \big) + \partial_t \hat F_t \big) \hat \rho_t,
\end{aligned} 
\end{equation}
where the third equality holds because $\partial_t \rho_t + \nabla \cdot (b_t \rho_t) = 0$ by the FPE \eqref{eq:fpe} with $\epsilon_t \equiv 0$, and in the fourth equality we used the definition $s_t(x) = \nabla \log \rho_t(x)$. Hence, when we replace $b_t$ by $\tilde{b}_t$, the terms $b_t \cdot \nabla r_t + \partial_t r_t$ get replaced by $b_t \cdot \nabla r_t + \partial_t r_t + \chi_t \big( \| \nabla r_t\|^2 + \Delta r_t + \langle \nabla r_t, s_t \rangle \big)$.

Given equation \eqref{eq:stePrhor:reward}, the proof of the statement \eqref{eq:expect:map-mod} 
is analogous to the proof of \Cref{prop:jar} in \Cref{app:proof:generic}, simply replacing $b_t$ by $b_t + \chi_t \nabla r_t$, and $b_t \cdot \nabla r_t + \partial_t r_t$ by $b_t \cdot \nabla r_t + \partial_t r_t + \chi_t \big( \| \nabla r_t\|^2 + \Delta r_t + \langle \nabla r_t, s_t \rangle \big)$. We omit a full proof.

\textbf{(ii) Proof of equation \eqref{eq:a_ode:reward2}}. We proceed to prove that \eqref{eq:a_ode:reward2} is equivalent to \eqref{eq:a_ode:reward}. Observe that the solution of equation \eqref{eq:a_ode:reward} is
\begin{align} \label{eq:A_t_integral}
    A_t = \int_0^t \bigg( 
    b_t(\tilde{x}_t) \cdot \nabla r_t(\tilde{x}_t) \! + \! \partial_t r_t(\tilde{x}_t)
    \! + \! \chi_{\tau} \big( \| \nabla r_{\tau}(\tilde{x}_{\tau})\|^2 \! + \! \Delta r_{\tau}(\tilde{x}_{\tau}) \! + \! \langle \nabla r_{\tau}(\tilde{x}_{\tau}), s_{\tau}(\tilde{x}_{\tau}) \rangle \big) \bigg) \, dt.
\end{align}
Next, we handle the term $\nabla r_t$. 
Applying \Cref{lem:difference_integrals_application}, we obtain that
\begin{align}
\begin{split} \label{eq:a2_application}
    &\int_0^t \chi_{\tau} \Delta r_{\tau}(\tilde{x}_{\tau}) \, d\tau = \int_0^{t} \tau 
    \chi_{\tau} \Delta (r \circ X_{\tau,1})(\tilde{x}_{\tau}) \, \mathrm{d}\tau \\ &= \int_t^{t'} \frac{\tau 
    \chi_{\tau}
    }{\sqrt{2 \epsilon_{\tau}}} \nabla (r \circ X_{\tau,1})(\tilde{x}_{\tau}) \cdot \mathrm{d}^{-}W_{\tau} 
    - \int_t^{t'} \frac{\tau \chi_{\tau}}{\sqrt{2 \epsilon_{\tau}}} \nabla (r \circ X_{\tau,1})(\tilde{x}_{\tau}) \cdot \mathrm{d}W_{\tau},
\end{split}
\end{align}
And plugging this back into \eqref{eq:A_t_integral} yields equation \eqref{eq:a_ode:reward2}.

\textbf{(iii) Proof of equation \eqref{eq:a_ode:reward3}}. Finally, we prove that \eqref{eq:a_ode:reward3} is equivalent to \eqref{eq:a_ode:reward}. We introduce the necessary concepts first. The semi-group for a (time-inhomogeneous) Markov process is defined as
\[
P_{s,t} f(x) = \mathbb{E}\!\left[ f(X_t) \,\big|\, X_s = x \right], \qquad 0 \le s \le t.
\]
The (time-dependent) infinitesimal generator is defined by
\begin{align}
\label{eq:inf_generator}
L_t f_t(x) = \lim_{h \downarrow 0} \frac{P_{t,t+h} f_{t+h}(x) - f_t(x)}{h} = \frac{\mathrm{d}}{\mathrm{d}s} \mathbb{E}[ f_s(X_s) \,|\, X_t = x ] \big|_{s = t}.
\end{align}
For the process $X^{\chi}$ that solves \eqref{eq:process_eta_t},
the infinitesimal generator takes the form
\begin{align} \label{eq:L_t_f_t}
    L^{\chi}_t f_t(x) = \partial_t f_t(x) + (b_t(x) + |\chi_t| s_t(x)) \cdot \nabla f_t(x) + |\chi_t| \Delta f_t(x).
\end{align}
We define the Hamilton-Jacobi-Bellman (HJB) residual $\mathcal{R}^{\chi}_t$ for the reward $r_t$ and the process $X^{\chi}$ as
\begin{align} \label{eq:residual_HJB_def}
     \mathcal{R}^{\chi}_t(x) = b_t(x) \cdot \nabla r_t(x) \! + \! \partial_t r_t(x) \! + \! |\chi_t| \big( \| \nabla r_t(x)\|^2 \! + \! \Delta r_t(x) \! + \! \nabla r_t(x) \cdot s_t(x) \big).
\end{align}
The term "HJB residual" stems from the fact that the Hamilton-Jacobi-Bellman equation for a time-dependent reward $r_t$ and process $X^{\chi}$ can be expressed as $\mathcal{R}^{\chi}_t(x) = 0$ for all $t$ and $x$\footnote{Note that in stochastic optimal control, the HJB equation is usually written for the value function $V_t = -r_t$.}.

We start with the case $\chi_t \geq 0$. The term in the right-hand side of \eqref{eq:a_ode:reward} can be rewritten as follows:
\begin{align}
\begin{split} \label{eq:R_chi_rewritten}
    &b_t(x) \cdot \nabla r_t(x) + \partial_t r_t(x) + \chi_t \big( \| \nabla r_t(x)\|^2 + \Delta r_t(x) + \langle \nabla r_t(x), s_t(x) \rangle \big) = \mathcal{R}^{\chi}_t(x) \\ &= e^{-r_t(x)} [L^{\chi}_t e^{r_t}](x) = e^{-r_t(x)} \frac{\mathrm{d}}{\mathrm{d}s} \mathbb{E}[ e^{r_s(X^{\chi}_s)} \,|\, X^{\chi}_t = x ] \big|_{s = t} = \frac{\mathrm{d}}{\mathrm{d}s} \mathbb{E}[ e^{r_s(X^{\chi}_s)-r_t(x)} \,|\, X^{\chi}_t = x ] \big|_{s = t}.
\end{split}
\end{align}
where the second equality holds by \Cref{lem:inf_gen_residual}, and the third equality holds by the definition of the infinitesimal generator in \eqref{eq:inf_generator}. This concludes the case $\chi_t \geq 0$. 

We move on to the case $\chi_t \geq 0$. The term in the right-hand side of \eqref{eq:a_ode:reward} can be rewritten as follows:
\begin{align}
\begin{split}
    &b_t(x) \cdot \nabla r_t(x) + \partial_t r_t(x) + \chi_t \big( \| \nabla r_t(x)\|^2 + \Delta r_t(x) + \langle \nabla r_t(x), s_t(x) \rangle \big) \\ &= 2 \big( b_t(x) \cdot \nabla r_t(x) + \partial_t r_t(x) \big) \\ &\quad - \big( b_t(x) \cdot \nabla r_t(x) + \partial_t r_t(x) + |\chi_t| \big( \| \nabla r_t(x)\|^2 + \Delta r_t(x) + \langle \nabla r_t(x), s_t(x) \rangle \big) \big) 
    \\ &= 2 \mathcal{R}^0_t(x) - \mathcal{R}^{\chi}_t(x) =  2 \frac{\mathrm{d}}{\mathrm{d}s} \big( e^{r_s(X^{0}_s)-r_t(x)} \big) \big|_{\substack{X^{0}_t = x \\ s = t}}  - \frac{\mathrm{d}}{\mathrm{d}s} \mathbb{E}[ e^{r_s(X^{\chi}_s)-r_t(\tilde{x}_t)} \,|\, X^{\chi}_t = x ] \big|_{s = t}
    \\ &=  2 \frac{\mathrm{d}}{\mathrm{d}s}  r_s(X^{0}_s) \big|_{\substack{X^{0}_t = x \\ s = t}}  - \frac{\mathrm{d}}{\mathrm{d}s} \mathbb{E}[ e^{r_s(X^{\chi}_s)-r_t(\tilde{x}_t)} \,|\, X^{\chi}_t = x ] \big|_{s = t}
\end{split}
\end{align}
where $\mathcal{R}^0_t$ is defined as $\mathcal{R}^{\chi}_t$ with $\chi = 0$, and the last equality holds by equation \eqref{eq:R_chi_rewritten} applied to $\mathcal{R}^0_t$ and $\mathcal{R}^{\chi}_t$.
\end{proof}

The following corollary particularizes \Cref{prop:map:tilt:reward} to the case in which the time-dependent reward is built from the flow map $X_{t,1}$, thus generalizing \Cref{prop:map:tilt}.
\begin{corollary}[Unbiased Map Tilting with reward-modified vector field from flow map] \label{cor:unbiased_flow_map}
    When
    $r_t(x) = t \cdot r(X_{t,1}(x))$, equations 
    \eqref{eq:a_ode:reward}, \eqref{eq:a_ode:reward2} and \eqref{eq:a_ode:reward3} simplify respectively to
    \begin{align}
        \mathrm{d}A_t \! &= \! \big( \frac{r_t(\tilde{x}_t)}{t} \! + \! \chi_t \big( \| \nabla r_t(\tilde{x}_t)\|^2 \! + \! \Delta r_t(\tilde{x}_t) \! + \! \langle \nabla r_t(\tilde{x}_t), s_t(\tilde{x}_t) \rangle \big) \big) \, \mathrm{d}t, \ A_0 = 0.
        \label{eq:a_ode:reward:simplified}
    \end{align}
    \begin{align}
        \begin{split}
        \mathrm{d}A_t \! &= \! \big( \frac{r_t(\tilde{x}_t)}{t} \! + \! \chi_t \big( \| \nabla r_t(\tilde{x}_t)\|^2 
        \! + \! \langle \nabla r_t(\tilde{x}_t), s_t(\tilde{x}_t) \rangle \big) \big) \, \mathrm{d}t \\ &\quad + \! \frac{\chi_t}{\sqrt{2 \epsilon_t}} 
        \nabla r_t (\tilde{x}_t)
        \cdot \mathrm{d}^{-}W_t 
        \! - \! \frac{\chi_t}{\sqrt{2 \epsilon_t}} \nabla r_t (\tilde{x}_t) \cdot \mathrm{d}W_t, \ A_0 = 0.
        \end{split}
        \label{eq:a_ode:reward2:simplified}
    \end{align}
    \begin{align}
        \begin{split}
        \mathrm{d}A_t \! &= \begin{cases} 
        \partial_s \mathbb{E}[ e^{r_s(X^{\chi}_s)-r_t(\tilde{x}_t)} \,|\, X^{\chi}_t = \tilde{x}_t ] \big|_{s = t} \, \mathrm{d}t, &\text{if } \chi_t \geq 0, \\
        \big( \frac{2 r_t(\tilde{x}_t)}{t}  - \partial_s \mathbb{E}[ e^{r_s(X^{\chi}_s)-r_t(\tilde{x}_t)} \,|\, X^{\chi}_t = \tilde{x}_t ] \big|_{s = t} \big) \, \mathrm{d}t, &\text{if } \chi_t < 0,
        \end{cases}
        \qquad
         A_0 = 0,
        \end{split}
        \label{eq:a_ode:reward3:simplified}
    \end{align}
\end{corollary}
\begin{proof}
    Using the argument in \Cref{app:proof:map:tilt} yields
    \begin{align}
        b_t(x) \cdot \nabla r_t(x) + \partial_t r_t(x) = r(X_{t,1}(x)) = \frac{r_t(x)}{t}.
    \end{align}
    Substituting this into equations \eqref{eq:a_ode:reward}, \eqref{eq:a_ode:reward2} and \eqref{eq:a_ode:reward3} yields the simplified forms.
\end{proof}

\begin{remark}[Setting $\chi_t = 0$ in \Cref{prop:map:tilt:reward} and \Cref{cor:unbiased_flow_map}] \label{rem:chi_zero}
    \Cref{prop:map:tilt:reward} and \Cref{cor:unbiased_flow_map} show that the default choice $\chi_t = 0$ is the only one for which the evolution of $A_t$ simplifies. Namely, when $\chi_t = 0$, equations \eqref{eq:a_ode:reward}, \eqref{eq:a_ode:reward2} and \eqref{eq:a_ode:reward3} in \Cref{prop:map:tilt:reward} simplify respectively to
    \begin{align}
        \mathrm{d}A_t \! &= \! 
        \big( b_t(\tilde{x}_t) \cdot \nabla r_t(\tilde{x}_t) \! + \! \partial_t r_t(\tilde{x}_t) \big)
        \, \mathrm{d}t, \\
        \mathrm{d}A_t \! &= \! 
        \big( b_t(\tilde{x}_t) \cdot \nabla r_t(\tilde{x}_t) \! + \! \partial_t r_t(\tilde{x}_t) \big)
        \, \mathrm{d}t, \\
        \mathrm{d}A_t \! &= \! \frac{\mathrm{d}}{\mathrm{d}s} r_s(X^{0}_s) \big|_{\substack{X^{0}_t = \tilde{x}_t \\ s = t}} \, \mathrm{d}t,
    \end{align}
    and equations \eqref{eq:a_ode:reward:simplified}, \eqref{eq:a_ode:reward2:simplified}, and \eqref{eq:a_ode:reward3:simplified} in \Cref{cor:unbiased_flow_map} become
    \begin{align}
        \mathrm{d}A_t \! &= \! \frac{r_t(\tilde{x}_t)}{t} \, \mathrm{d}t, \\
        \mathrm{d}A_t \! &= \! \frac{r_t(\tilde{x}_t)}{t} \, \mathrm{d}t, \\
        \mathrm{d}A_t \! &= \! \frac{\mathrm{d}}{\mathrm{d}s} r_s(X^{0}_s) \big|_{\substack{X^{0}_t = \tilde{x}_t \\ s = t}} \, \mathrm{d}t,
    \end{align}
\end{remark}

\subsection{Solving the SDE/ODE System Numerically} \label{subsec:solving_numerically}
We need to simulate the coupled dynamics on weights and positions numerically. The SDE \eqref{eq:tilde_x_sde:reward} for $\tilde{x}_t$ can be solved via a $K$-step Euler-Maruyama scheme with time discretization $(t_k)_{k=0}^{K}$ and updates of the form
\begin{align}
    \begin{split}
X_k \! &= \! X_{k-1} \! + \! (t_{k} \! - \! t_{k-1}) [b_{t_{k-1}}(X_{k-1}) \! + \! \chi_{t_{k-1}} \nabla r_{t_{k-1}}(X_{k-1}) \\ &\qquad\qquad \ + \! \epsilon_{t_{k-1}} (s_{t_{k-1}}(X_{k-1}) \! + \!
    \nabla r_{t_{k-1}}(X_{k-1}))] 
    + \sqrt{2 \epsilon_k (t_{k} \! - \! t_{k-1})} \xi_{k-1},
\end{split}
\end{align}
where $\xi_{k-1} \sim N(0,\mathrm{I})$.
For generic $\chi_t$, solving the dynamics for the log-weight $A_t$ requires one of the following approaches, which correspond to equations 
\eqref{eq:a_ode:reward}, \eqref{eq:a_ode:reward2} and \eqref{eq:a_ode:reward3}, respectively:
\begin{enumerate}[left=-2pt,label=(\roman*)]
    \item \textbf{Estimating the Laplacian $\Delta r_t$}: If solving \eqref{eq:a_ode:reward}, we need to estimate the Laplacian $\Delta r_t(x)$, which can be done using the Hutchinson trace estimator $\widehat{\Delta r_t}(x) = \frac{1}{2M\varepsilon} \sum_{m=1}^{M}
z_m \cdot \bigl[\nabla r_t(x+\varepsilon z_m) - \nabla r_t(x-\varepsilon z_m)\bigr]$, where $z_m \sim \mathcal{N}(0,I)\ \text{or Rademacher},\ \varepsilon \ll 1$. Then, the log-weights can be simulated with Euler updates of the form:
    \begin{align}
    \begin{split} \label{eq:A_k_Laplacian}
        A_k \! &= \! A_{k-1} \! + \! (t_{k} \! - \! t_{k-1}) \big[ b_{t_{k-1}}(X_{k-1}) \cdot \nabla r_{t_{k-1}}(X_{k-1}) \! + \! \partial_t r_{t_{k-1}}(X_{k-1}) \\ &\qquad\qquad\qquad\qquad\quad + \! \chi_{t_{k-1}} \big( \| \nabla r_{t_{k-1}}(X_{k-1})\|^2 \! + \! \widehat{\Delta r_{t_{k-1}}}(X_{k-1}) \\ &\qquad\qquad\qquad\qquad\qquad\qquad \ + \! \langle \nabla r_{t_{k-1}}(X_{k-1}), s_{t_{k-1}}(X_{k-1}) \rangle \big) \big]
    \end{split}
    \end{align}
    When $\chi_{(k-1)h} = 0$, we do not need to compute the estimate the Laplacian, which makes the update much faster.
    \item \textbf{Estimating the difference of forward and backward Itô integrals}: If solving \eqref{eq:a_ode:reward2}, we estimate
    \begin{align}
    \begin{split} \label{eq:diff_ito_approx}
        &\int_{t_{k-1}}^{t_k} \frac{\chi_{\tau}}{\sqrt{2 \epsilon_{\tau}}} 
    \nabla r_{\tau} (X_{k-1})
    \cdot \mathrm{d}^{-}W_{\tau} 
    \! - \! \int_{t_{k-1}}^{t_k} \frac{\chi_{\tau}}{\sqrt{2 \epsilon_{\tau}}} \nabla r_{\tau} (\tilde{x}_{\tau}) \cdot \mathrm{d}W_{\tau} \\ &\approx \chi_{t_{k}} \sqrt{\frac{t_{k} \! - \! t_{k-1}}{2 \epsilon_{t_{k}}}} \nabla r_{t_{k}} (X_{k}) \cdot \xi_{k-1} \! - \! \chi_{t_{k-1}} \sqrt{\frac{t_{k} \! - \! t_{k-1}}{2 \epsilon_{t_{k-1}}}} \nabla r_{t_{k-1}} (X_{k-1}) \cdot \xi_{k-1}.
    \end{split}
    \end{align}
    Thus, the update reads
    \begin{align}
    \begin{split} \label{eq:A_k_ito}
        A_k \! &= \! A_{k-1} \! + \! (t_k-t_{k-1}) \bigg[ 
        b_{t_{k-1}}(X_{k-1}) \cdot \nabla r_{t_{k-1}}(X_{k-1}) \! + \! \partial_t r_{t_{k-1}}(X_{k-1})
        \\ &\qquad\qquad\qquad\qquad\quad + \! \chi_{t_{k-1}} \big( \|\nabla r_{t_{k-1}}\|^2 \! + \! \nabla r_{t_{k-1}} \cdot s_{t_{k-1}} \big)(X_{k-1}) \bigg] \\ &\qquad\quad + \! \chi_{t_{k}} \sqrt{\frac{t_{k} \! - \! t_{k-1}}{2 \epsilon_{t_{k}}}} \nabla r_{t_{k}} (X_{k-1}) \cdot \xi_{k-1}^{n} \! - \! \chi_{t_{k-1}} \sqrt{\frac{t_{k} \! - \! t_{k-1}}{2 \epsilon_{t_{k-1}}}} \nabla r_{t_{k-1}} (X_{k-1}) \cdot \xi_{k-1}^{n}.
    \end{split}
    \end{align}
    \item \textbf{Estimating expectations of $\exp(r_t)$}: If solving \eqref{eq:a_ode:reward3}, we approximate
    \begin{align}
    \begin{split}
        &\frac{\mathrm{d}}{\mathrm{d}s} \mathbb{E}[ e^{r_s(X^{\chi}_s)-r_t(x)} \,|\, X^{\chi}_t = x ] \big|_{s = t} \\ &\approx \frac{1}{M(t_{k} - t_{k-1})} \sum_{m=1}^{M} \exp\bigg( r_{t_k}(x + (t_{k} - t_{k-1}) (b_{t_{k-1}}(x) + |\chi_{t_{k-1}}| s_{t_{k-1}}(x) ) \\ &\qquad\qquad\qquad\qquad\qquad\qquad\quad + \sqrt{2 |\chi_{t_{k-1}}| (t_{k} \! - \! t_{k-1})} \xi_{k-1}^{(m)} ) 
        - r_{t_{k-1}}(x) \bigg), \quad \xi_{k-1}^{(m)} \sim N(0,\mathrm{I}).
    \end{split}
    \end{align}
    where we used a finite-difference approximation of the derivative, and the Euler-Maruyama update corresponding to the SDE \eqref{eq:process_eta_t}. Thus, for $\chi_{(k-1)h} \geq 0$ the update reads
    \begin{align}
    \begin{split} \label{eq:approach_3_positive}
        A_k \! &= \! A_{k-1} \! + \! \frac{1}{M} \sum_{m=1}^{M} \exp\bigg( r_{t_k}(X_{k-1} \! + \! (t_{k} \! - \! t_{k-1}) (b_{t_{k-1}}(X_{k-1}) \! + \! |\chi_{t_{k-1}}| s_{t_{k-1}}(X_{k-1}) ) 
        \\ &\qquad\qquad\qquad\qquad\qquad\quad \
        + \! \sqrt{2 |\chi_{t_{k-1}}| (t_{k} \! - \! t_{k-1})} \xi_{k-1}^{(m)} ) 
        - r_{t_{k-1}}(X_{k-1}) \bigg), \quad \xi_{k-1}^{(m)} \sim N(0,\mathrm{I}).
    \end{split}
    \end{align}
    Similarly, 
    \begin{align}
        \frac{\mathrm{d}}{\mathrm{d}s}  r_s(X^{0}_s) \big|_{\substack{X^{0}_t = x \\ s = t}} \approx \frac{r_{t_k}(x + (t_{k} - t_{k-1}) b_{t_k}(x)) - r_{t_k}(x)}{t_{k} - t_{k-1}}.
    \end{align}
    Thus, for $\chi_{(k-1)h} \leq 0$ the update reads
    \begin{align}
    \begin{split} \label{eq:approach_3_negative}
        A_k \! &= \! A_{k-1} \! + \! 2 \big( r_{t_k}(X_{k-1} + (t_{k} - t_{k-1}) b_{t_k}(X_{k-1})) - r_{t_{k-1}}(X_{k-1}) \big) \\ &\qquad\quad \ - \! \frac{1}{M} \sum_{m=1}^{M} \exp\bigg( r_{t_k}(X_{k-1} \! + \! (t_{k} \! - \! t_{k-1}) (b_{t_{k-1}}(X_{k-1}) \! + \! |\chi_{t_{k-1}}| s_{t_{k-1}}(X_{k-1}) ) 
        \\ &\qquad\qquad\qquad\qquad\qquad\quad \
        + \! \sqrt{2 \epsilon_k (t_{k} \! - \! t_{k-1})} \xi_{k-1}^{n} ) 
        - r_{t_{k-1}}(X_{k-1}) \bigg), \quad \xi_{k-1}^{(m)} \sim N(0,\mathrm{I}).
    \end{split}    
    \end{align}
    Observe that for $\chi_{(k-1)h} = 0$, equations \eqref{eq:approach_3_negative} and \eqref{eq:approach_3_positive} simplify to 
    \begin{align}
        A_k \! &= \! A_{k-1} \! + \! r_{t_k}(X_{k-1} + (t_{k} - t_{k-1}) b_{t_k}(X_{k-1})) - r_{t_{k-1}}(X_{k-1}),
    \end{align}
    which makes the update much less expensive.
\end{enumerate}

Observe that following \Cref{cor:unbiased_flow_map}, when $r_t(x) = t \cdot r(X_{t,1}(x))$ the terms $b_{t_{k-1}}(X_{k-1}) \cdot \nabla r_{t_{k-1}}(X_{k-1}) \! + \! \partial_t r_{t_{k-1}}(X_{k-1})$ in \eqref{eq:A_k_Laplacian} and \eqref{eq:A_k_ito} can be replaced by $\frac{r_{t_{k-1}}(X_{k-1})}{t_{k-1}}$.

\subsubsection{On the best-performing approach in practice} \label{subsubsec:best_performing}
When $\chi_t \neq 0$, the approach \emph{(i)} requires evaluating the gradient $\nabla r_t$ at $M$ points to obtain an estimate with variance $\Theta(1/M)$, and the approach \emph{(iii)} requires evaluating $r_t$ at $M$ points to obtain an estimate with variance $\Theta(1/M)$. Meanwhile, the approach \emph{(ii)} only requires evaluating the gradient $\nabla r_{t}$ at the iterates $(X_k)_{k=0}^{K}$. Hence, a priori, the approach \emph{(ii)} is preferable. 

However, when the reward functions $r$ is a neural network, such as a classifier, a vision-language model, or a CLIP model, the gradients $\nabla r$ are very noisy as $r$ is highly non-smooth: if we perturb a sample $x$ with an amount of Gaussian noise small enough that the noiseless sample $x$ and the noised sample $\hat{x}$ have very similar reward values, and compute cosine similarity of the gradients $\nabla r(x)$ and $\nabla r(\hat{x})$, we generally obtain low values ($\ll 0.5$). The noisy gradients cause the difference of integrals in \eqref{eq:diff_ito_approx} to be poor, as the gradients $\nabla r_{t_{k}} (X_{k})$ and $\nabla r_{t_{k-1}} (X_{k-1})$ are much more misaligned than they would be if the function was smoother. The noisy gradient issue also affect approach \emph{(i)}, but it does not affect approach \emph{(ii)}, which only relies on evaluations of $r_t$ but not its gradient. The issue can be mitigated by replacing the gradient evaluations $\nabla r_{t}$ by averages of the evaluations on noised samples, effectively estimating the gradient of $r_t$ convolved with a Gaussian. However, running approach \emph{(ii)} with this mitigation requires using several evaluations of $\nabla r_t$ anyway. In practice, we observe that approach \emph{(iii)} performs best, and that is the approach we used in our MNIST experiments in \Cref{app:additional_MNIST_experiments}.

\subsection{Lemmas used in \Cref{app:proof:map:tilt:reward}}
\begin{lemma}[Simulating the local tilt dynamics] \label{prop:cond_density_max_2}
    Consider the Euler-Maruyama discretization of the SDE $\mathrm{d}\tilde x_t = b^{\epsilon}_t(\tilde x_t) \mathrm{d}t + \sqrt{2 \epsilon_t} \mathrm{d}W_t$, where $b^{\epsilon}_t(x) = b_t(x) + \epsilon_t s_t(x)
    $,
    which corresponds to the conditional density 
    \begin{align}
    P(x_{(k+1)h}|x_{kh}) &= \frac{\exp \big( - \| x_{(k+1)h} - x_{kh} - h b^{\epsilon}_{kh}(x_{kh})\|^2 / (4h\epsilon_{k h}) \big)}{(4 \pi h \epsilon_{k h})^{d/2}}.
    \end{align}
    If we define the conditional density
    \begin{align} \label{eq:p_r_conditional}
        P^{r}(x_{(k+1)h}|x_{kh}) \propto P(x_{(k+1)h}|x_{kh}) \exp(r_{(k+1)h}(x_{(k+1)h})),
    \end{align}
    in the limit $h \to 0$, this conditional density can be written as
    \begin{align}
    P^{r}(x_{(k+1)h}|x_{kh}) &= \frac{\exp \big( - \| x_{(k+1)h} - x_{kh} - h (b^{\epsilon}_{kh}(x_{kh}) + 
    2 \epsilon_{kh}
    \nabla r_{kh}(x_{kh}) )\|^2 / (4h\epsilon_{k h}) \big)}{(4 \pi h \epsilon_{k h})^{d/2}},
    \end{align}
    which is the conditional density for the Euler-Maruyama discretization of the SDE
    \begin{align}
        dX_t = (b^{\epsilon}_t(X_t) + 2 \epsilon_t \nabla r_t(X_t)) \mathrm{d}t + \sqrt{2 \epsilon_t} \mathrm{d}B_t.
    \end{align}
\end{lemma}
\begin{proof}
We use a discrete version of Itô's lemma:
\begin{align} 
\begin{split} \label{eq:update_log_p_ito_2}
    r_{(k+1)h}(x_{(k+1)h}) &\approx r_{kh}(x_{kh}) + h \big( \partial_t r_{kh}(x_{kh}) + %
    \epsilon_{kh}
    \Delta r_{kh}(x_k) 
    \big) \\ &\qquad %
    + \langle \nabla r_{kh}(x_k), x_{(k+1)h} - x_{kh} \rangle + O(h^{3/2}).
\end{split}
\end{align}
Thus,
\begin{align}
\begin{split}
    &P^{r}(x_{(k+1)h}|x_{kh}) \\ &\propto \exp \bigg( - \frac{\|x_{(k+1)h} - x_{kh} - h b^{\epsilon}_{kh}(x_{kh}) 
    \|^2}{4 h \epsilon_{kh}} + \langle \nabla r_{kh}(x_k), x_{(k+1)h} - x_{kh} \rangle + O(h^{3/2}) \bigg) \\ &\propto \exp \bigg( - \frac{\|x_{(k+1)h} - x_{kh} - h (b^{\epsilon}_{kh}(x_{kh}) + 2 \epsilon_{kh} \nabla r_{kh}(x_k))
    \|^2}{4 h \epsilon_{kh}} + O(h^{3/2}) \bigg).
\end{split}
\end{align}    
\end{proof}

\begin{lemma} \label{lem:difference_integrals_application}
Assume that $(\tilde{x}_t)_{t \in [0,1]}$ satisfies the SDE \eqref{eq:tilde_x_sde:reward}.
We have that
\begin{align}
\begin{split}
    \int_t^{t'} \tau 
    \chi_{\tau} \Delta (r \circ X_{\tau,1})(\tilde{x}_{\tau}) \, \mathrm{d}\tau &= \int_t^{t'} \frac{\tau 
    \chi_{\tau}
    }{\sqrt{2 \epsilon_{\tau}}} \nabla (r \circ X_{\tau,1})(\tilde{x}_{\tau}) \cdot \mathrm{d}^{-}W_{\tau} 
    - \int_t^{t'} \frac{\tau \chi_{\tau}}{\sqrt{2 \epsilon_{\tau}}} \nabla (r \circ X_{\tau,1})(\tilde{x}_{\tau}) \cdot \mathrm{d}W_{\tau},
\end{split}
\end{align}
where the forward and backward Itô integrals are defined respectively as
\begin{align}
\int_t^{t'} H_{\tau} \cdot \mathrm{d}W_{\tau}
&\;\coloneqq\;
\lim_{|\pi|\to 0}\;
\sum_{k=0}^{n-1} H_{t_k}\cdot\big(W_{t_{k+1}}-W_{t_k}\big), \\ \int_t^{t'} H_{\tau} \cdot \mathrm{d}^{-}W_{\tau}
&\;\coloneqq\;
\lim_{|\pi|\to 0}\;
\sum_{k=0}^{n-1} H_{t_{k+1}}\cdot\big(W_{t_{k+1}}-W_{t_k}\big).
\end{align}
Here, $(H_s)_{t \leq s \leq t'}$ is a process adapted to the filtration induced by the Brownian motion $W$ such that $\mathbb{E}[\int_t^{t'} \|H_s\|^2 \, ds] < +\infty$, $\pi=\{t=t_0<t_1<\cdots<t_n=t'\}$ is a partition of $[t,t']$ with mesh $|\pi|=\max_k(t_{k+1}-t_k)$, and the limits are $L^2$ limits.
\end{lemma}
\begin{proof}
    By definition of the forward and backward Itô integrals,
    \begin{align}
        \int_0^t H_{\tau}\,d^-W_{\tau} \;-\; \int_0^t H_{\tau}\,dW_{\tau} = \lim_{|\pi|\to 0}\; \sum_{k=0}^{n-1} \big(H_{t_{k+1}}-H_{t_k}\big)\cdot\big(W_{t_{k+1}}-W_{t_k}\big) := [H,W]_t,
    \end{align}
    where $[H,W]_t$ is known as the quadratic variation of $H$ and $W$. 
    Let us set $H_{\tau} = \frac{\tau 
    \chi_{\tau}
    }{\sqrt{2 \epsilon_{\tau}}} \nabla (r \circ X_{\tau,1})(\tilde{x}_{\tau}) = \gamma_{\tau} \nabla (r \circ X_{\tau,1})(\tilde{x}_{\tau})$, where we defined $\gamma_{\tau} = \frac{\tau 
    \chi_{\tau}
    }{\sqrt{2 \epsilon_{\tau}}}$. By Itô's lemma,
    \begin{align}
    \begin{split}
        &d(\gamma_{\tau} \cdot \nabla (r \circ X_{\tau,1}))(\tilde{x}_{\tau}) \\ &= \gamma_t \bigg( \partial_t \nabla (r \circ X_{t,1})(\tilde{x}_{\tau}) + \nabla^2 (r \circ X_{t,1})(\tilde{x}_{\tau}) \cdot \big( \tilde{b}_t + \epsilon_t s_t - \epsilon_t \nabla r_t \big)(\tilde{x}_{\tau}) + \epsilon_t \nabla \cdot \nabla^2 (r \circ X_{t,1})(\tilde{x}_{\tau}) \bigg) \, \mathrm{d}t \\ &\quad + \dot{\gamma}_t \nabla (r \circ X_{t,1})(\tilde{x}_{\tau}) \, \mathrm{d}t + \sqrt{2 \epsilon_t} \gamma_t \nabla^2 (r \circ X_{t,1})(\tilde{x}_{\tau}) dW_t.
    \end{split}    
    \end{align}
    When we simplify the quadratic variation, only the stochastic term survives:
    \begin{align}
    \begin{split}
        [H,W]_t &= \lim_{|\pi|\to 0}\; \sum_{k=0}^{n-1} \sqrt{2 \epsilon_{t_k}} \gamma_{t_k} \langle \nabla^2 (r \circ X_{t_k,1})(\tilde{x}_{t_k}) (W_{t_{k+1}}-W_{t_k}), W_{t_{k+1}}-W_{t_k} \rangle \\ &= \lim_{|\pi|\to 0}\; \sum_{k=0}^{n-1} \sqrt{2 \epsilon_{t_k}} \gamma_{t_k} \mathrm{Tr}\big( \nabla^2 (r \circ X_{t_k,1})(\tilde{x}_{t_k}) \big) (t_{k+1} - t_k) = \int_0^t \sqrt{2 \epsilon_{\tau}} \gamma_{\tau} \Delta (r \circ X_{\tau,1})(\tilde{x}_{\tau}) \, d\tau \\
        &= \int_0^t \tau \chi_{\tau} \Delta (r \circ X_{\tau,1})(\tilde{x}_{\tau}) \, d\tau,
    \end{split}
    \end{align}
    which concludes the proof.
\end{proof}

\begin{lemma} \label{lem:inf_gen_residual}
    The infinitesimal generator $L^{\chi}_t$ defined in \eqref{eq:L_t_f_t} and the HJB residual defined in \eqref{eq:residual_HJB_def} fulfill:
    \begin{align}
    [L^{\chi}_t e^{r_t}](x) = e^{r_t(x)} \mathcal{R}^{\chi}_t(x).
    \end{align}
\end{lemma}
\begin{proof}
Applying the Hopf-Cole transformation, we obtain that
\begin{align} 
\begin{split} \label{eq:e_r_t_evolution_lemma}
    \partial_t e^{r_t(x)} &= e^{r_t(x)} \partial_t r_t(x) \\ &= e^{r_t(x)} \big( \mathcal{R}^{\chi}_t(x) - b_t(x) \cdot \nabla r_t(x) \! - \! |\chi_t| \big( \| \nabla r_t(x)\|^2 \! + \! \Delta r_t(x) \! + \! \nabla r_t(x) \cdot s_t(x) \big) \big).
\end{split}
\end{align}
And observe that
\begin{align}
\begin{split}
    &(b_t(x) + |\chi_t| s_t(x)) \cdot \nabla e^{r_t(x)} + |\chi_t| \Delta e^{r_t(x)} \\ &= e^{r_t(x)} \big( b_t(x) \cdot \nabla r_t(x) \! + \! |\chi_t| \big( \| \nabla r_t(x)\|^2 \! + \! \Delta r_t(x) \! + \! \nabla r_t(x) \cdot s_t(x) \big) \big).
\end{split}
\end{align}
Plugging this into \eqref{eq:e_r_t_evolution_lemma} yields
\begin{align} \label{eq:e_r_t_evolution_lemma_2}
    \partial_t e^{r_t(x)} = e^{r_t(x)} \mathcal{R}^{\chi}_t(x) - (b_t(x) + |\chi_t| s_t(x)) \cdot \nabla e^{r_t(x)} - |\chi_t| \Delta e^{r_t(x)}.
\end{align}
And plugging the expression of the infinitesimal generator in \eqref{eq:L_t_f_t} into \eqref{eq:e_r_t_evolution_lemma_2} yields
\begin{align}
    0 = e^{r_t(x)} \mathcal{R}^{\chi}_t(x) - [L^{\chi}_t e^{r_t}](x).
\end{align}
\end{proof}

\subsection{Flow Map Look-ahead is Optimal for Linear Rewards in a Gaussian Setting} 
\label{app:toy_theory_flowmap_optimal}
In this section, we show that in a simplified Gaussian setting with a linear reward, the flow map is the unique scalar linear look-ahead that minimizes the accumulated variance of the SMC incremental log-weights. A linear reward simply corresponds to a mean shift of the terminal Gaussian distribution, so this setting is tractable and all relevant quantities have closed form solutions.

Formally, let $\rho_{0}(x) = \mathcal N(0, I)$ and $\rho_{data} = \rho_{1}(x) = \mathcal{N}(0, c^2 I)$. We consider the stochastic interpolant
\begin{equation}
    x_t = (1-t)x_0 + tx_1, \qquad t \in [0,1],
\end{equation}
where $x_0 \sim \rho_0, x_1 \sim \rho_1$ are independently drawn. The marginal distribution of $x_t$ is
\begin{equation}
    \rho_t = \mathcal N(0,\sigma_t^2 I_d),
    \qquad
    \sigma_t^2 = (1-t)^2 + c^2 t^2.
\end{equation}
The score is therefore
\begin{equation}
    s_t(x) = \nabla \log \rho_t(x) = -\frac{x}{\sigma_t^2}.
\end{equation}

Using this, the flow matching field can be derived
\begin{equation}
    b_t(x) = \mathbb E[x_1-x_0 \mid x_t=x] = \left(\frac{c^2t - (1 - t)}{\sigma_t^2} \right)x = a_t x,
\end{equation}
where 
\begin{equation}
    a_t = \frac{c^2t-(1-t)}{\sigma_t^2}.
\end{equation}

Equivalently,
\begin{equation}
    a_t = \frac{\dot \sigma_t}{\sigma_t}.
\end{equation}

Given this, we can write the probability-flow ODE as
\begin{equation}
    \dot x_t = b_t(x_t) = a_t x_t.
\end{equation}
Since $a_t = \dot \sigma_t/\sigma_t$, the flow map can be derived in closed form
\begin{equation}
    X_{s,t}(x) = \frac{\sigma_t}{\sigma_s}x.
\end{equation}

In particular, because $\sigma_1=c$,
\begin{equation}
    X_{t,1}(x) = \frac{c}{\sigma_t}x.
\end{equation}

Similarly, the one-step denoiser can be derived as
\begin{equation}
    D_t(x) = \mathbb E[X_1\mid I_t=x] = \frac{c^2t}{\sigma_t^2}x.
\end{equation}

So, the three look-aheads are all linear functions of the form $L_t(x) = \ell_t\cdot x$ where
\begin{equation}
    \ell_t^{\mathrm{Flowmap}} = \frac{c}{\sigma_t},
    \qquad
    \ell_t^{\mathrm{Denoiser}} = \frac{c^2t}{\sigma_t^2},
    \qquad
    \ell_t^{\mathrm{NoLookahead}} = 1.
\end{equation}

\begin{theorem}[Optimality of the flow map for linear rewards]
Let
\begin{equation}
    r(y) = \beta u^\top y, \qquad \|u\|=1,
\end{equation}
for some reward strength $\beta>0$. 
Consider scalar linear look-aheads of the form
\begin{equation}
    L_t(x) = \ell_t x,
\end{equation}
where $\ell_1 = 1$, and define the time-dependent reward as
\begin{equation}
    r_t^\ell(x) = t r(L_t(x)).
\end{equation}
For the default Jarzynski/SMC correction with $\chi_t\equiv 0$, the ideal-case variance of the SMC incremental log-weights is
\begin{equation}
    \mathcal I_\ell
    =
    \beta^2 \int_0^1
    \left[
        \frac{d}{dt}\left(t\sigma_t\ell_t\right)
    \right]^2 dt.
\end{equation}
Furthermore $\mathcal I_\ell \geq \beta^2c^2$, with equality achieved if and only if using the flow map look-ahead. 
\end{theorem}

\begin{proof}
Let $q_t = t\ell_t$, so that
\begin{equation}
    r_t^\ell(x) = \beta q_t u^\top x.
\end{equation}
For $\chi_t\equiv 0$, according to \Cref{prop:jar}, the log-weight dynamics are
\begin{equation}
    \frac{dA_t}{dt} = b_t(x_t)\cdot \nabla r_t^\ell(x_t) +
    \partial_t r_t^\ell(x_t).
\end{equation}
Since
\begin{equation}
    \nabla r_t^\ell(x) = \beta q_t u,
    \qquad
    \partial_t r_t^\ell(x)=\beta \dot q_t u^\top x,
\end{equation}
and $b_t(x)=a_tx$ with $a_t=\dot \sigma_t/\sigma_t$, we have
\begin{align}
    \frac{dA_t}{dt}
    &= \beta a_t q_t u^\top x_t + \beta \dot q_t u^\top x_t \\
    &=
    \beta(a_tq_t + \dot q_t)u^\top x_t \\
    &=
    \beta\left(\frac{\dot\sigma_t}{\sigma_t}q_t + \dot q_t\right)u^\top x_t \\
    &=
    \frac{\beta}{\sigma_t}\frac{d}{dt}(\sigma_t q_t)u^\top x_t.
\end{align}
Define
\begin{equation}
    m_\ell(t) = \sigma_t q_t = t\sigma_t\ell_t.
\end{equation}
Then
\begin{equation}
    \frac{dA_t}{dt}
    =
    \beta\frac{\dot m_\ell(t)}{\sigma_t}u^\top x_t.
\end{equation}

The tilted intermediate density is
\begin{equation}
    \widehat \rho_t^\ell(x)
    \propto
    \rho_t(x)\exp(r_t^\ell(x)).
\end{equation}
Because $r_t^\ell$ is linear and $\rho_t$ is Gaussian, this tilt only changes the mean of $\rho_t$ and leaves the covariance equal to $\sigma_t^2 I_d$. As a result, 
\begin{equation}
    \mathrm{Var}_{\widehat\rho_t^\ell}(u^\top X)=\sigma_t^2.
\end{equation}
Therefore
\begin{align}
    \mathrm{Var}_{\widehat\rho_t^\ell}
    \left[
        \frac{dA_t}{dt}
    \right]
    &=
    \beta^2\frac{(\dot m_\ell(t))^2}{\sigma_t^2}
    \mathrm{Var}_{\widehat\rho_t^\ell}(u^\top X) \\
    &=
    \beta^2(\dot m_\ell(t))^2.
\end{align}
So, the accumulated local variance is
\begin{equation}
    \mathcal I_\ell
    =
    \beta^2\int_0^1(\dot m_\ell(t))^2dt
    =
    \beta^2 \int_0^1
    \left[
        \frac{d}{dt}(t\sigma_t\ell_t)
    \right]^2dt.
\end{equation}

Earlier, we assumed $\ell_1 = 1$, which implies 
\begin{equation}
    m_\ell(0)=0,
    \qquad
    m_\ell(1)=\sigma_1\ell_1=c.
\end{equation}
By Cauchy--Schwarz, we have
\begin{align}
    \int_0^1(\dot m_\ell(t))^2dt
    &\geq
    \left(\int_0^1 \dot m_\ell(t)dt\right)^2  =
    (m_\ell(1)-m_\ell(0))^2 =
    c^2.
\end{align}
Equality holds if and only if $m_\ell'(t)$ is constant almost everywhere. Since $m_\ell(0)=0$ and $m_\ell(1)=c$, this would mean $m_\ell(t)=ct$. Plugging this in to the definition of $m_\ell(t) = t\sigma_t\ell_t$ we have:
\begin{equation}
    \ell_t=\frac{c}{\sigma_t},
\end{equation}
which is exactly the flow map look-ahead $X_{t,1}(x)=\frac{c}{\sigma_t} x$, completing the proof.
\end{proof}

Note that in the proof above, we assume an ideal case where $x_t$ is distributed according to the tilted distribution $\widehat \rho_t^\ell(x) \propto \rho_t(x)\exp(r_t^\ell(x))$ for all $t$. In practice, with finite particles, this is only approximately true. Therefore, the value $\mathcal I_\ell$ should be viewed as an ideal asymptotic variance for the large-particle limit with sufficient resampling operations.
\section{Analyzing the Performance of Test-Time Sampling Algorithms} \label{app:analyzing}

\subsection{Simulating the Dynamics with SMC}
The natural approach to handle the weights $e^{A_t}$ in \Cref{prop:map:tilt} and in \Cref{prop:map:tilt:reward} is sequential Monte Carlo (SMC), which is implemented in \Cref{alg1:inference_time_adaptation}. Let $\mathcal{T} = {(t_k)}_{k=0}^{K}$ be an annealing schedule satisfying $0 = t_0 <\cdots < t_K = 1$.
The SMC sampling procedure starts with \(N\) particles \(\mathbf{X}_0=(X_0^n)_{n\in[N]}\) drawn from the reference, \(X_0^n \sim \rho_0\), and initial weights \(\mathbf{w}_0=(w_0^n)_{n\in[N]}\) with \(w_0^n=1\). For each subsequent iteration \(k\in[K]\), we produce \(\mathbf{X}_k\) and \(\mathbf{w}_k\) by propagating, reweighting, and optionally resampling\footnote{While in \Cref{alg1:inference_time_adaptation} we allow for a number of clones $C$ greater than one, for simplicity, the analysis we perform in this section is with $C=1$.}. In what follows, we use a notation similar to the one of \cite{syed2024optimised}.

\noindent\textbf{Propagate.} Evolve \(\mathbf{X}_{k-1}\) forward with the Markov transition kernel \(M_{t_{k-1},t_k}(x_{k-1}
)\) to obtain \(\mathbf{X}_k=(X_k^n)_{n\in[N]}\), where
\begin{align} \label{eq:M_t}
X_k^n \sim M_{t_{k-1},t_k}(X_{k-1}^n
).
\end{align}
For the dynamics of \Cref{prop:map:tilt} and \Cref{prop:map:tilt:reward}, the Markov transition kernels are the Euler-Maruyama updates for the SDEs \eqref{eq:position_SDE} and \eqref{eq:sde:mod}, respectively:
\begin{align}
    \begin{split}
    X_k^{n} &= X_{k-1}^{n} + [b_{t_{k-1}}(X_{k-1}^{n}) \! + \! \epsilon_{t_{k-1}} (s_{t_{k-1}}(X_{k-1}^{n}) \! + \!
        \nabla r_{t_{k-1}}(X_{k-1}^{n}))] (t_{k} \! - \! t_{k-1}) \\ &\quad + \sqrt{2 \epsilon_k (t_{k} \! - \! t_{k-1})} \xi_{k-1}^{n}, \quad \xi_{k-1}^{n} \sim N(0,\mathrm{I}), 
    \end{split} \\
    \begin{split}
    X_k^{n} &= X_{k-1}^{n} + [b_{t_{k-1}}(X_{k-1}^{n}) \! + \! \chi_{t_{k-1}} \nabla r_{t_{k-1}}(X_{k-1}^{n}) \! + \! \epsilon_{t_{k-1}} (s_{t_{k-1}}(X_{k-1}^{n}) \! + \!
        \nabla r_{t_{k-1}}(X_{k-1}^{n}))] (t_{k} \! - \! t_{k-1}) \\ &\quad + \sqrt{2 \epsilon_k (t_{k} \! - \! t_{k-1})} \xi_{k-1}^{n}, \quad \xi_{k-1}^{n} \sim N(0,\mathrm{I}),
    \end{split}
\end{align}

\noindent\textbf{Reweight.} Update the weights using the incremental weight function \(g_{t_{k-1},t_k}(x_{k-1},x_k)\):
\begin{align} \label{eq:w_k_g_t}
\mathbf{w}_k=(w_k^n)_{n\in[N]},\qquad
w_k^n = w_{k-1}^n\, g_{t_{k-1},t_k}(X_{k-1}^n,X_k^n).
\end{align}
The incremental weight functions $g_{t_{k-1},t_k}(X_{k-1}^n,X_k^n)$ and weights $w_k^n$ can be computed using the framework of \Cref{subsec:solving_numerically}:
\begin{align}
    g_{t_{k-1},t_k}(X_{k-1}^n,X_k^n) = \exp \big( A^n_k - A^n_{k-1} \big), \qquad w_k^n = \exp \big( A^n_k \big).
\end{align}
Here, the difference $A^n_k - A^n_{k-1}$ is computed for each particle using one of the approaches described in \Cref{subsec:solving_numerically}. 

\noindent\textbf{Resample (optional).} On a (possibly random) subset of iterations \(\mathcal{R}\subseteq [K]\) determined by a criterion depending on \((\mathbf{X}_k,\mathbf{w}_k)\), apply a resampling step
\[
\mathbf{X}_k \leftarrow \mathrm{resample}(\mathbf{X}_k,\mathbf{w}_k),
\]
to stabilize \(\mathbf{w}_k\) and favor propagation of particles with higher relative weight. Concretely, set \(X_k^n \leftarrow X_k^{a_k^n}\), where \(a_k=(a_k^n)_{n\in[N]}\) is a random ancestor index vector with \(a_k^n\in[N]\) and
\[
\mathbb{P}\!\left(a_k^n=m \,\middle|\, \mathbf{w}_k\right)
= \frac{w_k^m}{\sum_{j\in[N]} w_k^j},\qquad m\in[N].
\]
After resampling, reset the weights via \(w_k^n \leftarrow 1\) for all \(n\). See \citep{chopin2022onresampling} for annealed-SMC-specific resampling schemes, and \citep{dai2020aninvitation} for a recent review of SMC samplers.

The SMC procedure yields an unbiased estimate of the expectation $\int_{\mathbb{R}^d} h(x) \hat{\rho}_{t_k}(x) dx$ for any test function $h$ and any $k \in [K]$. Namely, if we let $\bar{\rho}_t$ be the unnormalized density as defined in \Cref{app:proof:generic}, recall that $\mathcal{R} \cap [k]$ denotes the subset of iterations where a resampling step happens, and for $r \in \mathcal{R}$ we let $w_r$ be the weight prior to resetting to 1,
we have that
\begin{align}
    \int_{\mathbb{R}^d} h(x) \bar{\rho}_{t_k}(x) dx = \mathbb{E}\bigg[ \bigg( \prod_{r \in \mathcal{R} \cap [k-1]} \frac{1}{N} \sum_{n=1}^{N} w^n_{r} \bigg) \frac{1}{N} \sum_{n=1}^{N} w^n_{k} h(X^n_{k})  \bigg],
\end{align}
Setting $h \equiv 1$ and recalling that $\hat{\rho}_{t_k}(x) = \bar{\rho}_{t_k}(x) / \int_{\mathbb{R}^d} \bar{\rho}_{t_k}(x) dx$, we obtain that
\begin{align} \label{eq:Z_k_SMC}
    \int_{\mathbb{R}^d} h(x) \hat{\rho}_{t_k}(x) dx = \frac{\mathbb{E}\bigg[ \hat{Z}^{(k)}_{\mathrm{SMC}} \frac{\frac{1}{N} \sum_{n=1}^{N} w^n_{t_k} h(X^n_{t_k})}{\frac{1}{N} \sum_{n=1}^{N} w^n_{t_k}}
    \bigg]}{\mathbb{E}\big[ \hat{Z}^{(k)}_{\mathrm{SMC}}
    \big]}, \quad \hat{Z}^{(k)}_{\mathrm{SMC}} = \bigg( \prod_{r \in \mathcal{R} \cap [k-1]} \frac{1}{N} \sum_{n=1}^{N} w^n_{r} \bigg) \times \bigg( \frac{1}{N} \sum_{n=1}^{N} w^n_{k} \bigg).
\end{align}
If we set $k = K$, we obtain that
\begin{align}
    \int_{\mathbb{R}^d} h(x) \hat{\rho}_{1}(x) dx = \frac{\mathbb{E}\bigg[ \hat{Z}_{\mathrm{SMC}} \frac{\frac{1}{N} \sum_{n=1}^{N} w^n_{K} h(X^n_{K})}{\frac{1}{N} \sum_{n=1}^{N} w^n_{K}} \bigg]}{\mathbb{E}\big[ \hat{Z}_{\mathrm{SMC}} \big]}, \qquad \text{where } \hat{Z}_{\mathrm{SMC}} = \prod_{r \in \mathcal{R}} \frac{1}{N} \sum_{n=1}^{N} w^n_{r}.
\end{align}
$\hat{Z}_{\mathrm{SMC}}$ is known as the SMC normalization constant. Observe that $\mathbb{E}[\hat{Z}_{\mathrm{SMC}}] = \int_{\mathbb{R}^d} \bar{\rho}_{1}(x) dx := Z$. Thus, low $\mathrm{Var}[\hat{Z}_{\mathrm{SMC}}/Z]$ is a proxy for good performance of the SMC procedure. In \Cref{subsec:variance_SMC} we reproduce a result by \cite{syed2024optimised} that expresses $\mathrm{Var}[\hat{Z}_{\mathrm{SMC}}/Z]$ in terms of the parameters of the SMC procedure, using the concept of total discrepancy from \Cref{subsec:discrepancy}.

\subsection{The Incremental and Total Discrepancies} \label{subsec:discrepancy}

Following \cite{syed2024optimised}, we define the normalized incremental weight function $G_{t,t'}$ as
\begin{align} \label{eq:normalized_incremental}
    G_{t,t'}(x,x') = \frac{g_{t,t'}(x,x')}{
    \mathbb{E}_{(X,X') \sim \hat{\rho}_{t} \otimes M_{t,t'}}[g_{t,t'}(X,X')]
    }
\end{align}
where $g_{t,t'}$ is the unnormalized incremental weight function defined in \eqref{eq:w_k_g_t} and $M_{t,t'}$ is the Markov transition kernel defined in \eqref{eq:M_t}.

Given $G_{t,t'}$ from time $t$ to time $t'$, the \textit{incremental discrepancy} $D(t,t')$ is defined as 
\begin{align}
D(t,t') = \log \big( 1 + \mathrm{Var}_{(X,X') \sim \hat{\rho}_t \otimes M_{t,t'}}\big(G_{t,t'}(X,X') \big) \big). 
\end{align}
Given a sequence of timesteps $\mathcal{T} = (t_{k})_{k=0}^{K}$ with $t_0 = 0$, $t_K = 1$, tor \( k \le k' \), define \( D(\mathcal{T}, t_k, t_{k'}) \) as the \textit{accumulated discrepancy} between
iterations \( k \) and \( k' \)
and \( D(\mathcal{T}) \) as the \textit{total discrepancy}:
\begin{align} \label{eq:cumulative_total_discrepancy}
D(\mathcal{T}, t_k, t_{k'}) = \sum_{k'' = k+1}^{k'} D(t_{k''-1}, t_{k''}), 
\qquad 
D(\mathcal{T}) = D(\mathcal{T}, 0, 1).
\end{align}

Observe that the incremental discrepancy can be expressed in terms of the first and second moments of $g_{t,t'}(X,X')$:
\begin{align}
\begin{split} \label{eq:D_expression}
    D(t,t') &= \log \bigg( 1 + \mathrm{Var}_{(X,X') \sim \hat{\rho}_t \otimes M_{t,t'}}\bigg[ \frac{g_{t,t'}(X,X')}{\mathbb{E}_{(X'',X''') \sim \hat{\rho}_{t} \otimes M_{t,t'}}[g_{t,t'}(X'',X''')]} \bigg] \bigg) \\ &= \log \bigg( 1 + \frac{\mathbb{E}_{(X,X') \sim \hat{\rho}_t \otimes M_{t,t'}}\big[ g_{t,t'}(X,X')^2 \big]}{\mathbb{E}_{(X'',X''') \sim \hat{\rho}_{t} \otimes M_{t,t'}}[g_{t,t'}(X'',X''')]^2} - \frac{\mathbb{E}_{(X,X') \sim \hat{\rho}_t \otimes M_{t,t'}}\big[ g_{t,t'}(X,X') \big]^2}{\mathbb{E}_{(X'',X''') \sim \hat{\rho}_{t} \otimes M_{t,t'}}[g_{t,t'}(X'',X''')]^2} \bigg) \\ &= \log \mathbb{E}_{(X,X') \sim \hat{\rho}_t \otimes M_{t,t'}}\big[ g_{t,t'}(X,X')^2 \big] - 2 \log \mathbb{E}_{(X,X') \sim \hat{\rho}_t \otimes M_{t,t'}}\big[ g_{t,t'}(X,X') \big]
\end{split}
\end{align}

We want to obtain a consistent estimator for $D(t,t')$. Following Sec 5.1 of \cite{syed2024optimised}, if we are using a single SMC run with $N$ particles, and we let $g^n_k = g_{t_{k-1},t_k}(X^n_{k-1},X^n_k)$, then the following estimator is consistent:
\begin{align} \label{eq:hatD_k}
    \hat{D}_k = \log \hat{g}_{k,2} - 2 \log \hat{g}_{k,1} - \log \hat{g}_{k,0}, \quad \text{where for $i \in \{0,1,2\}$,} \quad \hat{g}_{k,i} = \sum_{n \in [N]} w^n_{k-1} \big( g^n_k \big)^i. 
\end{align}
To get a consistent estimator using $N_{R}$ SMC runs, each with $N$ particles, we compute 
the normalization constant $\hat{Z}^{(k,j)}_{\mathrm{SMC}}$ in \eqref{eq:Z_k_SMC} for each SMC run $j = 1:N_{R}$, and define $\hat{D}_k$ as in \eqref{eq:hatD_k}, but where $\hat{g}_{k,i}$ takes the form
\begin{align}
    \hat{g}_{k,i} = 
    \sum_{j=1}^{N_R} \hat{Z}^{(k,j)}_{\mathrm{SMC}} \frac{\sum_{n \in [N]} w^{(n,j)}_{k-1} \big( g^{(n,j)}_{k} \big)^i}{\sum_{n \in [N]} w^{(n,j)}_{k-1}},
    \qquad \text{where} \quad g^{(n,j)}_{k} = g_{t_{k-1},t_k}(X^{(n,j)}_{k-1},X^{(n,j)}_k),
\end{align}
and $X^{(n,j)}_k$, $w^{(n,j)}_k$ is the $n$-th particle and weight of $j$-th SMC run at iteration $k$.

\subsection{The Variance of the SMC Normalization Constant} \label{subsec:variance_SMC}

The total discrepancy defined in equation \eqref{eq:cumulative_total_discrepancy} is related to the variance of the SMC normalization constant $Z_{\mathrm{SMC}}$ via the following result, which was proven by \cite{syed2024optimised} as a generalization of a result of \cite{dai2020aninvitation}: 

\begin{theorem}[Theorem 1, \cite{syed2024optimised}]
Suppose that the following assumptions on the normalized incremental weights
\(G_k^n = G_{t_{k-1},t_k}(X_{k-1}^n,X_k^n)\) defined in \eqref{eq:normalized_incremental} hold:
\begin{itemize}[left=5pt]
\item \emph{Assumption 1 (Integrability).}
For all \(n \in [N]\), \(k \in [K]\), \(G_k^n\) has finite variance with respect to
\(\hat{\rho}_{t_k} \otimes M_{t_{k-1},t_k}\).

\item \emph{Assumption 2 (Temporal indep.).}
For \(n \in [N]\), \((G_k^n)_{t \in [T]}\) are independent.

\item \emph{Assumption 3 (Particle indep.).}
For \(k \in [K]\), \((G_k^n)_{n \in [N]}\) are independent.

\item \emph{Assumption 4 (Efficient local moves).}
For each \(n \in [N]\) and \(k \in [k]\),
\[
G_k^n \stackrel{d}{=} G_{t_{k-1},t_k}(X_{k-1},X_k),
\qquad (X_{k-1},X_k) \sim \pi_{t_{k-1}} \otimes M_{t_{k-1},t_k}.
\]
\end{itemize}

Assume also that \( N > 1 \), \( D(\mathcal{T}) > 0 \).
For every resample schedule \( \mathcal{T}_R = (t_r)_{r=0}^R \), there exists a unique 
\( 1 \le R_{\mathrm{eff}} \le \mathbb{E}[R] \) such that
\begin{align} \label{eq:var_SMC}
\operatorname{Var} \left( \frac{\hat Z_{\mathrm{SMC}}}{Z} \right)
= \frac{1}{N} \left( \exp\!\left( \frac{D(\mathcal{T})}{R_{\mathrm{eff}}} \right) - 1 \right) R_{\mathrm{eff}} - 1.
\end{align}
Moreover, \( R_{\mathrm{eff}} = 1 \) if and only if 
\( D(\mathcal{T}, t_{r-1}, t_r) \stackrel{\mathrm{a.s.}}{=} D(\mathcal{T}) \) for some \( r \in [R] \), 
and \( R_{\mathrm{eff}} = \mathbb{E}[R] \) if and only if \( R \) is a.s.\ constant and 
\( D(\mathcal{T}, t_{r-1}, t_r) \stackrel{\mathrm{a.s.}}{=} D(\mathcal{T})/R \) for some \( r \in [R] \).
\end{theorem}

\begin{remark}
    As remarked by \cite{syed2024optimised}, Assumptions 1--4 constitute an idealized model similar to the one considered by other works in the area \citep{grosse2013annealing,dai2020aninvitation}. While Assumption 1 is weak, Assumptions 2--4 are not. Assumption 3 only holds when no resampling is performed, and Assumptions 2--4 only hold (approximately) when a number of MCMC steps are interleaved with the SMC updates. However, \citet[Sec.~6.1]{syed2024optimised} show that empirically, the scaling \eqref{eq:var_SMC} is consistent with empirical observation.
\end{remark}

\subsection{Optimizing the Annealing Schedule to Minimize the Total Discrepancy}

As defined in \eqref{eq:cumulative_total_discrepancy}, the total discrepancy depends not only on the continuous time dynamics for positions and weights, but also on the specific annealing schedule $\mathcal{T} = {(t_k)}_{k=0}^{K}$. For a fixed $K$, it is possible to characterize and find the annealing schedule that minimizes the total discrepancy.

Under technical regularity assumptions (see Sec 4.1 of \cite{syed2024optimised}), the incremental discrepancy admits the asymptotic expansion
\[
G_{t, t + \Delta t} = 1 + S_t \, \Delta t + o(\Delta t),
\]
and hence the local changes and variance of the incremental discrepancy 
\( G_{t, t + \Delta t} \) are encoded in \( S_t \) and its variance 
\(\delta(t)\), defined as
\[
S_t 
= 
\left|\frac{\partial}{\partial t'} G_{t, t'}\right|_{t' = t},
\qquad
\delta(t) = \operatorname{Var}_{t, t}[S_t].
\]
Using this expansion, we can expand the incremental discrepancy as follows:
\begin{align} \label{eq:incremental_discrepancy_expansion}
D(t,t + \Delta t) = \delta(t) \Delta t + O(\Delta t^3).
\end{align}

\paragraph{Scheduler generators} A \emph{schedule generator} is a continuously twice-differentiable function 
\(\varphi : [0,1] \to [0,1]\) such that 
\(\varphi(0) = 0\), \(\varphi(1) = 1\), and 
\(\dot{\varphi}(u) = \frac{d}{du} \varphi(u) > 0\).
Given \( K \in \mathbb{N} \), \(\varphi\) generates an annealing schedule 
\( \mathcal{T} = (t_k)_{k=0}^K \) where 
\[
t_{k} = \varphi(u_k), \qquad u_k = \frac{k}{K}.
\]
In the following, without loss of generality, we restrict our attention to schedules generated by schedule generator.

By the mean value theorem, we have
\[
t_k - t_{k-1} \approx \frac{\dot{\varphi}(u_k)}{K}.
\]
Combining this with equation \eqref{eq:incremental_discrepancy_expansion}, we obtain
\begin{align} \label{eq:D_t_k_m_1_t_k}
D(t_{k-1}, t_k) \approx \frac{\delta(\varphi(u_k)) \, \dot{\varphi}(u_k)^{2}}{K^{2}}.
\end{align}
By summing over \( k \) and using Riemann approximations, we can approximate 
\( D(\mathcal{T}, t_k, t_{k'}) \) and \( D(\mathcal{T}) \) in terms of \( E(\varphi, u_{t_k}, u_{t_{k'}}) \) and 
\( E(\varphi) \), defined as the integral of \( \delta(\varphi(u)) \, \dot{\varphi}(u)^{2} \),
\[
E(\varphi, u, u') = \int_{u}^{u'} \delta(\varphi(v)) \, \dot{\varphi}(v)^{2} \, dv,
\qquad
E(\varphi) = E(\varphi, 0, 1).
\]

\begin{proposition}[Proposition 1, \cite{syed2024optimised}]
Suppose Assumptions~5 to~8 hold. There exists \( C_D(\varphi) > 0 \) such that,
for \( k, k' \in [K] \),
\[
\left| D(\mathcal{T}, t_{k}, t_{k'}) - \frac{1}{K} E(\varphi, u_{k}, u_{k'}) \right|
\le \frac{C_D(\varphi) \, |t_{k'} - t_k|}{T^{3}}.
\]
\end{proposition}

An immediate consequence of Proposition~1 is that, in the dense schedule limit as 
\( K \to \infty \), the total discrepancy \( D(\mathcal{T}) \) is asymptotically equivalent to 
\(\frac{E(\varphi)}{K}\). Hence, for a fixed $K$, optimizing $D(\mathcal{T})$ with respect to $\mathcal{T}$ is asymptotically equivalent to the following problem: 
\begin{align}
    \min_{\varphi : [0,1] \to [0,1]} \int_0^1 \delta(\varphi(u)) \, \dot{\varphi}(u)^{2} \, du, \quad \text{s.t.} \quad \int_0^1 \dot{\varphi}(u) \, du = 1.
\end{align}
Jensen's inequality implies that
\begin{align}
    \int_0^1 \delta(\varphi(u)) \, \dot{\varphi}(u)^{2} \, du \geq \bigg( \int_0^1 \sqrt{\delta(\varphi(u))} \, \dot{\varphi}(u) \, du \bigg)^{2},
\end{align}
with equality if and only if there exists a constant $\Lambda > 0$ such that $\sqrt{\delta(\varphi^{\star}(u))} \, \dot{\varphi^{\star}}(u) = \Lambda$ for $u$ a.e. in $[0,1]$. Defining 
\begin{align}
    \Lambda(t) = \int_0^t \sqrt{\delta(u)} \, du, 
\end{align}
by the chain rule, we have equivalently that
\begin{align} \label{eq:derivative_Lambda}
\Lambda = \Lambda'(\varphi^{\star}(t)) \, \dot{\varphi^{\star}}(t) = \frac{d}{dt} \Lambda(\varphi^{\star}(t)) \implies \Lambda(\varphi^{\star}(t)) = \Lambda t \implies \varphi^{\star}(t) = \Lambda^{-1}(\Lambda t). 
\end{align}
Setting $t = 1$ in $\Lambda(\varphi^{\star}(t)) = \Lambda t$ also implies that $\Lambda = \Lambda(\varphi^{\star}(1)) = \Lambda(1) = \int_0^1 \sqrt{\delta(u)} \, du$. \cite{syed2024optimised} refer to $\Lambda(t)$ and $\Lambda$ as the \emph{local barrier} and the \emph{global barrier} associated to the SMC algorithm. We refer to $\Lambda$ as the \textbf{thermodynamic length} associated to the algorithm.

A change of the integration variable implies that for any schedule generator $\varphi$,
\begin{align}
\begin{split} \label{eq:thermodynamic_length_def}
    &\Lambda(\varphi(t)) = \int_0^{\varphi(t)} \sqrt{\delta(u)} \, du = \int_0^{t} \sqrt{\delta(\varphi(u))} \dot{\varphi}(u) \, du, \\
    &\implies \Lambda(t_k) = \Lambda(\varphi(u_k)) = \int_0^{u_k} \sqrt{\delta(\varphi(u))} \dot{\varphi}(u) \, du \approx \sum_{k'=1}^{k} \frac{\sqrt{\delta(\varphi(u_k))} \, \dot{\varphi}(u_k)}{K} = \sum_{k'=1}^{k} \sqrt{D(t_{k-1}, t_k)}, \\
    &\qquad\quad \Lambda = \int_0^{1} \sqrt{\delta(\varphi(u))} \dot{\varphi}(u) \, du \approx \sum_{k=1}^{K} \sqrt{D(t_{k-1}, t_k)}.
\end{split}
\end{align}
where the last equality holds by \eqref{eq:D_t_k_m_1_t_k}. 
This allows us to approximate the local and global barriers using \eqref{eq:D_expression} to compute the incremental discrepancy $D(t_{k-1}, t_k)$. Once we have an estimate $\hat{\Lambda}$ of the barrier, \cite{syed2024optimised} propose to iteratively refine the annealing schedule by resetting $t_k \gets \hat{\Lambda}^{-1}(\hat{\Lambda} k/K)$. 

Observe that given the quantities $\sum_{k'=1}^{k} D(t_{k'-1}, t_{k'})$ and $\sum_{k'=1}^{k} \sqrt{D(t_{k'-1}, t_{k'})}$, we have that
\begin{align}
    \sum_{k'=1}^{k} D(t_{k'-1}, t_{k'}) =  D(\mathcal{T}, 0, t_{k'}) \approx \frac{1}{K} E(\varphi, u_{0}, u_{k'}) \geq \frac{1}{K} \Lambda(t_k)^2 \approx \frac{1}{K} \bigg( \sum_{k'=1}^{k} \sqrt{D(t_{k'-1}, t_{k'})} \bigg)^2,
\end{align}
Thus, the quantity
\begin{align}
    \frac{\bigg( \sum_{k'=1}^{k} \sqrt{D(t_{k'-1}, t_{k'})} \bigg)^2}{K \sum_{k'=1}^{k} D(t_{k'-1}, t_{k'})}
\end{align}
should fall within $[0,1]$ and close to 1 when the annealing schedule $\mathcal{T}$ is close to the optimal one.

\section{Implementation Details} \label{app:algorithms}
The complete pseudocode of FMTT is given in \Cref{alg1:inference_time_adaptation}.  
\begin{algorithm}[h]
\caption{Inference-time adaptation of flow maps}
\label{alg1:inference_time_adaptation}
\begin{algorithmic}[1] %
\State \textbf{Input}: $\#$ simulation steps $K$, $\#$ resampling steps $R$, $\#$ particles $N$, $\#$ particle clones $C$, time sequence $(t_k)_{k=0:K}$, sequences $(\epsilon_k)_{k=0:K}$ and $(\chi_k)_{k=0:K}$
\State \textbf{for} $i = 1:N$, initialize $x^0_i = N(0,\mathrm{I})$ i.i.d. Let $X^0 = (x^0_i)_{i=1:N}$.
\State Clone the particles: $\bar{X}^0 = (x^0_{ij})_{i=1:N,j=1:C}$, where $(x^0_{ij})^{j=1:C}$ are equal copies of $x^0_i$.
\State \textbf{if} \textcolor{red}{Sampling} \textbf{then} \textbf{for} $i = 1:N$ and $j=1:C$, initialize $A^{0}_{ij} = 0$.
\For{$k = 0:K-1$}
    \State $\Delta t \gets t_{k+1} - t_k$
    \For{$i = 1:N$ and $j=1:C$} \Comment{Update particles}
        \State $x^{k+1}_{ij} = x^{k}_{ij} + [v_{t_k,t_k}(x^{k}_{ij}) \! + \! 
        \chi_k
        \nabla r_{t_k}(x^{k}_{ij}) + \epsilon_k (s_{t_k}(x^{k}_{ij}) \! + \!
        \nabla r_{t_k}(x^{k}_{ij}))] \Delta t + \sqrt{2 \epsilon_k \Delta t} \xi^{k}_{ij}, \quad \xi^{k}_{ij} \sim N(0,\mathrm{I})$.
        \If{\textcolor{red}{Sampling}}
            \State 
            Compute $A^{k+1}_{ij}$ from $A^{k}_{ij}$ using approach \emph{(i)} \eqref{eq:A_k_Laplacian}, approach \emph{(ii)} \eqref{eq:A_k_ito} or approach \emph{(iii)} \eqref{eq:approach_3_positive}-\eqref{eq:approach_3_negative} from \Cref{subsec:solving_numerically}. Note that simplifications occur when $\chi_k \equiv 0$ (\Cref{rem:chi_zero}).
        \EndIf
    \EndFor
    \If{$k = 0 \ (\mathrm{mod} \ K/R)$ and $k > 0$} \Comment{Resample / select particles}
        \If{\textcolor{red}{Sampling}}
        \State Define probabilities $p^k = \mathrm{softmax}(A^{k}) = \big(\exp(A^{k}_{ij})/\sum_{i'j'}\exp(A^{k}_{i'j'}) \big)_{i'=1:N,j'=1:C}$
            \State Resample $X^k = (x^k_i)_{i=1:N} \sim \sum_{i'=1}^{n} \sum_{j'=1}^{C} p^k_{i'j'} \delta_{x^k_{i'j'}}$ i.i.d., or using Quasi-Monte Carlo
        \State Set $A^k_{ij} = 0$.
        \ElsIf{\textcolor{blue}{Searching}}
        \State Select $X^k = (x^k_i)_{i=1:N}$ as the top-$n$ samples among $\bar{X}^k$ with respect to $r_{t_k}(x^k_{ij})$.
        \EndIf
        \State Clone the particles: $\bar{X}^{k} \gets (x^{k}_{ij})_{i=1:N}^{j=1:C}$, where $(x^k_{ij})^{j=1:C}$ are equal copies of $x^k_i$.
    \EndIf
\EndFor
\State \Return $x$
\end{algorithmic}
\end{algorithm}

\section{Additional Experiments}

\subsection{Toy 2D Experiments}
\label{app:toy_2d_experiments}
\begin{figure}[t]
    \centering
    \includegraphics[width=\linewidth]{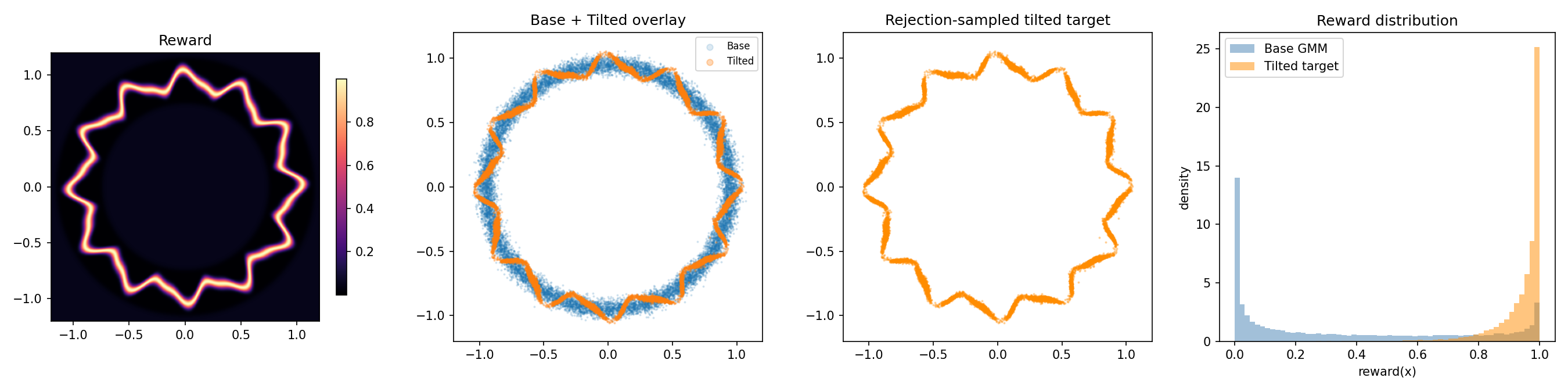}
    \caption{\textbf{Tilting a 2D toy distribution}. We train a diffusion model on the ring-shaped distribution, and choose a reward that tilts it toward the target zigzag shape. Running SMC with 500 steps, no-lookahead and denoiser-lookahead fail to recover the target, while flow map- lookahead can effectively capture it. See \Cref{app:toy_2d_experiments} for details.\looseness=-1}
    \label{fig:toy_2d_reward_analysis}
\end{figure}

In this section, we provide additional experiments for our 2D toy data setting. We consider a ring-shaped base distribution and define a reward function $r$ that has a high value for samples within the base distribution that lie on a zigzag-shaped pattern, and low values elsewhere. The reward along with the base and tilted distributions can be seen in \Cref{fig:toy_2d_reward_analysis}. We choose this reward since it mimics the behavior of complex VLM-based rewards, due to the fact that its values are relatively flat away from the data manifold resulting in unhelpful gradients. 

\begin{figure}[h]
    \centering

    \begin{subfigure}{\linewidth}
        \centering
        \includegraphics[width=0.8\linewidth]{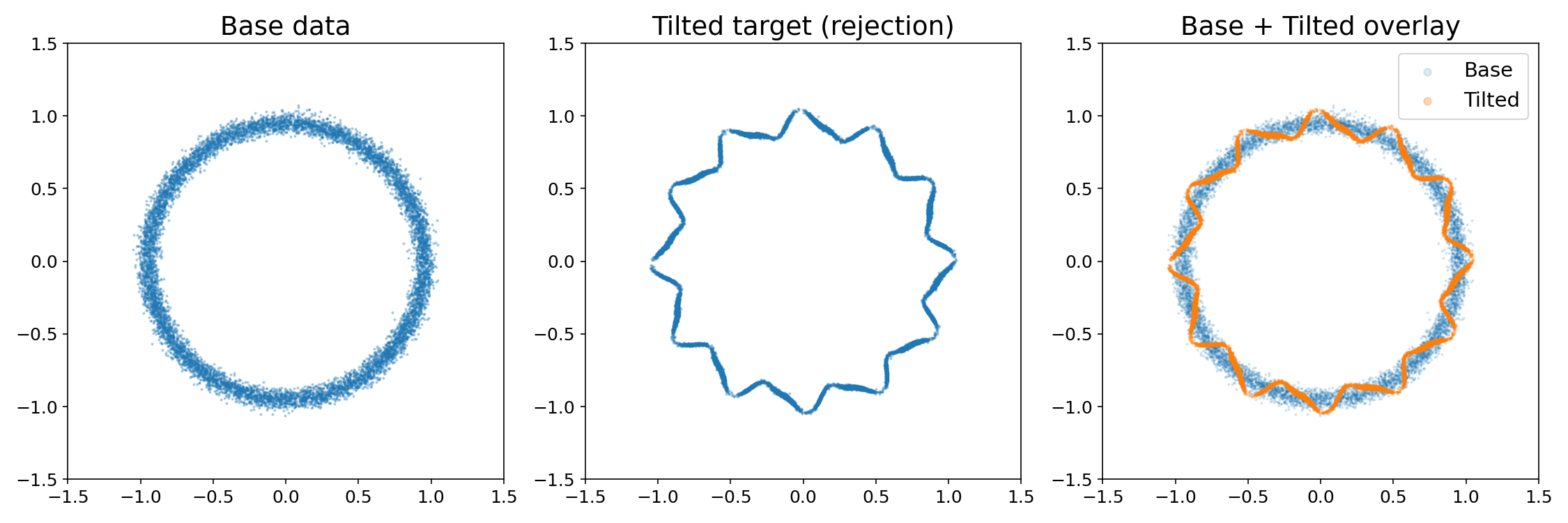}
        \caption{Base and reward-tilted distributions.}
        \label{fig:toy_2d_shared}
    \end{subfigure}

    \vspace{1mm}

    \begin{subfigure}{0.48\linewidth}
        \centering
        \includegraphics[width=\linewidth]{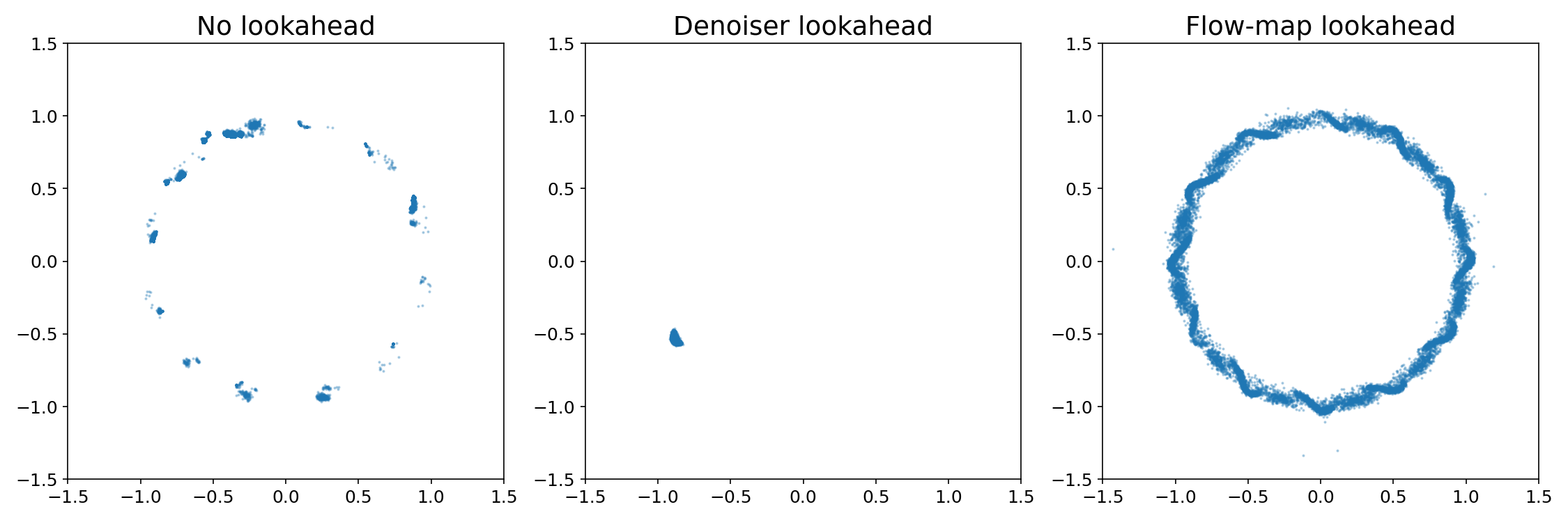}
        \caption{50 SMC steps}
        \label{fig:toy_2d_smc50}
    \end{subfigure}
    \hfill
    \begin{subfigure}{0.48\linewidth}
        \centering
        \includegraphics[width=\linewidth]{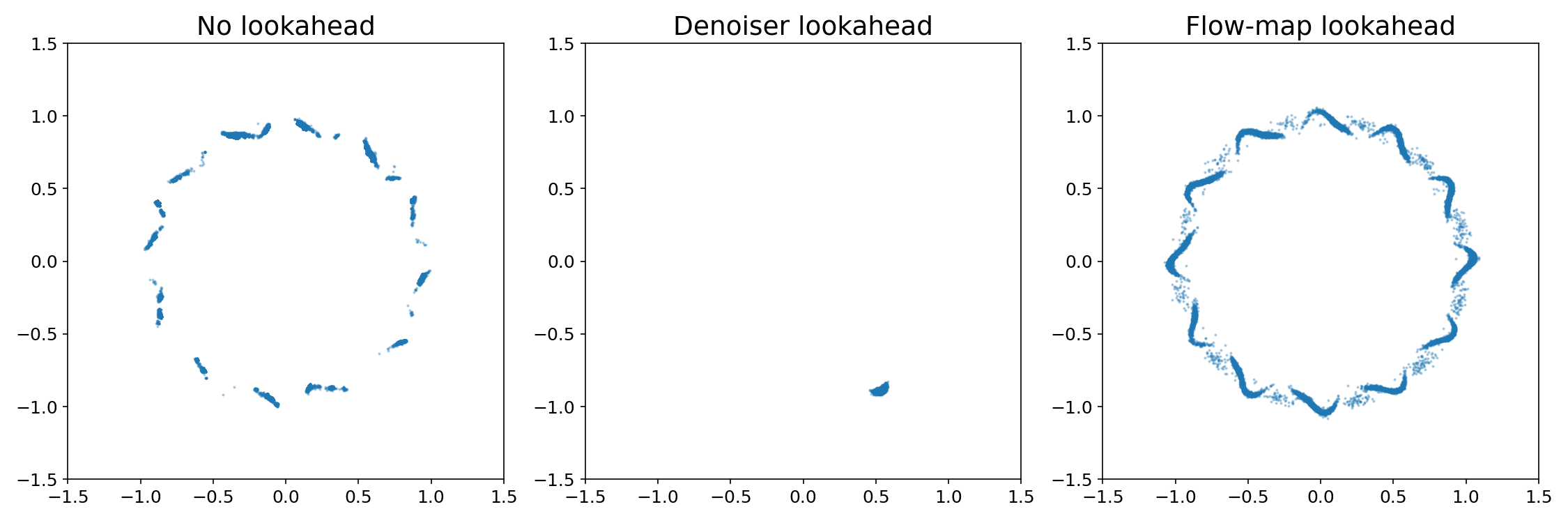}
        \caption{100 SMC steps}
        \label{fig:toy_2d_smc100}
    \end{subfigure}

    \vspace{1mm}

    \begin{subfigure}{0.48\linewidth}
        \centering
        \includegraphics[width=\linewidth]{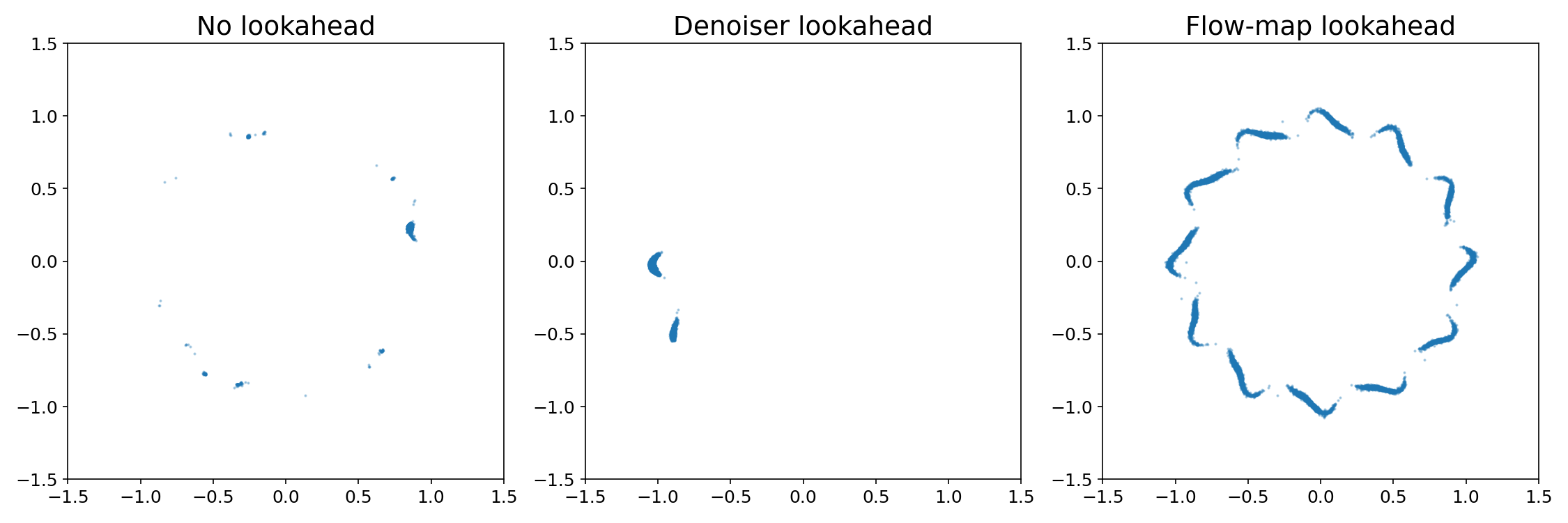}
        \caption{200 SMC steps}
        \label{fig:toy_2d_smc200}
    \end{subfigure}
    \hfill
    \begin{subfigure}{0.48\linewidth}
        \centering
        \includegraphics[width=\linewidth]{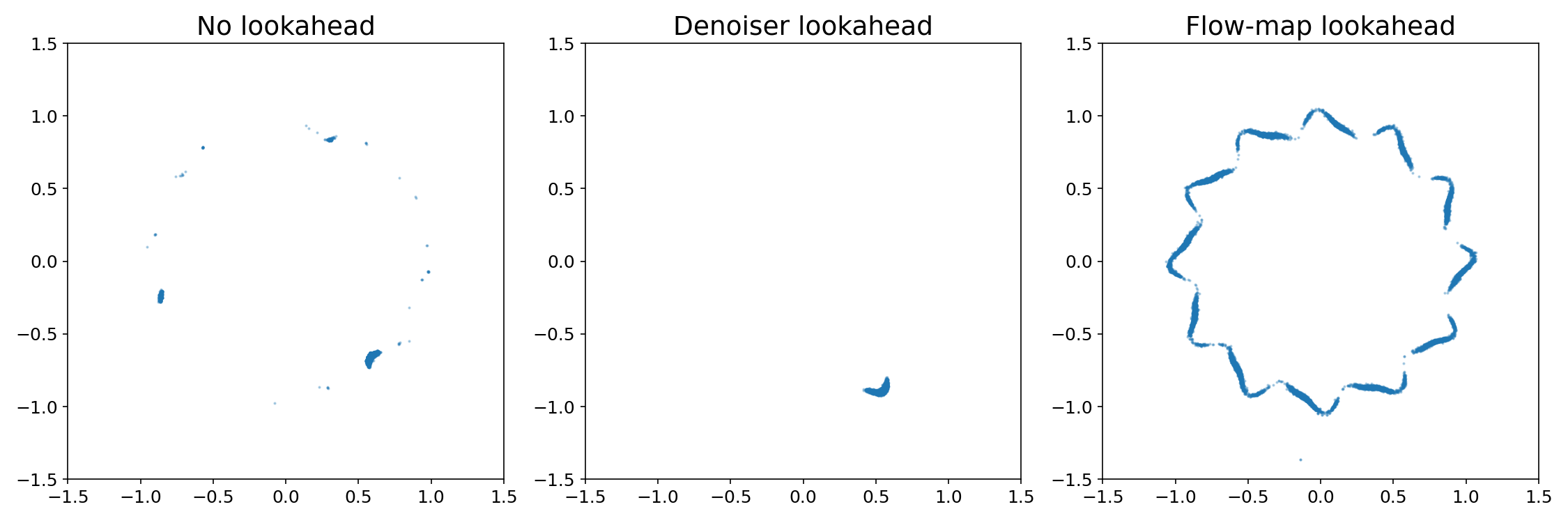}
        \caption{500 SMC steps}
        \label{fig:toy_2d_smc500}
    \end{subfigure}

    \caption{\textbf{Tilting a 2D toy distribution.}
    We run SMC with varying numbers of steps. Across step counts, no-lookahead and denoiser-lookahead fail to recover the target, while flow map-lookahead captures it much more effectively.\looseness=-1}
    \label{fig:toy_2d_all_runs}
\end{figure}

To model the base distribution, we train a simple continuous-time flow matching model to learn the velocity field, followed by distilling into a flow map. We train both models for 100,000 steps with a batch size of 8092. 
After training, we run SMC algorithm defined by \Cref{eq:position_SDE,eq:sde:A} for 50, 100, 200, and 500 sampling steps with 10,000 particles, and perform resampling when the effective sample size falls beneath $0.9$ using a stochasticity factor of $\epsilon_t = 0.1 \times (1 - t)$. 
We calculate Maximum Mean Discrepancy (MMD) and Sliced Wasserstein-2 (SW2) against samples obtained via rejection sampling, and summarize the results in \Cref{tab:toy_2d_smc_sweep}. Qualitative results can be seen in \Cref{fig:toy_2d_all_runs}.  

As can be seen from the results, both naive- and denoiser- look-aheads fail to capture the true tilted distribution, with their outputs degenerating into small blob-like subsets. This is also reflected quantitatively: despite achieving high mean rewards, these methods have much larger MMD and SW2 distances, indicating that they over-optimize the reward while losing coverage of the tilted target. In contrast, flow map look-ahead consistently achieves much lower MMD and SW2 across all step counts, while requiring far fewer resampling steps. This suggests that the flow map provides a much more reliable look-ahead for SMC, allowing the particle system to track the tilted distribution without collapsing. Furthermore, the lower number of resampling steps suggests more stable SMC behavior with smaller importance weights variance. 

\begin{table}[h]
\caption{\textbf{Quantitative results for the 2D toy experiment.}
We compare no-lookahead, denoiser-lookahead, and flow map-lookahead across different numbers of SMC steps. Lower Maximum Mean Discrepancy (MMD) and Sliced Wasserstein-2 (SW2) indicate better agreement with the tilted target distribution.}
\vspace{2mm}
\centering
\label{tab:toy_2d_smc_sweep}
\begin{tabular}{llcccc}
\toprule
\textbf{Steps} & \textbf{Method} & \textbf{MMD $\downarrow$} & \textbf{SW2 $\downarrow$} & \textbf{Mean Reward $\uparrow$} & \textbf{$\#$ of Resamples $\downarrow$} \\
\midrule
\multirow{3}{*}{50}
& No lookahead & 0.996 & 0.578 & 0.911 & 46 \\
& Denoiser & 3.973 & 0.988 & 0.917 & 47 \\
& \textbf{Flow map} & \textbf{0.043} & \textbf{0.060} & 0.621 & 5 \\
\midrule
\multirow{3}{*}{100}
& No lookahead & 0.460 & 0.443 & 0.958 & 64 \\
& Denoiser & 4.131 & 0.979 & 0.949 & 81 \\
& \textbf{Flow map} & \textbf{0.065} & \textbf{0.083} & 0.925 & 10 \\
\midrule
\multirow{3}{*}{200}
& No lookahead & 4.332 & 0.902 & 0.977 & 84 \\
& Denoiser & 3.016 & 0.942 & 0.964 & 135 \\
& \textbf{Flow map} & \textbf{0.065} & \textbf{0.126} & 0.967 & 9 \\
\midrule
\multirow{3}{*}{500}
& No lookahead & 3.459 & 0.857 & 0.972 & 155 \\
& Denoiser & 3.717 & 0.977 & 0.976 & 198 \\
& \textbf{Flow map} & \textbf{0.031} & \textbf{0.089} & 0.971 & 6 \\
\bottomrule
\end{tabular}
\end{table}

\newpage
\subsection{MNIST Experiments}
\label{app:additional_MNIST_experiments}

We ran MNIST experiments using a flow map model that we trained from scratch, with stochastic interpolants $\alpha_t = 1-t$, $\beta_t = t$. We use the unconditional flow map, which samples the distribution $\rho_1$ of all MNIST digits. We run \Cref{alg1:inference_time_adaptation} with $N=128$ particles and $K=200$ simulation steps.

We ablate the following look-ahead approaches introduced in \Cref{subsec:fixing_inference_time}:
\begin{itemize}[left=3pt]
    \item Naive look-ahead / no look-ahead (NL): $r_t(x) = t \cdot r(x)$.
    \item Denoiser look-ahead (WD): $r_t(x) = t \cdot r( \mathsf D _t(x))$.
    \item Flow map look-ahead (FMTT): $r_t(x) = t \cdot r( X_{t,1}(x))$.
\end{itemize}
We set $\epsilon_t = \alpha_t = 1-t$. We consider the four choices of $\chi_t$ described in \Cref{app:proof:map:tilt:reward}: 
\begin{itemize}[left=3pt]
\item Default: $\chi_t = 0$.
\item Tilted score dynamics: $\chi_t = \eta_t :=\alpha_t (\frac{\dot{\beta}_t}{\beta_t} \alpha_t - \dot{\alpha}_t) = (1-t) \big( \frac{1-t}{t} + 1 \big) = \frac{1-t}{t}$. In practice, we add an offset of $0.05$ to $\beta_t$ in the denominator: $\chi_t =\alpha_t (\frac{\dot{\beta}_t}{\beta_t + 0.05} \alpha_t - \dot{\alpha}_t) = (1-t) \big( \frac{1-t}{t + 0.05} + 1 \big) = \frac{1.05 \times (1-t)}{t + 0.05}$.
\item Local tilt dynamics: $\chi_t = \epsilon_t = 1-t$.
\item Base dynamics: $\chi_t = - \epsilon_t = -1 + t$.
\end{itemize}

\begin{figure}[h]
    \centering
    \begin{minipage}{0.025\linewidth}
        \centering
        \rotatebox{90}{$\chi_t = 0$}
    \end{minipage}%
    \begin{minipage}{0.97\linewidth}
        \includegraphics[width=\linewidth]{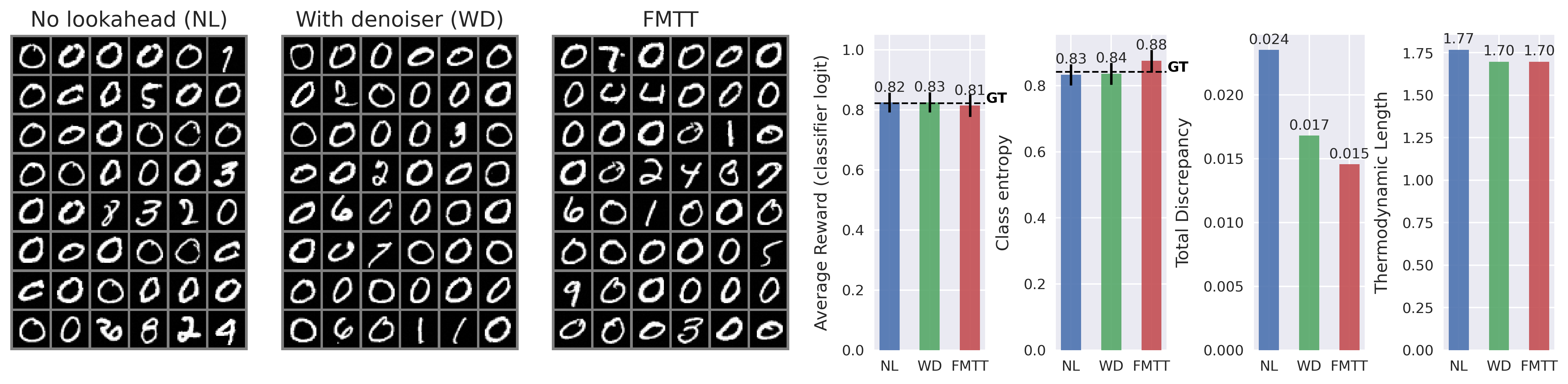}
    \end{minipage}

    \begin{minipage}{0.025\linewidth}
        \centering
        \rotatebox{90}{$\chi_t = \eta_t$}
    \end{minipage}%
    \begin{minipage}{0.97\linewidth}
        \includegraphics[width=\linewidth]{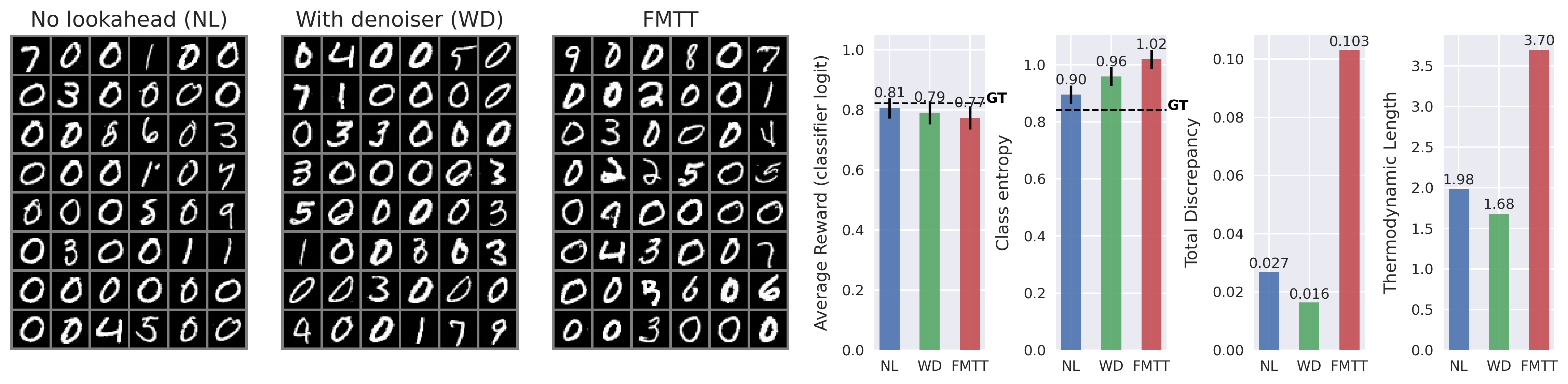}
    \end{minipage}

    \begin{minipage}{0.025\linewidth}
        \centering
        \rotatebox{90}{$\chi_t = \epsilon_t$}
    \end{minipage}%
    \begin{minipage}{0.97\linewidth}
        \includegraphics[width=\linewidth]{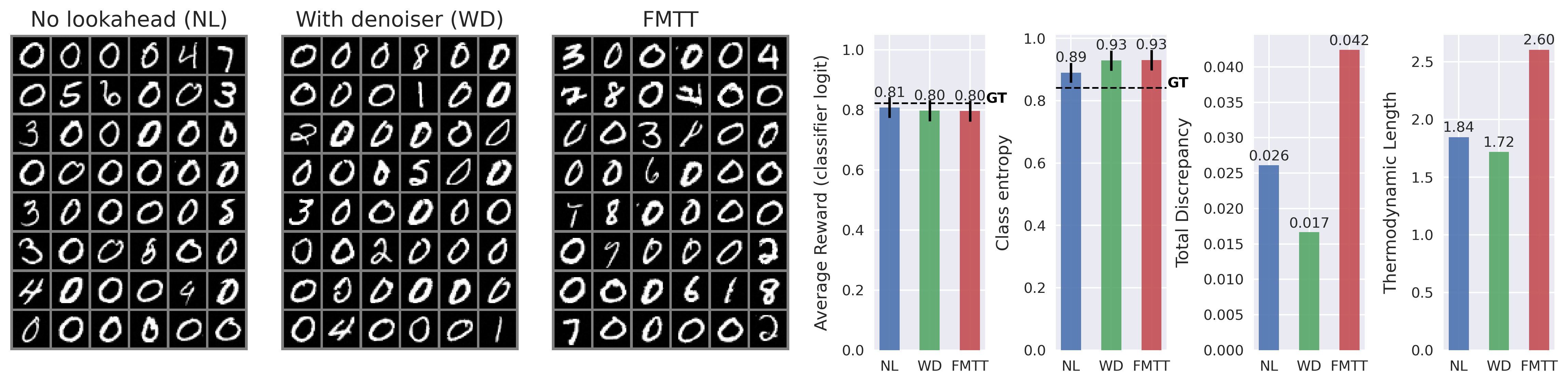}
    \end{minipage}

    \begin{minipage}{0.025\linewidth}
        \centering
        \rotatebox{90}{$\chi_t = -\epsilon_t$}
    \end{minipage}%
    \begin{minipage}{0.97\linewidth}
        \includegraphics[width=\linewidth]{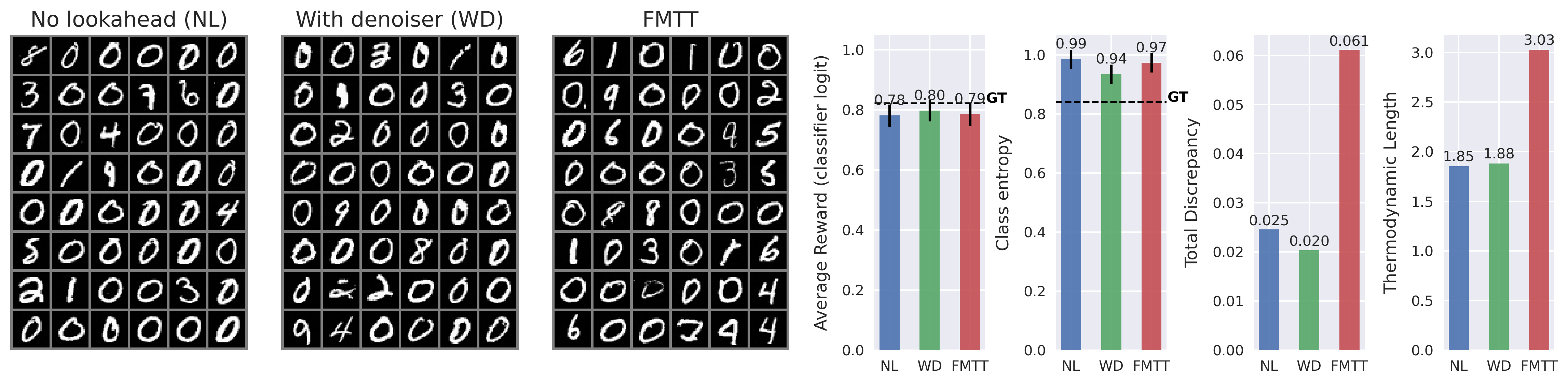}
    \end{minipage}

    \caption{Comparison of MNIST tilted sampling to generate digits that would be classified as zeros: the reward is $r(x) = 0.1 \times \log p(0|x)$. The four rows correspond to the dynamics with different $\chi_t$ choices described in \Cref{app:proof:map:tilt:reward}.}
    \label{fig:mnist}
\end{figure}

\begin{figure}[h]
    \centering

    \begin{minipage}{0.025\linewidth}
        \centering
        \rotatebox{90}{$\chi_t = 0$}
    \end{minipage}%
    \begin{minipage}{0.97\linewidth}
        \includegraphics[width=\linewidth]{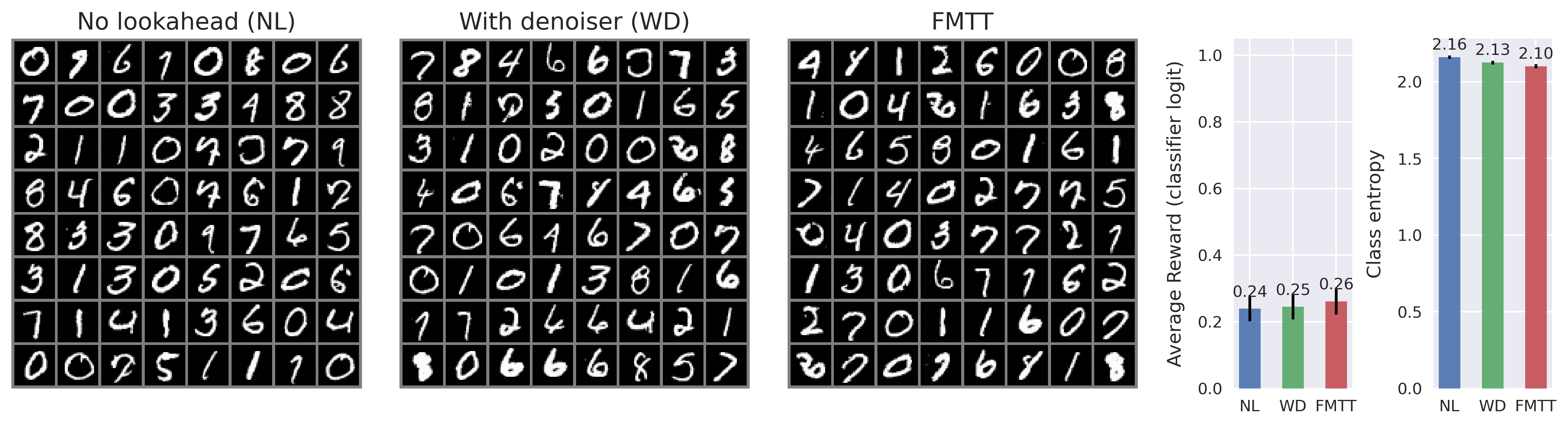}
    \end{minipage}

    \begin{minipage}{0.025\linewidth}
        \centering
        \rotatebox{90}{$\chi_t = \eta_t$}
    \end{minipage}%
    \begin{minipage}{0.97\linewidth}
        \includegraphics[width=\linewidth]{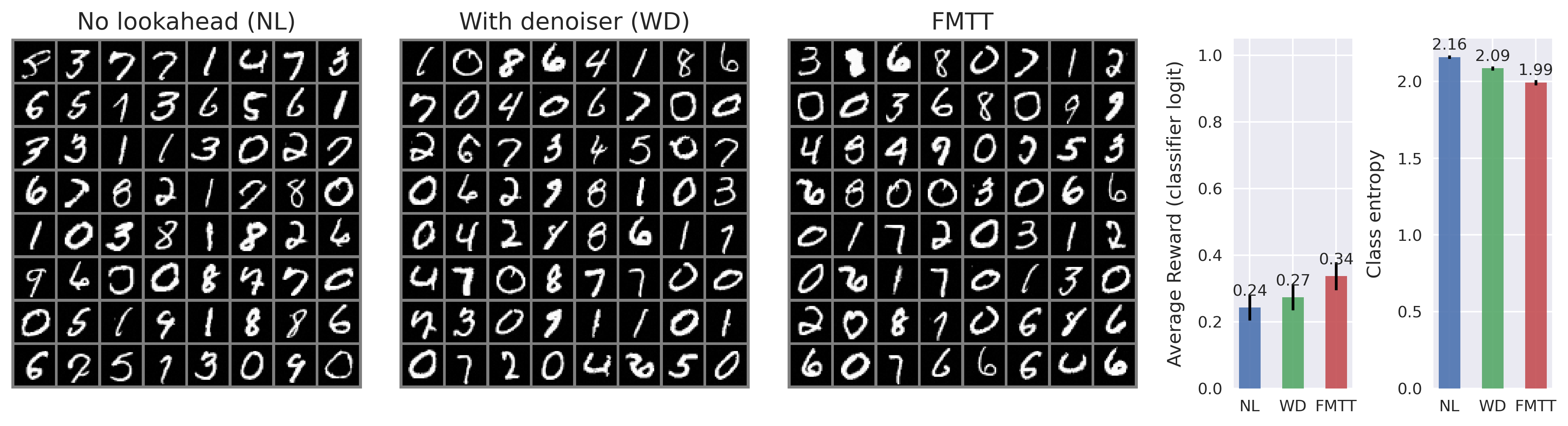}
    \end{minipage}

    \begin{minipage}{0.025\linewidth}
        \centering
        \rotatebox{90}{$\chi_t = \epsilon_t$}
    \end{minipage}%
    \begin{minipage}{0.97\linewidth}
        \includegraphics[width=\linewidth]{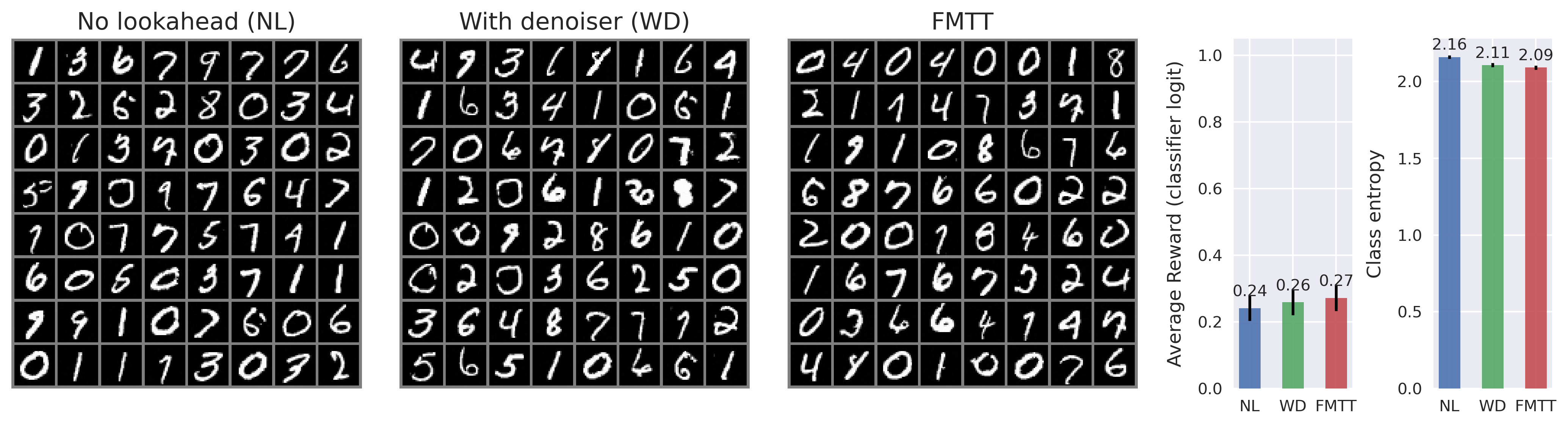}
    \end{minipage}

    \begin{minipage}{0.025\linewidth}
        \centering
        \rotatebox{90}{$\chi_t = -\epsilon_t$}
    \end{minipage}%
    \begin{minipage}{0.97\linewidth}
        \includegraphics[width=\linewidth]{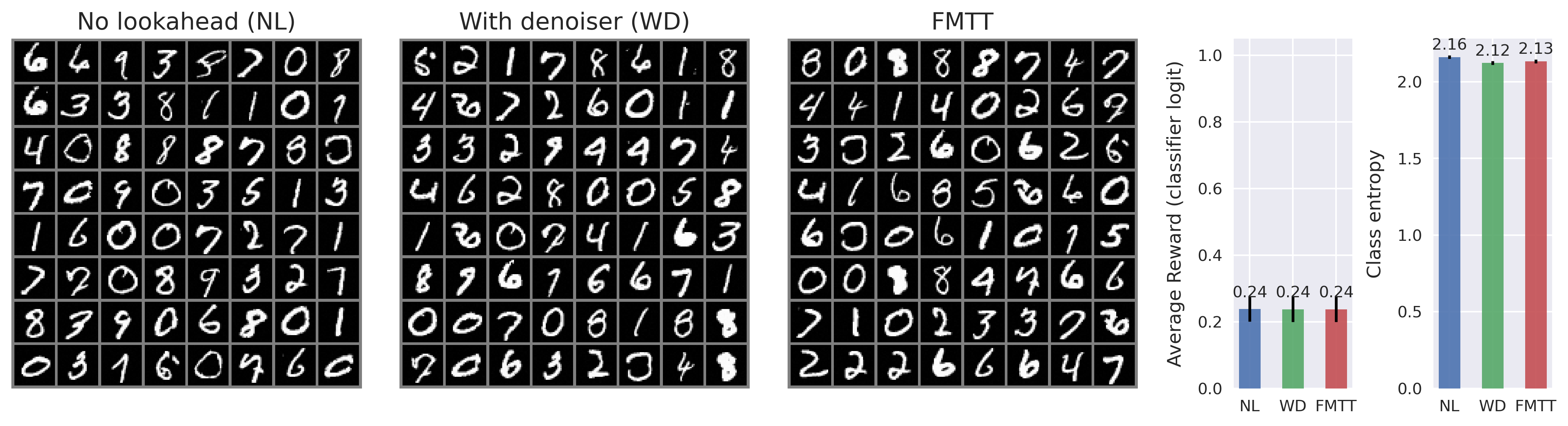}
    \end{minipage}

    \caption{Comparison of MNIST greedy search to generate digits that would be classified as zeros: the reward is $r(x) = 0.05 \times \log p(0|x)$. The four rows correspond to the dynamics with different $\chi_t$ choices described in \Cref{app:proof:map:tilt:reward}.}
    \label{fig:mnist_greedy}
\end{figure}

We run \textcolor{red}{Sampling} and \textcolor{blue}{Searching} experiments.
\begin{itemize}[left=3pt]
    \item For \textcolor{red}{Sampling} experiments (\Cref{fig:mnist}), we set $r(x) = 0.1 \times \log p_{\theta}(0|x)$, where $p_{\theta}$ is a pre-trained MNIST classifier and $\log p_{\theta}(0|x)$ is the probability that the classifier assigns to the digit 0 for the image $x$. Following \Cref{sec:methodology}, the target distribution that we aim to sample is the tilted probability distribution is $\hat{\rho}_1(x) \propto \rho_1(x) \exp \big(0.1 \times \log p_{\theta}(0|x) \big)$. We run the sampling experiments with $C = 1$ clone. We perform the log-weight update in Line 10 using approach \emph{(iii)} (see \Cref{subsubsec:best_performing} for a discussion on the different approaches). We remark again that the choice $\chi_t = 0$ is much faster computationally, as it does not require estimating expectations. For the other three choices of $\chi_t$, we use $M = 400$ noisy samples when computing \eqref{eq:approach_3_positive} or \eqref{eq:approach_3_negative}. Instead of resampling every $R$ steps with $R$ fixed, we resample when the effective sample size, which is computed as $\sum_{i=1}^{N}\exp(A^{k}_{i})^2/\big(\sum_{i=1}^{N}\exp(A^{k}_{i} \big)^2$, falls below $0.85 \times N$.
    \item For \textcolor{blue}{Searching} experiments (\Cref{fig:mnist_greedy}), $r(x) = 0.05 \times \log p_{\theta}(0|x)$. Note that in this case, there does not exist an explicit form for the target probability distribution, which is different for each algorithm. We run the simulation with $C=2$ clones, and do one resampling step after the 100th simulation step (in the middle of the generation). 
\end{itemize}

\paragraph{\textcolor{red}{Sampling} results} In \Cref{fig:mnist}, we show samples generated using each algorithm and plot four metrics. For each algorithm, the metrics were computed from 16 runs, each with $N=128$ particles, and the bars show the average value plus/minus the standard error.
\begin{itemize}[left=3pt]
    \item Average reward: this is the expected value of $\log p_{\theta}(0|x)$ for the samples generated with the algorithm (without the multiplier $0.1$). We also print the ground truth value, which we compute by sampling $400\times128=51200$ independent samples from the unconditional model, and reweighting them by the factor $\exp \big(0.1 \times \log p_{\theta}(0|x) \big)$ upon normalization.
    \item Class entropy: this is the average value of the entropy of the probability distribution $(p_{\theta}(i|x))_{i=0}^{9}$ induced by the classifier $p_{\theta}$ for each generated sample $x$, i.e. $- \sum_{i=0}^{9} \log p_{\theta}(i|x)$. We also print the ground truth value, which we compute by sampling $51200$ samples and reweighting them accordingly.
    \item Total discrepancy: this is the quantity $D(\mathcal{T}) = \sum_{k=1}^{K} D(t_{k-1}, t_k)$ defined in equation \eqref{eq:cumulative_total_discrepancy}, where $D(t_{k-1}, t_k)$ are computed using the procedure described in \Cref{subsec:discrepancy}.
    \item Thermodynamic length: this is the $\Lambda = \sum_{k=1}^{K} \sqrt{D(t_{k-1}, t_k)}$ defined in \eqref{eq:thermodynamic_length_def} (strictly speaking, this is an approximation of the actual thermodynamic length).
\end{itemize}
We observe that the only $\chi_t$ choice for which the ground truth values for the average reward and the class entropy fall within the error bars is $\chi_t = 0$. For this choice, the look-ahead approach that results in lower total discrepancy and thermodynamic length is FMTT. To understand why this is the case, recall that 
the evolution of the log-weights for $\chi_t = 0$ is given by the ODE $\frac{\mathrm{d}A_t}{\mathrm{d}t} \! = \! \frac{\mathrm{d}}{\mathrm{d}s} r_s(X^{0}_s) \big|_{\substack{X^{0}_t = \tilde{x}_t, s = t}}$. Observe that when $r_t(x) = r^{\star}_t(x) := r(X_{t,1}(x))$, then $\frac{\mathrm{d}A_t}{\mathrm{d}t} = 0$, which implies that the total discrepancy and the thermodynamic length are zero, and it is easy to see that this is the only choice of $r_t$ for which they are zero\footnote{Observe that $r_t(x) = r^{\star}_t(x) = r(X_{t,1}(x))$ cannot be used in practice because we require that $r_0(x)$ is constant so that we can sample from $\hat{\rho}_0$}. Thus, the total discrepancy and the thermodynamic length values can be regarded as a measure of how close a given $r_t$ is to $r^{\star}_t := r \circ X_{t,1}$. From their values in the first row of \Cref{fig:mnist}, we deduce that the flow map look-ahead reward $r_t(x) = t \cdot r(X_{t,1}(x))$ is the closest one to $r^{\star}_t(x) = r(X_{t,1}(x))$, which is unsurprising because they only differ by a factor $t$.

For all other $\chi_t$ choices, the ground truth class entropy values fall outside of the error bars of the model. Moreover, for these other $\chi_t$ choices, the total discrepancy and thermodynamic length are substantially higher for FMTT than for NL and WD, and are lowest for WD. To understand why this is the case, observe that the evolution of the log-weights for $\chi_t = \eta_t$ and $\chi_t = \epsilon_t$ is given by the ODE $\frac{\mathrm{d}A_t}{\mathrm{d}t} = \frac{\mathrm{d}}{\mathrm{d}s} \mathbb{E}[ e^{r_s(X^{\chi}_s)-r_t(\tilde{x}_t)} \,|\, X^{\chi}_t = \tilde{x}_t ] \big|_{s = t}$. Note that when $r_t$ is the optimal time-dependent reward in the stochastic optimal control sense, i.e., $r_t(x) = r^{\star}_t(x) := \log \mathbb{E}[\exp(r_1(X^{\chi}_1)) | X^{\chi}_t = \tilde{x}_t]$\footnote{This holds by the path integral characterization of stochastic optimal control.}, we have that $\mathbb{E}[ e^{r_s(X^{\chi}_s)-r_t(\tilde{x}_t)} \,|\, X^{\chi}_t = \tilde{x}_t ] = 0$ for all $0 \leq t \leq s \leq 1$. Hence, for this choice of $r_t$, the total discrepancy and the thermodynamic length are zero, and it is the only choice for which they are zero. Like before, the total discrepancy and the thermodynamic length values can be regarded as a measure of how close a given $r_t$ is to $r^{\star}_t$. From their values in the second and third row of \Cref{fig:mnist}, we deduce that the denoiser look-ahead reward $r_t(x) = t \cdot r(\mathsf D_t(x))$ is the closest one to $r^{\star}_t(x)$, and $r_t(x) = t \cdot r(X_{t,1}(x))$ is the farthest apart.

\paragraph{\textcolor{blue}{Searching} results} In \Cref{fig:mnist_greedy}, we show samples generated using each algorithm and plot the average reward and average class entropy methods, which are also computed from 16 runs with $N=128$ particles each, as in the sampling case. In this case, there are no ground truth values to compare to. We observe that all else equal, FMTT with $\chi_t = \eta_t$ is the most effective choice in that it achieves the highest average reward and the lowest class entropy, hence sampling the digit 0 more often. Accordingly, FMTT with $\chi_t = \eta_t$ is the choice we make for our high dimensional search experiments throughout the paper.

\FloatBarrier
\newpage

\subsection{Text-to-Image Experiments}
\subsubsection{Additional Geometric Reward Examples}
\label{app:additional_geometric_examples}
\begin{figure}[!htbp]
  \centering
  \includegraphics[width=\linewidth]{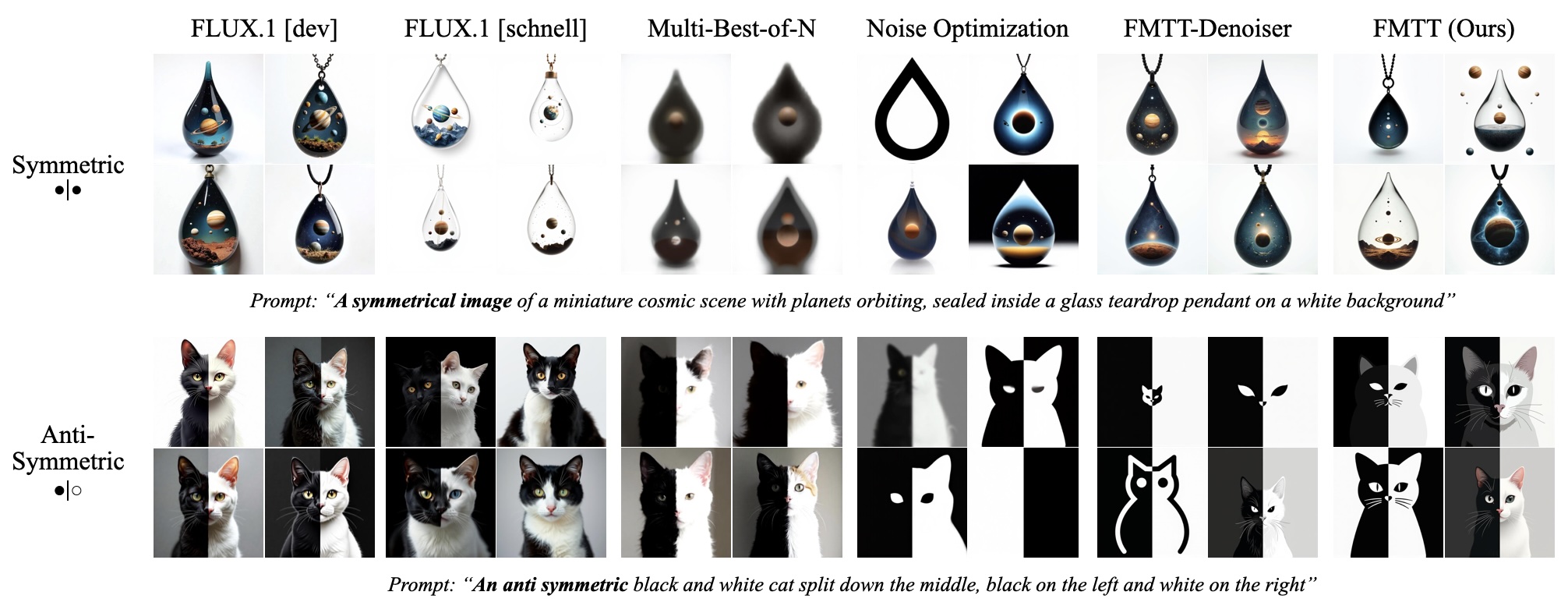}
  \caption{Qualitative comparison on two basic geometric rewards (symmetry, anti-symmetry). The gradient-based methods that change the generative dynamics produce sharper images that satisfy the constraints more reliably than prior methods. Note how Multi-best-of-N and ReNO-style noise optimization result in blurry images while FMTT methods produce clear results that properly satisfy the constraints. See \Cref{app:additional_geometric_examples} for additional examples.\looseness=-1}
  \label{fig:symm_rot_reward}
\end{figure}

\begin{figure}[!htbp]
    \centering
    \includegraphics[width=0.87\linewidth]{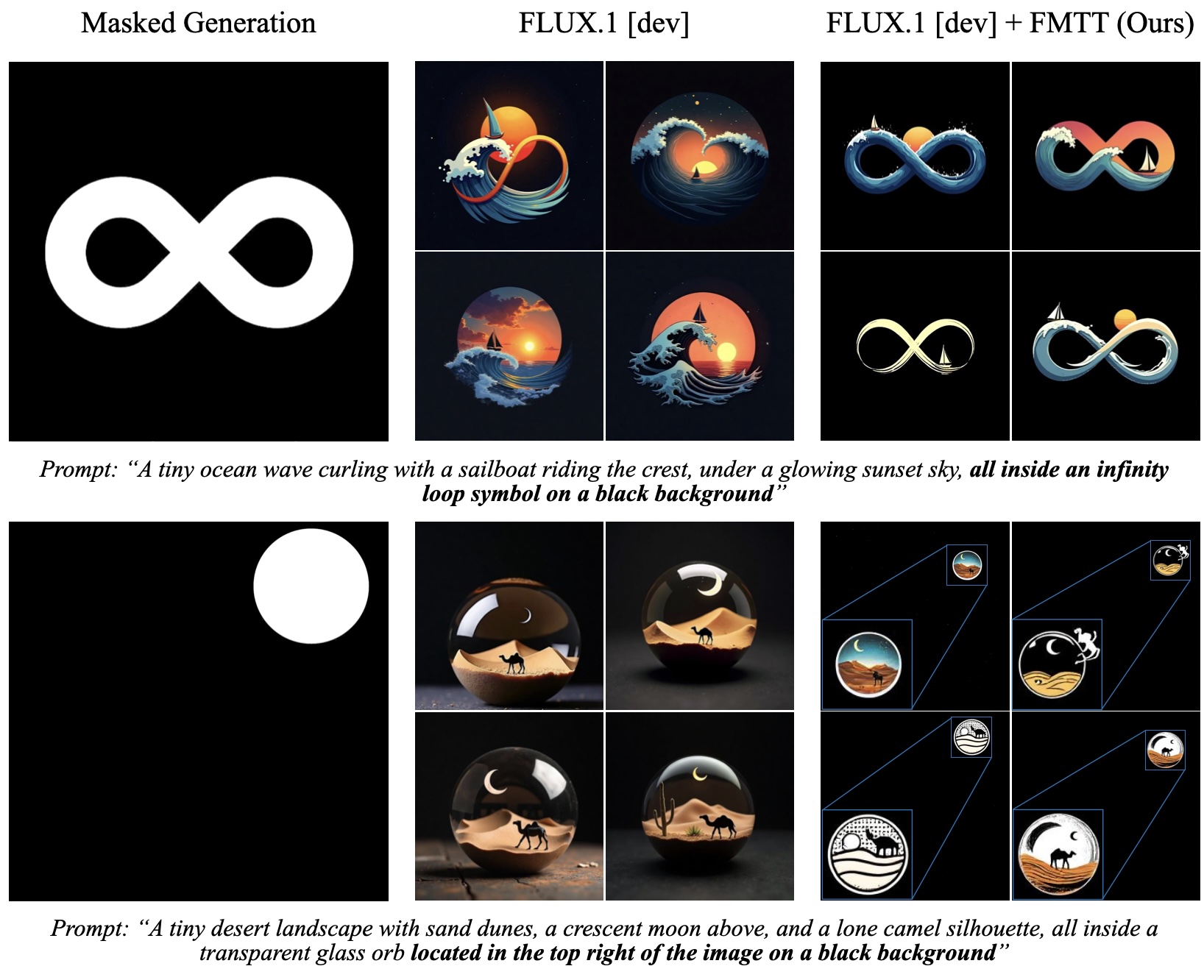}
    \caption{Additional qualitative comparisons using masked rewards. Note the zoom-ins in the second example for FMTT.}
    \label{fig:additional_geometric}
\end{figure}

\begin{figure}[!htbp]
    \centering
    \includegraphics[width=0.87\linewidth]{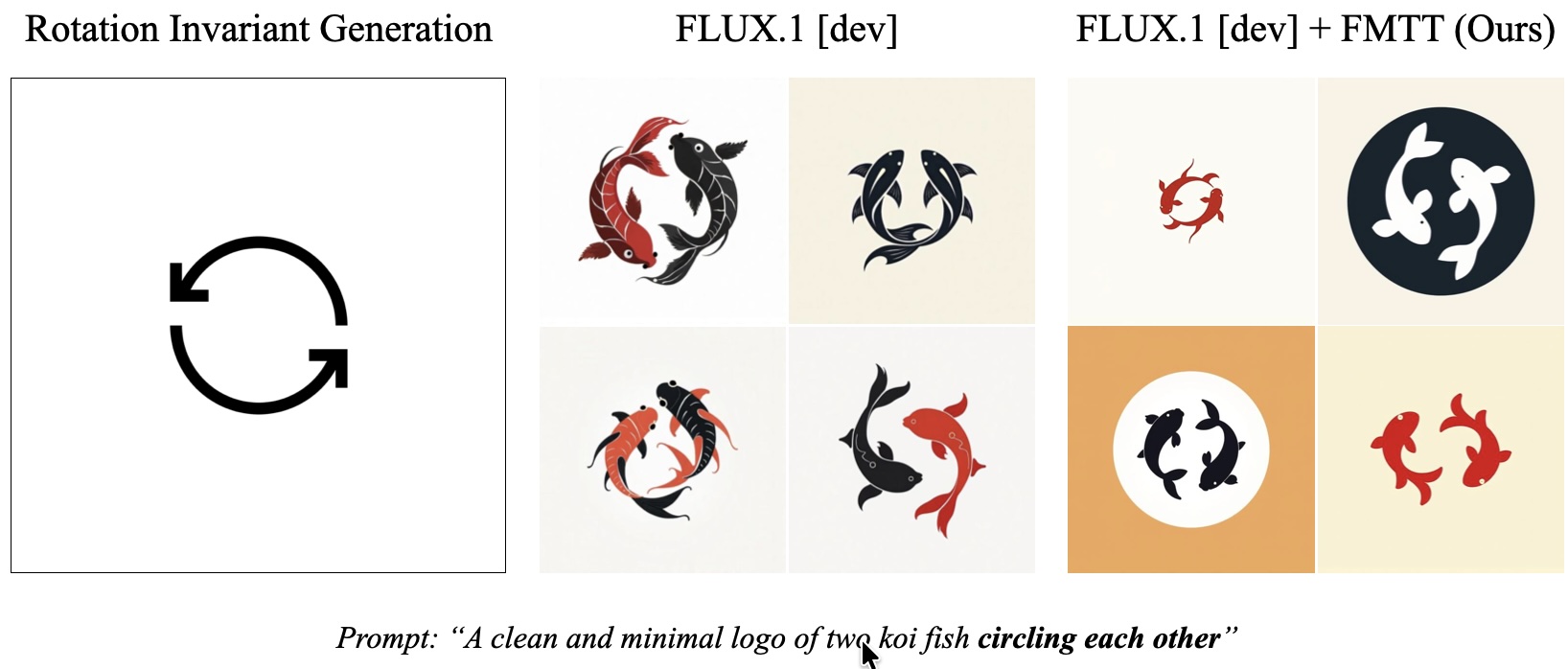}
    \caption{Additional qualitative comparison using rotation invariant rewards.}
    \label{fig:additional_geometric_2}
\end{figure}

Additional qualitative comparisons when using a geometric reward can be seen in \Cref{fig:symm_rot_reward,fig:additional_geometric,fig:additional_geometric_2}.

\FloatBarrier
\newpage

\vspace{-5mm}
\subsubsection{Additional VLM-as-judge Examples}
\label{app:additional_vlm_examples}
\begin{figure}[!h]
    \centering
    \includegraphics[width=0.99\linewidth]{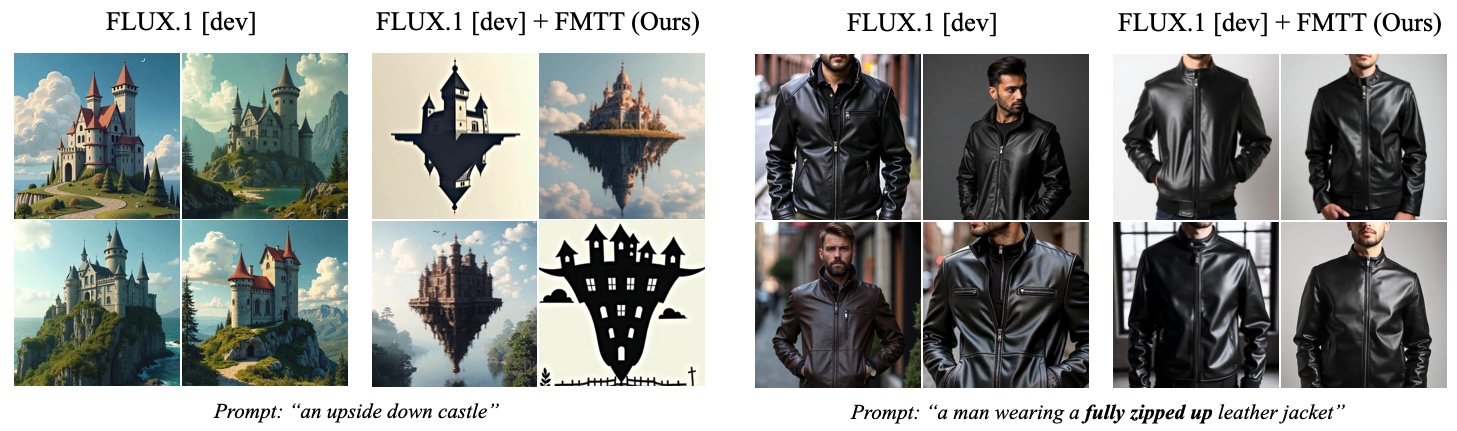}
    \caption{Additional qualitative comparison using VLM-as-judge rewards.}
    \label{fig:additional_vlms_1}
\end{figure}

Additional qualitative results with VLM-as-judge rewards can be seen in \Cref{fig:additional_vlms_1}. 

\begin{figure}[!h]
    \centering
    \includegraphics[width=0.7\linewidth]{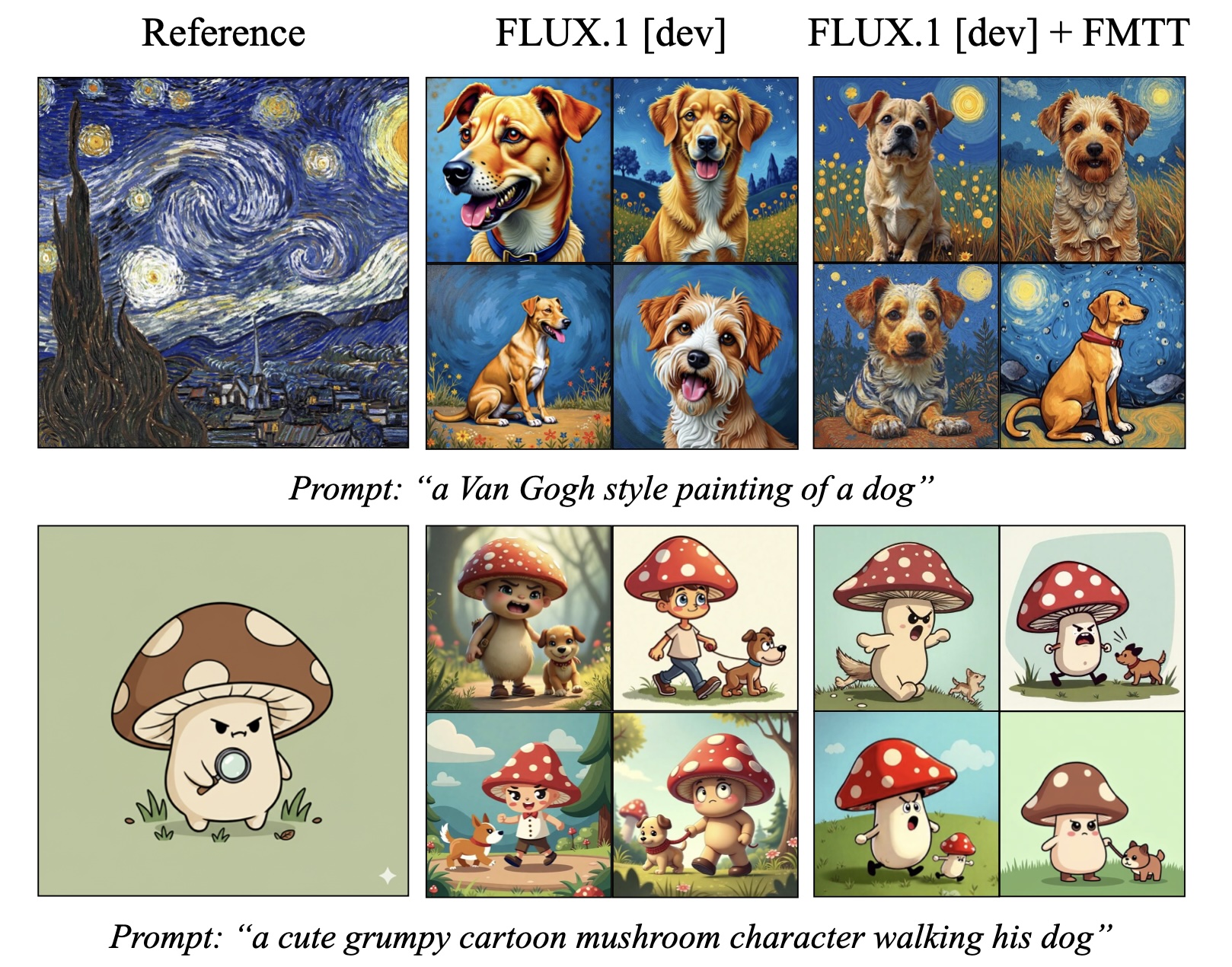}
    \caption{Style consistency via VLM-based rewards. Given a reference image, FMTT produces images that better match its art style than the base model.}
    \label{fig:qwenvl_style_consistency}
\end{figure}

As described in the paper, we also use Qwen2.5-VL-7B-Instruct to define rewards expressed as yes/no questions over one or more context images. This makes it possible to cast diverse objectives as test-time search problems, including style consistency and character consistency. Here, we demonstrate the style consistency case. The VLM receives both a reference image and a generated image and is asked whether they share the same art style. FMTT then optimizes this reward, producing generations more closely aligned with the reference style. The qualitative results are shown in \Cref{fig:qwenvl_style_consistency}.  

\subsubsection{VLM Reward Hacking} 
\label{app:vlm_reward_hacking}
\begin{figure}[H]
    \centering
    \includegraphics[width=0.5\linewidth]{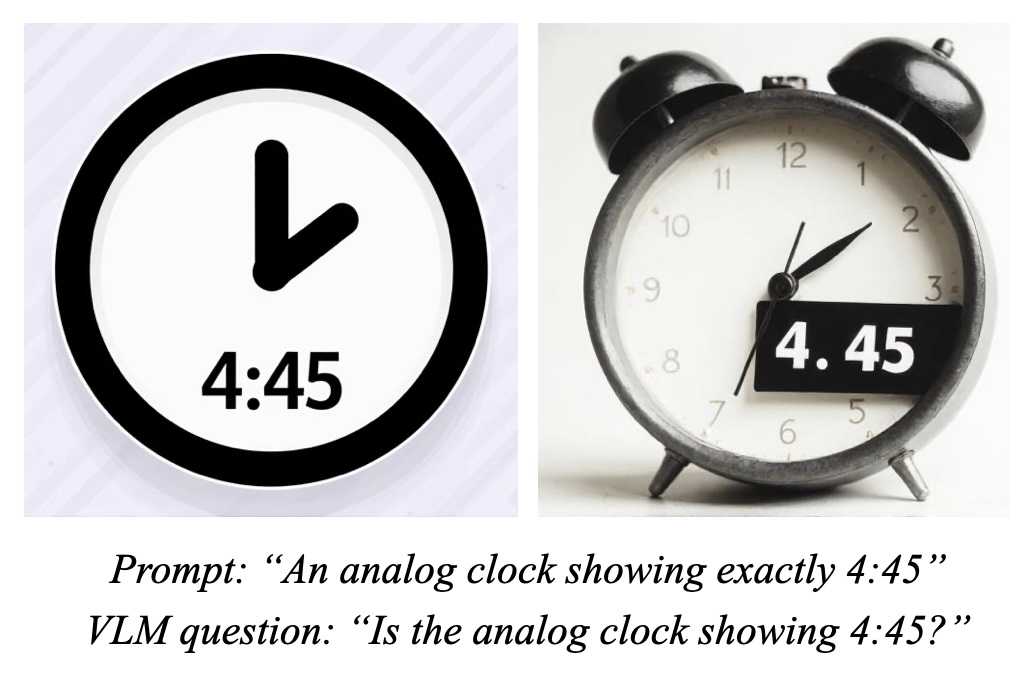}
    \caption{VLM reward hacking. Instead of the clock being at 4:45, the search process finds a way to ``cheat'' by writing the text \texttt{4:45} on the clock face and achieving high rewards from the VLM.}
    \label{fig:vlm_reward_hack}
\end{figure}

As discussed in the paper, a challenge of using VLMs (or any non-verifiable reward model) is the risk of the search process exploiting loopholes. This happens when the algorithm produces images that either act as adversarial examples for the VLM or satisfy the literal question without achieving the intended effect. \Cref{fig:vlm_reward_hack} shows such a case.  

We explored two solutions. The first is to craft the VLM prompt to be as verbose and unambiguous as possible, explicitly discouraging potential ``cheats''. This works when only a few edge cases exist, but becomes brittle when many (4+) conditions are needed, at which point the reward model grows opaque and the search converges to local maxima. 
The second approach is to decompose the binary question into several simpler sub-questions and define the reward as their sum. While this adds computational overhead by requiring multiple VLM inferences, it proved more robust in practice.  

For reference, to achieve the results in \Cref{fig:teaser}, we used the following three questions:  
\begin{itemize}
    \item Is the hour hand pointing between 4 and 5?  
    \item Is the minute hand pointing at 9?  
    \item Is the second hand pointing at 12?  
\end{itemize}

\subsubsection{Complete UniGenBench++ Results} \label{app:unigenbench_details}
\Cref{tab:unigenbench} reports the complete per-dimension scores used by UniGenBench++, including entity layout, text rendering, world knowledge, spatial reasoning, and all other evaluation categories.

\begin{table}[h!]
\caption{Quantitative results on UniGenBench++. Entries show the percentage of images judged by the evaluation model to satisfy each criterion. The standard deviation is computed across the 4 images generated for every prompt. $N$ represents the number of distinct samples/particles used during generation.}
\centering
\label{tab:unigenbench}
\resizebox{0.99\linewidth}{!}{
\begin{tabular}{lcccccccccccc}
\toprule
\textbf{Method} & \textbf{Mean} & \textbf{Style} & \textbf{World Knowledge} & \textbf{Attribute} & \textbf{Action} & \textbf{Relationship} & \textbf{Compound} & \textbf{Grammar} & \textbf{Logical Reasoning} & \textbf{Entity Layout} & \textbf{Text Generation} & \textbf{NFE}   \\
\midrule
\multicolumn{9}{l}
{\textbf{Diffusions}} \\
\midrule
FLUX.1 [dev] & 61.78 $\pm$ {\footnotesize 3.58} & 85.20 $\pm$ {\footnotesize 2.30} & 86.23 $\pm$ {\footnotesize 2.62} & 64.53 $\pm$ {\footnotesize 2.82} & 60.46 $\pm$ {\footnotesize 3.80} & 63.07 $\pm$ {\footnotesize 5.46} & 41.24  $\pm$ {\footnotesize 5.33} & 60.96  $\pm$ {\footnotesize 2.95} & 24.08  $\pm$ {\footnotesize 3.63} & 67.16  $\pm$ {\footnotesize 4.32}  & 30.17  $\pm$ {\footnotesize 5.77} & 180 \\
Flow Map & 58.16 $\pm$ {\footnotesize 0.66} & 79.70 $\pm$ {\footnotesize 2.42} & 80.22 $\pm$ {\footnotesize 1.04} & 59.51 $\pm$ {\footnotesize 1.26} & 57.22 $\pm$ {\footnotesize 1.44} & 62.44 $\pm$ {\footnotesize 0.95} & 42.01  $\pm$ {\footnotesize 2.20} & 57.62  $\pm$ {\footnotesize 1.33} & 25.92  $\pm$ {\footnotesize 2.37} & 70.90  $\pm$ {\footnotesize 1.75}  & 3.74  $\pm$ {\footnotesize 0.50} & 16 \\
\midrule{\textbf{Gradient-Free Search}} \\
\midrule
FLUX.1 [dev] + Best-of-$N$ \\
- $N = 8$& 65.40 $\pm$ {\footnotesize 2.56} & 87.40 $\pm$ {\footnotesize 2.09} & 89.72  $\pm$ {\footnotesize 1.75} & 68.27  $\pm$ {\footnotesize 2.50} & 62.26  $\pm$ {\footnotesize 3.64} & 68.02  $\pm$ {\footnotesize 4.38} & 47.29  $\pm$ {\footnotesize 1.80} & 63.37  $\pm$ {\footnotesize 2.16} & 29.36  $\pm$ {\footnotesize 3.24} & 71.08  $\pm$ {\footnotesize 71.08} & 35.06  $\pm$ {\footnotesize 5.66} & 360 \\
- $N = 16$& 67.57  $\pm$ {\footnotesize 1.53} & 88.60  $\pm$ {\footnotesize 1.40} & 90.82  $\pm$ {\footnotesize 1.30}  & 69.44 $\pm$ {\footnotesize 1.37} & 64.92  $\pm$ {\footnotesize 2.33} & 70.81 $\pm$ {\footnotesize 3.10} & 52.71 $\pm$ {\footnotesize 2.48} & 64.97 $\pm$ {\footnotesize 1.10} & 31.19 $\pm$ {\footnotesize 2.34} & 72.95 $\pm$ {\footnotesize 2.26} & 36.49 $\pm$ {\footnotesize 4.01} & 720 \\
- $N = 32$& 69.18 $\pm$ {\footnotesize 0.87} & 89.80 $\pm$ {\footnotesize 1.40} & 90.82  $\pm$ {\footnotesize 1.30} & 72.33 $\pm$ {\footnotesize 1.22} & 66.06 $\pm$ {\footnotesize 1.30} & 72.97 $\pm$ {\footnotesize 1.98} & 54.64 $\pm$ {\footnotesize 1.59} & 65.24 $\pm$ {\footnotesize 1.65} & 31.88 $\pm$ {\footnotesize 1.00} & 76.49 $\pm$ {\footnotesize 1.54} & 39.37 $\pm$ {\footnotesize 4.55} & 1440 \\
- $N = 64$& 70.93 $\pm$ {\footnotesize 1.38} & 90.50 $\pm$ {\footnotesize 0.59} & 90.98 $\pm$ {\footnotesize 0.82} & 74.15 $\pm$ {\footnotesize 1.37} & 67.97 $\pm$ {\footnotesize 0.73} & 75.00  $\pm$ {\footnotesize 0.66} & 56.96 $\pm$ {\footnotesize 4.13} & 65.24 $\pm$ {\footnotesize 3.00} & 38.30 $\pm$ {\footnotesize 4.12} & 77.99 $\pm$ {\footnotesize 1.71} & 42.82 $\pm$ {\footnotesize 3.29} & 2880 \\
- $N = 128$& 72.77 $\pm$ {\footnotesize 1.26} & 91.40 $\pm$ {\footnotesize 0.66} & 91.77 $\pm$ {\footnotesize 0.45} & 75.00 $\pm$ {\footnotesize 1.56} & 70.15 $\pm$ {\footnotesize 1.38} & 74.24 $\pm$ {\footnotesize 2.36} & 60.95 $\pm$ {\footnotesize 2.20} & 68.85 $\pm$ {\footnotesize 2.49} & 41.97 $\pm$ {\footnotesize 2.86} & 80.60 $\pm$ {\footnotesize 1.40} & 44.54 $\pm$ {\footnotesize 5.71} & 5760 \\
- $N = 256$& 73.75 $\pm$ {\footnotesize 1.06} & 91.10 $\pm$ {\footnotesize 0.59} & 92.88 $\pm$ {\footnotesize 0.52} & 76.39 $\pm$ {\footnotesize 0.82} & 71.58 $\pm$ {\footnotesize 1.83} & 75.51 $\pm$ {\footnotesize 1.54} & 61.86 $\pm$ {\footnotesize 3.44} & 69.52 $\pm$ {\footnotesize 0.53} & 45.18 $\pm$ {\footnotesize 3.14} & 80.97 $\pm$ {\footnotesize 1.54} & 44.83 $\pm$ {\footnotesize 3.54} & 11520 \\

\midrule
FLUX.1 [schnell] + Best-of-$N$ \\
- $N = 4$   & 67.37 $\pm$ {\footnotesize 3.74} & 89.40 $\pm$ {\footnotesize 2.27} & 89.40 $\pm$ {\footnotesize 2.91} & 67.31 $\pm$ {\footnotesize 5.10} & 65.97 $\pm$ {\footnotesize 3.12} & 70.69 $\pm$ {\footnotesize 3.97} & 52.06 $\pm$ {\footnotesize 6.57} & 63.37 $\pm$ {\footnotesize 3.24} & 35.09 $\pm$ {\footnotesize 4.27} & 78.92 $\pm$ {\footnotesize 3.35} & 26.44 $\pm$ {\footnotesize 7.67} & 16 \\
- $N = 8$   & 71.25 $\pm$ {\footnotesize 1.87} & 91.30 $\pm$ {\footnotesize 2.37} & 92.56 $\pm$ {\footnotesize 0.69} & 72.44 $\pm$ {\footnotesize 2.43} & 69.77 $\pm$ {\footnotesize 3.83} & 75.00 $\pm$ {\footnotesize 0.66} & 57.86 $\pm$ {\footnotesize 3.04} & 66.58 $\pm$ {\footnotesize 2.22} & 39.68 $\pm$ {\footnotesize 1.64} & 82.09 $\pm$ {\footnotesize 2.74} & 30.46 $\pm$ {\footnotesize 1.72} & 32 \\
- $N = 16$  & 73.59 $\pm$ {\footnotesize 1.83} & 91.90 $\pm$ {\footnotesize 1.34} & 93.83 $\pm$ {\footnotesize 0.94} & 74.79 $\pm$ {\footnotesize 3.11} & 73.95 $\pm$ {\footnotesize 2.83} & 76.52 $\pm$ {\footnotesize 2.77} & 61.73 $\pm$ {\footnotesize 2.17} & 67.91 $\pm$ {\footnotesize 1.96} & 43.35 $\pm$ {\footnotesize 2.70} & 85.63 $\pm$ {\footnotesize 1.33} & 31.32 $\pm$ {\footnotesize 2.49} & 64 \\
- $N = 32$  & 75.18 $\pm$ {\footnotesize 1.13} & 92.70 $\pm$ {\footnotesize 1.58} & 94.62 $\pm$ {\footnotesize 1.05} & 77.35 $\pm$ {\footnotesize 1.09} & 74.52 $\pm$ {\footnotesize 1.61} & 79.95 $\pm$ {\footnotesize 2.05} & 63.92 $\pm$ {\footnotesize 1.96} & 69.65 $\pm$ {\footnotesize 1.28} & 46.10 $\pm$ {\footnotesize 0.76} & 85.26 $\pm$ {\footnotesize 1.62} & 32.76 $\pm$ {\footnotesize 4.49} & 128 \\
- $N = 64$  & 75.88 $\pm$ {\footnotesize 0.92} & 93.50 $\pm$ {\footnotesize 0.91} & 95.41 $\pm$ {\footnotesize 0.52} & 77.46 $\pm$ {\footnotesize 2.08} & \textbf{75.10 $\pm$ {\footnotesize 0.95}} & 81.22 $\pm$ {\footnotesize 2.46} & 66.11 $\pm$ {\footnotesize 1.56} & 69.39 $\pm$ {\footnotesize 2.12} & 45.64 $\pm$ {\footnotesize 4.07} & 85.45 $\pm$ {\footnotesize 1.94} & 34.77 $\pm$ {\footnotesize 2.62} & 256 \\
- $N = 128$ & 76.89 $\pm$ {\footnotesize 0.51} & \textbf{94.50 $\pm$ {\footnotesize 0.59}} & 94.62 $\pm$ {\footnotesize 1.14} & 78.42 $\pm$ {\footnotesize 1.26} & \textbf{75.10 $\pm$ {\footnotesize 1.97}} & 82.36 $\pm$ {\footnotesize 1.50} & 67.78 $\pm$ {\footnotesize 1.73} & 71.39 $\pm$ {\footnotesize 1.23} & 50.00 $\pm$ {\footnotesize 2.64} & 86.94 $\pm$ {\footnotesize 2.01} & 33.33 $\pm$ {\footnotesize 3.35} & 512 \\
- $N = 256$ & \textbf{77.41 $\pm$ {\footnotesize 1.26}} & 93.40 $\pm$ {\footnotesize 1.73} & \textbf{94.94 $\pm$ {\footnotesize 1.73}} & \textbf{79.17 $\pm$ {\footnotesize 1.67}} & 75.00 $\pm$ {\footnotesize 1.24} & \textbf{83.63 $\pm$ {\footnotesize 1.31}} & \textbf{68.56 $\pm$ {\footnotesize 0.36}} & 71.79 $\pm$ {\footnotesize 2.52} & 51.15 $\pm$ {\footnotesize 4.22} & \textbf{88.62 $\pm$ {\footnotesize 2.26}} & 35.63 $\pm$ {\footnotesize 3.54} & 1024 \\

\midrule
Flow Map + Multi-best-of-$N$ (10 rounds) \\ 
- $N=8$ & 67.58 $\pm$ {\footnotesize 0.56} & 86.70 $\pm$ {\footnotesize 0.77} & 90.66 $\pm$ {\footnotesize 1.22} & 70.41 $\pm$ {\footnotesize 1.06} & 67.40 $\pm$ {\footnotesize 1.87} & 71.32 $\pm$ {\footnotesize 2.05} & 52.32 $\pm$ {\footnotesize 1.39} & 64.71 $\pm$ {\footnotesize 2.11} & 38.30 $\pm$ {\footnotesize 3.46} & 76.49 $\pm$ {\footnotesize 1.63} & 18.39 $\pm$ {\footnotesize 0.81} & 320 \\
- $N=16$ & 68.01 $\pm$ {\footnotesize 1.05} & 85.90 $\pm$ {\footnotesize 1.80} & 88.29 $\pm$ {\footnotesize 1.05} & 72.33 $\pm$ {\footnotesize 1.97} & 67.97 $\pm$ {\footnotesize 1.24} & 71.57 $\pm$ {\footnotesize 1.52} & 52.84 $\pm$ {\footnotesize 2.30} & 67.38 $\pm$ {\footnotesize 2.30} & 38.76  $\pm$ {\footnotesize 3.14} & 77.05 $\pm$ {\footnotesize 1.43} & 18.10 $\pm$ {\footnotesize 2.35} & 640 \\
- $N=32$ & 70.53 $\pm$ {\footnotesize 0.80} & 89.20 $\pm$ {\footnotesize 0.49} & 91.14 $\pm$ {\footnotesize 1.85} & 74.04 $\pm$ {\footnotesize 1.67} & 70.06 $\pm$ {\footnotesize 2.99} & 73.73 $\pm$ {\footnotesize 2.57} & 55.93 $\pm$ {\footnotesize 0.85} & 70.32 $\pm$ {\footnotesize 1.79} & 39.22 $\pm$ {\footnotesize 1.19} & 79.29 $\pm$ {\footnotesize 2.44} & 22.99 $\pm$ {\footnotesize 2.93} & 1280 \\
- $N=64$ & 70.90 $\pm$ {\footnotesize 0.77} & 88.80 $\pm$ {\footnotesize 1.67} & 92.56 $\pm$ {\footnotesize 2.21} & 73.18 $\pm$ {\footnotesize 1.49} & 70.63 $\pm$ {\footnotesize 1.53} & 73.98 $\pm$ {\footnotesize 2.34} & 57.73 $\pm$ {\footnotesize 2.80} & 69.79 $\pm$ {\footnotesize 3.04} & 41.74 $\pm$ {\footnotesize 2.47} & 79.29 $\pm$ {\footnotesize 3.23} & 23.28 $\pm$ {\footnotesize 2.21} & 2560 \\
- $N=128$ & 72.60 $\pm$ {\footnotesize 0.63} & 90.40 $\pm$ {\footnotesize 1.50} & 91.46 $\pm$ {\footnotesize 0.32} & 75.75 $\pm$ {\footnotesize 1.33} & 70.91 $\pm$ {\footnotesize 0.87} & 79.70 $\pm$ {\footnotesize 2.06} & 60.57 $\pm$ {\footnotesize 1.73} & 71.93 $\pm$ {\footnotesize 1.79} & 42.89 $\pm$ {\footnotesize 3.00} & 78.73 $\pm$ {\footnotesize 2.87} & 23.85 $\pm$ {\footnotesize 1.49} & 5120 \\
- $N=256$ & 73.47 $\pm$ {\footnotesize 0.36} & 90.00 $\pm$ {\footnotesize 1.47} & 93.04 $\pm$ {\footnotesize 1.17} & 77.35 $\pm$ {\footnotesize 1.24} & 75.00 $\pm$ {\footnotesize 2.55} & 76.27 $\pm$ {\footnotesize 0.92} & 61.73 $\pm$ {\footnotesize 2.16} & 72.46 $\pm$ {\footnotesize 2.16} & 46.56 $\pm$ {\footnotesize 3.00} & 79.66 $\pm$ {\footnotesize 2.14} & 21.55 $\pm$ {\footnotesize 2.97} & 10240 \\
\midrule
FK-Steering - $N=64$  \\ 
- 1-step denoiser look-ahead & 66.13 $\pm$ {\footnotesize 0.45} & 87.70 $\pm$ {\footnotesize 0.91} & 87.82 $\pm$ {\footnotesize 0.94} & 66.88 $\pm$ {\footnotesize 1.83} & 63.69 $\pm$ {\footnotesize 0.99} & 70.05 $\pm$ {\footnotesize 1.87} & 48.97 $\pm$ {\footnotesize 1.50} & 63.77 $\pm$ {\footnotesize 1.79} & 32.34 $\pm$ {\footnotesize 2.28} & 71.83 $\pm$ {\footnotesize 2.00} & 38.22 $\pm$ {\footnotesize 2.86} & 2880  \\
- 1-step flow map look-ahead & 68.81 $\pm$ {\footnotesize 0.21} & 88.80 $\pm$ {\footnotesize 1.20} & 89.24 $\pm$ {\footnotesize 1.34} & 71.26 $\pm$ {\footnotesize 2.17} & 68.35 $\pm$ {\footnotesize 1.48} & 71.19 $\pm$ {\footnotesize 1.81} & 52.96 $\pm$ {\footnotesize 0.99} & 67.11 $\pm$ {\footnotesize 3.22} & 32.80 $\pm$ {\footnotesize 3.00} & 75.93 $\pm$ {\footnotesize 1.62} & 36.78 $\pm$ {\footnotesize 1.82} & 3264  \\
\midrule
\multicolumn{9}{l}{\textbf{Gradient-Based Search}} \\
\midrule
DAS - $N=64$   \\
- 1-step denoiser look-ahead & 66.42 $\pm$ {\footnotesize 1.02} & 86.90 $\pm$ {\footnotesize 1.34} & 87.66 $\pm$ {\footnotesize 0.95} & 68.38 $\pm$ {\footnotesize 0.91} & 64.35 $\pm$ {\footnotesize 2.05} & 69.29 $\pm$ {\footnotesize 1.37} & 52.32 $\pm$ {\footnotesize 3.02} & 63.90 $\pm$ {\footnotesize 2.52} & 33.72 $\pm$ {\footnotesize 0.76} & 72.01 $\pm$ {\footnotesize 2.66} & 32.76 $\pm$ {\footnotesize 1.72} & 2880  \\
- 1-step flow map look-ahead & 68.14 $\pm$ {\footnotesize 0.86} & 88.50 $\pm$ {\footnotesize 0.91} & 89.72 $\pm$ {\footnotesize 0.82} & 70.09 $\pm$ {\footnotesize 1.38} & 66.25 $\pm$ {\footnotesize 1.83} & 70.05 $\pm$ {\footnotesize 2.41} & 54.16 $\pm$ {\footnotesize 1.88} & 63.77 $\pm$ {\footnotesize 1.03} & 33.26 $\pm$ {\footnotesize 2.63} & 73.51 $\pm$ {\footnotesize 1.24} & 34.77 $\pm$ {\footnotesize 2.49} & 1920  \\
\midrule
FMTT - 1-step denoiser look-ahead  \\
- $N=4$ & 62.00 $\pm$ {\footnotesize 4.93} & 85.00 $\pm$ {\footnotesize 3.26} & 84.02 $\pm$ {\footnotesize 4.79} & 66.88 $\pm$ {\footnotesize 4.63} & 60.08 $\pm$ {\footnotesize 4.79} & 63.96 $\pm$ {\footnotesize 7.57} & 42.91 $\pm$ {\footnotesize 5.79} & 58.96 $\pm$ {\footnotesize 3.99} & 24.77 $\pm$ {\footnotesize 5.31} & 69.59 $\pm$ {\footnotesize 7.13} & 28.16 $\pm$ {\footnotesize 6.32} & 180  \\
- $N=8$ & 65.73 $\pm$ {\footnotesize 1.83} & 87.70 $\pm$ {\footnotesize 1.21} & 88.29 $\pm$ {\footnotesize 3.30} & 69.12 $\pm$ {\footnotesize 2.27} & 62.83 $\pm$ {\footnotesize 1.33} & 69.04 $\pm$ {\footnotesize 2.24} & 47.81 $\pm$ {\footnotesize 3.54} & 62.83 $\pm$ {\footnotesize 3.04} & 30.05 $\pm$ {\footnotesize 2.78} & 72.57 $\pm$ {\footnotesize 0.97} & 34.20 $\pm$ {\footnotesize 1.70} & 360  \\
- $N=16$ & 67.75 $\pm$ {\footnotesize 1.82} & 88.20 $\pm$ {\footnotesize 0.87} & 88.92 $\pm$ {\footnotesize 1.14} & 70.09 $\pm$ {\footnotesize 2.16} & 65.49 $\pm$ {\footnotesize 1.33} & 71.70 $\pm$ {\footnotesize 2.92} & 49.23 $\pm$ {\footnotesize 3.45} & 64.57 $\pm$ {\footnotesize 3.77} & 32.57 $\pm$ {\footnotesize 3.52} & 76.49 $\pm$ {\footnotesize 3.05} & 40.80 $\pm$ {\footnotesize 4.34} & 720  \\
- $N=32$ & 69.79 $\pm$ {\footnotesize 1.06} & 90.00 $\pm$ {\footnotesize 1.30} & 90.51 $\pm$ {\footnotesize 1.55} & 71.90 $\pm$ {\footnotesize 2.35} & 67.30 $\pm$ {\footnotesize 1.17} & 73.35 $\pm$ {\footnotesize 1.16} & 52.71 $\pm$ {\footnotesize 1.56} & 67.38 $\pm$ {\footnotesize 1.81} & 37.84 $\pm$ {\footnotesize 5.44} & 77.05 $\pm$ {\footnotesize 2.07} & 39.94 $\pm$ {\footnotesize 1.49} & 1440  \\
- $N=64$ & 70.89 $\pm$ {\footnotesize 1.32} & 90.20 $\pm$ {\footnotesize 1.54} & 90.66 $\pm$ {\footnotesize 1.51} & 73.72 $\pm$ {\footnotesize 0.88} & 68.92 $\pm$ {\footnotesize 2.89} & 73.73 $\pm$ {\footnotesize 2.60} & 56.06 $\pm$ {\footnotesize 1.80} & 66.58 $\pm$ {\footnotesize 1.58} & 37.61 $\pm$ {\footnotesize 4.05} & 78.36 $\pm$ {\footnotesize 3.03} & 43.97 $\pm$ {\footnotesize 2.21} & 2880  \\
- $N=128$ & 72.52 $\pm$ {\footnotesize 1.42} & 91.60 $\pm$ {\footnotesize 0.85} & 92.56 $\pm$ {\footnotesize 0.27} & 73.40 $\pm$ {\footnotesize 2.82} & 69.49 $\pm$ {\footnotesize 2.75} & 76.27 $\pm$ {\footnotesize 1.58} & 58.12 $\pm$ {\footnotesize 2.92} & 69.52 $\pm$ {\footnotesize 2.30} & 42.20 $\pm$ {\footnotesize 5.07} & 81.53 $\pm$ {\footnotesize 2.55} & 42.24 $\pm$ {\footnotesize 0.95} & 5760  \\
- $N=256$ & 72.90 $\pm$ {\footnotesize 0.89} & 91.00 $\pm$ {\footnotesize 1.31} & 90.51 $\pm$ {\footnotesize 1.85} & 73.93 $\pm$ {\footnotesize 1.98} & 70.72 $\pm$ {\footnotesize 1.57} & 78.30 $\pm$ {\footnotesize 0.91} & 60.31 $\pm$ {\footnotesize 0.82} & 69.65 $\pm$ {\footnotesize 3.37} & 42.89 $\pm$ {\footnotesize 4.12} & 81.53 $\pm$ {\footnotesize 1.70} & 39.94 $\pm$ {\footnotesize 1.88} & 11520  \\

\midrule

FMTT - 1-step flow map look-ahead  \\
- $N=16$ & 63.28 $\pm$ {\footnotesize 0.77} & 85.10 $\pm$ {\footnotesize 0.95} & 85.28 $\pm$ {\footnotesize 1.57} & 67.09 $\pm$ {\footnotesize 2.58} & 60.93 $\pm$ {\footnotesize 1.55} & 67.13 $\pm$ {\footnotesize 1.31} & 45.36 $\pm$ {\footnotesize 3.75} & 61.90 $\pm$ {\footnotesize 1.33} & 27.06 $\pm$ {\footnotesize 1.38} & 69.22 $\pm$ {\footnotesize 1.43} & 27.87 $\pm$ {\footnotesize 3.39} & 240  \\
- $N=32$ & 67.55 $\pm$ {\footnotesize 0.86} & 88.60 $\pm$ {\footnotesize 0.82} & 88.92 $\pm$ {\footnotesize 1.05} & 68.06 $\pm$ {\footnotesize 1.69} & 65.97 $\pm$ {\footnotesize 1.46} & 70.94 $\pm$ {\footnotesize 0.83} & 51.93 $\pm$ {\footnotesize 1.17} & 65.37 $\pm$ {\footnotesize 0.95} & 31.65 $\pm$ {\footnotesize 3.70} & 75.37 $\pm$ {\footnotesize 3.34} & 36.49 $\pm$ {\footnotesize 4.33} & 480  \\
- $N=64$ & 71.13 $\pm$ {\footnotesize 0.62} & 89.60 $\pm$ {\footnotesize 0.63} & 91.14 $\pm$ {\footnotesize 0.45} & 73.40 $\pm$ {\footnotesize 2.17} & 67.87 $\pm$ {\footnotesize 2.26} & 75.25 $\pm$ {\footnotesize 1.98} & 57.35 $\pm$ {\footnotesize 1.52} & 68.72 $\pm$ {\footnotesize 1.34} & 37.39 $\pm$ {\footnotesize 3.39} & 77.05 $\pm$ {\footnotesize 2.66} & 45.11 $\pm$ {\footnotesize 3.39} & 960  \\
- $N=128$ & 73.32 $\pm$ {\footnotesize 0.68} & 91.90 $\pm$ {\footnotesize 0.52} & 90.98 $\pm$ {\footnotesize 0.94} & 76.39 $\pm$ {\footnotesize 2.01} & 71.29 $\pm$ {\footnotesize 1.46} & 75.63 $\pm$ {\footnotesize 1.61} & 59.79 $\pm$ {\footnotesize 1.67} & 69.65 $\pm$ {\footnotesize 0.95} & 42.43 $\pm$ {\footnotesize 1.76} & 80.04 $\pm$ {\footnotesize 1.93} & 46.84 $\pm$ {\footnotesize 2.86} & 1920  \\
- $N=256$ & 74.86 $\pm$ {\footnotesize 0.50} & 91.90 $\pm$ {\footnotesize 0.77} & 92.72 $\pm$ {\footnotesize 1.30} & 77.78 $\pm$ {\footnotesize 1.89} & 71.86 $\pm$ {\footnotesize 0.60} & 77.41 $\pm$ {\footnotesize 2.17} & 62.50 $\pm$ {\footnotesize 1.76} & 72.59 $\pm$ {\footnotesize 1.53} & 46.33 $\pm$ {\footnotesize 1.38} & 81.72 $\pm$ {\footnotesize 2.14} & 46.55 $\pm$ {\footnotesize 4.10} & 3840  \\

\midrule
FMTT - 1-step Schnell look-ahead  \\
- $N=256$ & 77.00 $\pm$ {\footnotesize 0.29} & 92.40 $\pm$ {\footnotesize 0.49} & 93.67 $\pm$ {\footnotesize 1.00} & 78.95 $\pm$ {\footnotesize 0.70} & 74.43 $\pm$ {\footnotesize 0.87} & 80.20 $\pm$ {\footnotesize 0.88} & 66.11 $\pm$ {\footnotesize 2.04} & \textbf{74.87 $\pm$ {\footnotesize 1.77}} & \textbf{51.38 $\pm$ {\footnotesize 4.35}} & 84.14 $\pm$ {\footnotesize 1.33} & 47.70 $\pm$ {\footnotesize 4.63} & 3840  \\
\midrule

FMTT - 2-step flow map look-ahead  \\
- $N=16$ & 64.73 $\pm$ {\footnotesize 0.74} & 87.40 $\pm$ {\footnotesize 2.09} & 87.34 $\pm$ {\footnotesize 2.15} & 65.60 $\pm$ {\footnotesize 2.35} & 61.79 $\pm$ {\footnotesize 0.83} & 68.27 $\pm$ {\footnotesize 1.81} & 47.94 $\pm$ {\footnotesize 1.63} & 61.90 $\pm$ {\footnotesize 1.38} & 30.50 $\pm$ {\footnotesize 0.76} & 71.27 $\pm$ {\footnotesize 3.10} & 33.33 $\pm$ {\footnotesize 2.15} & 360  \\
- $N=32$ & 68.04 $\pm$ {\footnotesize 1.34} & 89.60 $\pm$ {\footnotesize 0.57} & 90.35 $\pm$ {\footnotesize 0.82} & 70.19 $\pm$ {\footnotesize 2.55} & 64.83 $\pm$ {\footnotesize 4.10} & 69.80 $\pm$ {\footnotesize 1.16} & 51.80 $\pm$ {\footnotesize 1.06} & 66.84 $\pm$ {\footnotesize 2.33} & 33.26 $\pm$ {\footnotesize 4.27} & 73.69 $\pm$ {\footnotesize 2.50} & 39.08 $\pm$ {\footnotesize 3.45} & 840  \\
- $N=64$ & 71.46 $\pm$ {\footnotesize 0.40} & 90.10 $\pm$ {\footnotesize 1.07} & 91.46 $\pm$ {\footnotesize 2.12} & 74.15 $\pm$ {\footnotesize 1.11} & 68.25 $\pm$ {\footnotesize 2.49} & 74.49 $\pm$ {\footnotesize 1.50} & 56.70 $\pm$ {\footnotesize 2.03} & 69.65 $\pm$ {\footnotesize 1.28} & 38.76 $\pm$ {\footnotesize 1.88} & 77.99 $\pm$ {\footnotesize 2.87} & 44.83 $\pm$ {\footnotesize 1.63} & 1680  \\
- $N=128$ & 74.31 $\pm$ {\footnotesize 0.78} & 91.20 $\pm$ {\footnotesize 1.20} & 92.56 $\pm$ {\footnotesize 0.82} & 74.79 $\pm$ {\footnotesize 2.28} & 71.48 $\pm$ {\footnotesize 2.81} & 78.05 $\pm$ {\footnotesize 0.55} & 62.89 $\pm$ {\footnotesize 1.26} & 72.46 $\pm$ {\footnotesize 1.26} & 46.33 $\pm$ {\footnotesize 2.00} & 80.78 $\pm$ {\footnotesize 2.32} & 45.98 $\pm$ {\footnotesize 2.44} & 3360  \\
- $N=256$ & 75.55 $\pm$ {\footnotesize 0.53} & 91.90 $\pm$ {\footnotesize 0.77} & 92.88 $\pm$ {\footnotesize 1.51} & 78.21 $\pm$ {\footnotesize 1.09} & 72.34 $\pm$ {\footnotesize 1.12} & 79.19 $\pm$ {\footnotesize 1.87} & 65.08 $\pm$ {\footnotesize 1.98} & 72.73 $\pm$ {\footnotesize 0.85} & 46.79 $\pm$ {\footnotesize 1.45} & 81.53 $\pm$ {\footnotesize 1.10} & 47.70 $\pm$ {\footnotesize 3.09} & 6720  \\

\midrule

FMTT - 4-step flow map look-ahead  \\
- $N=16$ & 65.90 $\pm$ {\footnotesize 0.45} & 86.90 $\pm$ {\footnotesize 1.21} & 84.34 $\pm$ {\footnotesize 1.64} & 68.27 $\pm$ {\footnotesize 1.55} & 63.97 $\pm$ {\footnotesize 1.43} & 69.92 $\pm$ {\footnotesize 1.81} & 49.23 $\pm$ {\footnotesize 1.65} & 63.90 $\pm$ {\footnotesize 0.96} & 31.42 $\pm$ {\footnotesize 1.76} & 75.37 $\pm$ {\footnotesize 0.53} & 32.47 $\pm$ {\footnotesize 1.25} & 600  \\
- $N=32$ & 68.12 $\pm$ {\footnotesize 0.62} & 89.00 $\pm$ {\footnotesize 1.00} & 87.03 $\pm$ {\footnotesize 2.22} & 70.51 $\pm$ {\footnotesize 1.17} & 65.02 $\pm$ {\footnotesize 2.56} & 71.07 $\pm$ {\footnotesize 1.83} & 52.19 $\pm$ {\footnotesize 1.33} & 66.84 $\pm$ {\footnotesize 0.38} & 34.17 $\pm$ {\footnotesize 1.00} & 74.81 $\pm$ {\footnotesize 2.14} & 40.52 $\pm$ {\footnotesize 4.25} & 1560  \\
- $N=64$ & 71.48 $\pm$ {\footnotesize 0.29} & 89.80 $\pm$ {\footnotesize 0.72} & 90.82 $\pm$ {\footnotesize 0.78} & 72.86 $\pm$ {\footnotesize 0.48} & 69.39 $\pm$ {\footnotesize 0.69} & 74.49 $\pm$ {\footnotesize 2.19} & 56.83 $\pm$ {\footnotesize 2.20} & 69.12 $\pm$ {\footnotesize 0.95} & 40.14 $\pm$ {\footnotesize 2.78} & 78.92 $\pm$ {\footnotesize 0.97} & 43.97 $\pm$ {\footnotesize 2.05} & 3120  \\
- $N=128$ & 74.32 $\pm$ {\footnotesize 0.32} & 90.90 $\pm$ {\footnotesize 0.71} & 90.55 $\pm$ {\footnotesize 0.78} & 78.31 $\pm$ {\footnotesize 0.56} & 70.63 $\pm$ {\footnotesize 0.78} & 76.65 $\pm$ {\footnotesize 1.19} & 60.44 $\pm$ {\footnotesize 1.72} & 74.20 $\pm$ {\footnotesize 1.38} & 46.56 $\pm$ {\footnotesize 3.00} & 80.22 $\pm$ {\footnotesize 0.83} & 49.43 $\pm$ {\footnotesize 2.82} & 6240  \\
- $N=256$ & 75.14 $\pm$ {\footnotesize 0.80} & 91.40 $\pm$ {\footnotesize 1.18} & 92.41 $\pm$ {\footnotesize 1.18} & 77.03 $\pm$ {\footnotesize 1.72} & 71.39 $\pm$ {\footnotesize 2.16} & 79.19 $\pm$ {\footnotesize 1.48} & 65.08 $\pm$ {\footnotesize 3.29} & 72.46 $\pm$ {\footnotesize 2.32} & 45.87 $\pm$ {\footnotesize 2.25} & 80.41 $\pm$ {\footnotesize 1.10} & \textbf{50.86 $\pm$ {\footnotesize 2.62}} & 12480  \\

\bottomrule
\end{tabular}
}
\end{table}

\newpage
\section{Broader Impact}
Flow Map Trajectory Tilting contributes to reward-aligned generation with diffusion models, enabling stronger results without retraining and potentially reducing the computational cost of adapting models to new objectives. While we focus on text-to-image generation, the framework is broadly applicable to other modalities wherever suitable reward feedback can be defined. However, generative models and reward-guided methods can be misused for producing deceptive imagery or optimizing toward harmful objectives, and may inherit biases from reward models. Such techniques should be deployed with appropriate safeguards and caution.\looseness=-1

\section{Limitations}
As discussed in the main paper, FMTT is a gradient-based method and therefore requires a differentiable reward function. As a result, it can be slower than purely gradient-free methods under the same hyperparameters, and it is not directly applicable to non-differentiable rewards. In addition, flow map look-ahead requires access to a pretrained flow map or a few-step distilled model, which may introduce an additional distillation cost when such a model is not already available. Finally, in text-to-image    experiments, FMTT can require up to several thousand function evaluations, which is orders of magnitude more expensive than sampling from the base flow map model alone. A promising direction for future work is to combine FMTT with online RL-based diffusion fine-tuning methods, which could amortize this test-time cost.

\section{Licenses}
Our models and code are built upon the following codebases and datasets:
\begin{itemize}
    \item \textbf{Diffusers} (\url{https://github.com/huggingface/diffusers}): Used for our text-to-image experiments. Licensed under the Apache License 2.0. We build on top of \texttt{FLUX.1-dev}, which is under a proprietary non-commercial license (\url{https://huggingface.co/black-forest-labs/FLUX.1-dev/blob/main/LICENSE.md}).
    
    \item \textbf{GenEval} 
    (\url{https://github.com/djghosh13/geneval}): Used as a validation benchmark. Licensed under the MIT license. 
    
    \item \textbf{UniGenBench++} 
    (\url{https://github.com/CodeGoat24/UniGenBench}): Used as a validation benchmark. Licensed under the CC BY 4.0 license. 
    
    \item \textbf{Skywork-VL} 
    (\url{https://huggingface.co/Skywork/Skywork-VL-Reward-7B}): Used as a VLM-based reward model. Licensed under the MIT license. 
    
    \item \textbf{Qwen2.5-VL} 
    (\url{https://huggingface.co/Qwen/Qwen2.5-VL-7B-Instruct}): Used as a VLM-based reward model. Licensed under the Apache 2.0 license.
    
    \item \textbf{Pickscore} 
    (\url{https://github.com/yuvalkirstain/pickscore}): Used as a human preference reward model. Licensed under the MIT license. 

    \item \textbf{HPSv2} 
    (\url{https://github.com/tgxs002/HPSv2}): Used as a human preference reward model. Licensed under the Apache 2.0 license. 

    \item \textbf{ImageReward} 
    (\url{https://github.com/zai-org/ImageReward}): Used as a human preference reward model. Licensed under the Apache 2.0 license. 

    \item \textbf{CLIP} 
    (\url{https://huggingface.co/laion/CLIP-ViT-H-14-laion2B-s32B-b79K}): Used as a text-image alignment reward model. Licensed under the MIT license. 
    
\end{itemize}

\end{document}